\documentclass{article}
\usepackage{arxiv}
\usepackage{graphics}
\usepackage{array}
\usepackage{macros}
\usepackage{csvsimple}
\usepackage{enumerate}
\usepackage{float}
\usepackage{enumitem}
\usepackage{booktabs}
\usepackage{colortbl}
\usepackage[table]{xcolor}

\usepackage{algorithm}
\usepackage{algpseudocode}
\usepackage{url,hyperref}

\usepackage{amsmath}
\usepackage{microtype}
\usepackage{subfigure}
\usepackage{amsmath}
\usepackage{amssymb}
\usepackage{mathtools}
\usepackage{amsthm}
\usepackage{bbm}
\graphicspath{ {./images/} }
\newcommand{\xhdr}[1]{\vspace{0.1mm}\noindent{{\bf #1.}}}

\title{GNUMAP: A Parameter-Free Approach to Unsupervised Dimensionality Reduction via Graph Neural Networks}

\author{
 Jihee You \\
   Data Science Institute\\
  University of Chicago\\
 Chicago, IL 60605 \\
  \texttt{jiheeyou@uchicago.edu} \\
   \And
 So Won Jeong \\
  Booth School of Business\\
  University of Chicago\\
 Chicago, IL 60605 \\
  \texttt{sowonjeong@uchicago.edu} \\
  \And
 Claire Donnat \\
  Department of Statistics\\
  University of Chicago\\
 Chicago, IL 60605 \\
  \texttt{cdonnat@uchicago.edu} \\
}

\begin{document}
\maketitle

\begin{abstract}
%Unsupervised node representation learning has recently undergone significant progress, particularly through 
With the proliferation of Graph Neural Network (GNN) methods stemming from contrastive learning, 
unsupervised node representation learning for graph data is rapidly gaining traction across various fields, from biology to molecular dynamics, where it is often used as a dimensionality reduction tool. However, 
%despite the widespread adoption and development of these methods, 
there remains a significant gap in understanding the quality of the low-dimensional node representations these methods produce, particularly beyond well-curated academic datasets. To address this gap, we propose here the first comprehensive benchmarking of various unsupervised node embedding techniques tailored for dimensionality reduction, encompassing a range of manifold learning tasks, along with various performance metrics. \textit{We emphasize the sensitivity of current methods to hyperparameter choices --- highlighting a fundamental issue as to their applicability in real-world settings where there is no established methodology for rigorous hyperparameter selection.} Addressing this issue, we introduce GNUMAP, a robust and parameter-free method for unsupervised node representation learning that merges the traditional UMAP approach with the expressivity of the GNN framework. We show that GNUMAP consistently outperforms existing state-of-the-art GNN embedding methods in a variety of contexts, including synthetic geometric datasets, citation networks, and real-world biomedical data --- making it a simple but reliable dimensionality reduction tool. %This paper highlights the critical importance of carefully selecting the node representation learning technique and paves the way for a more robust assessment of the potential use of GNNs as a dimensionality reduction tool.
%%hard to argue it outperforms road network
\end{abstract}

\maketitle

\section{Introduction}
Consider the following biological problem: given a set of gene expressions at various locations within a tissue sample (for instance, a slice of mouse brain tissue), how can we aggregate information to discover spatial domains with coherent gene expression patterns? Recent methods, including \cite{hu2021spagcn} from which this example is adapted, have turned towards using a graph-based approach --- and subsequently, Graph Neural Networks (GNNs) --- to resolve this conundrum. In this setting, the data is first represented as a graph $\mathcal{G}$ on $n$ nodes corresponding here to the various spatial locations within the sample, each endowed with a feature vector $X_v \in \R^d$  (the gene expression data). Under this new formalism, Graph Neural Networks (GNNs) \cite{scarselli2008graph, Kipf2016semi} come as a natural tool for visualization and the subsequent discovery of new patterns in the data. Through the use of recursive neighborhood ``convolutions'', GNNs allow indeed the creation of rich node representations that capture topological information, feature data and essential neighborhood properties, and integrate this information in a Euclidean vector representation that is amenable to any downstream machine learning task.
%Heralded as the breakthrough for machine learning on graphs that would allow the same ``AI renaissance’' \cite{battaglia2018relational} that standard neural networks have brought to Computer Vision and Natural Language Processing, GNNs have been suggested as a panacea for a wide number of tasks across disciplines, including molecular design \cite{stokes2020deep,duvenaud2015convolutional},  traffic prediction \cite{cui2019high,casas2019spatially}, biological networks \cite{zitnik2018modeling,ma2019genn,li2021representation} or recommender systems. \\

\xhdr{Unsupervised Learning on Graph Data}
While the early rise in Graph Neural Network (GNN) advancements predominantly emphasized supervised learning, there has been growing interest in developing unsupervised GNN methods for learning node representations. This shift is driven by the scarcity of labeled data in numerous real-world situations, such as in the example described above, along with an increasing demand for techniques that facilitate exploratory data analysis and dimensionality reduction for graph data. However, there seems to have been a concurrent development of unsupervised learning approaches in the application community \cite{hu2021spagcn,partel2021spage2vec,zhang2020dango,li2021scgslc,ishiai2024graph} and in the method community \cite{cca-ssg, dgi, bgrl, grace,hong2023simplified}. This parallel development may underscore a disconnect between the approaches devised by the methods community and the practical reality of real-world data. Specifically, the deployment of state-of-the-art approaches evaluated on academic benchmarks appears to be impeded by two significant challenges: \textit{(a) Ease of deployment}, and \textit{(b) Trustworthiness of the learned representations}.

\textit{(a) Ease of deployment. }From a methodological point of view, a growing number of recent approaches are leaning toward the adoption of self-supervised contrastive learning frameworks to effectively represent nodes \cite{cca-ssg, grace, bgrl, dgi,hong2023simplified}. In this setting, the trick usually consists of perturbing the input data (e.g., by masking features with probability $p_f$, dropping edges with probability $p_e$) to create two modified versions of the original data. Each of these perturbed versions of the data is then processed by a Graph Neural Network, which is trained to identify pairs of node embeddings from the two perturbed graphs that correspond to the same node in the original dataset. Examples of such training objectives include the loss proposed by \cite{grace} for their method GRACE, $\mathcal{L} = \frac{1}{2N} \left( \sum_{i=1}^N \left[ \ell( u_i,v_i)+\ell(v_i,u_i)\right] \right),$ 
where
$\ell(u_i,v_i)=\log\left(\frac{e^{s(u_i,v_i)/\tau}}{\sum_{k=1}^{N}\left[ \mathbbm{1}_{\{k\neq i\}} e^{s(u_i,u_k)/\tau}+e^{s(u_i,v_k)/\tau}\right]}\right)$, $s:\mathbb{R}^{p}\times \mathbb{R}^{p}\rightarrow \mathbb{R}$ is the cosine similarity function,  and $u_i,v_i\in\mathbb{R}^p$ are the node representation for node $i$ stemming from the two perturbed versions of the graph. Finally, $\tau>0$ is a temperature parameter that must be tuned. Alternatives include the loss used in CCA-SSG \cite{cca-ssg}, $\mathcal{L}=\sum_{i=1}^n\|u_i - v_i\|^2+\lambda\left( \| U^T U- I \|_F^2+\| V^T V-I \|_F^2 \right)$
where $U,V\in \mathbb{R}^{N\times p}$ are the node representation matrix of two views, $\lambda>0$ is a hyperparameter, and $\| \cdot \|_F$ denotes the Frobenius norm. 
While these self-supervised learning losses have achieved state-of-the-art performance in a number of academic benchmarks, few of these methods have yet been deployed in applied settings.

\begin{figure*}[h!]
    \centering
    \includegraphics[width=\textwidth]{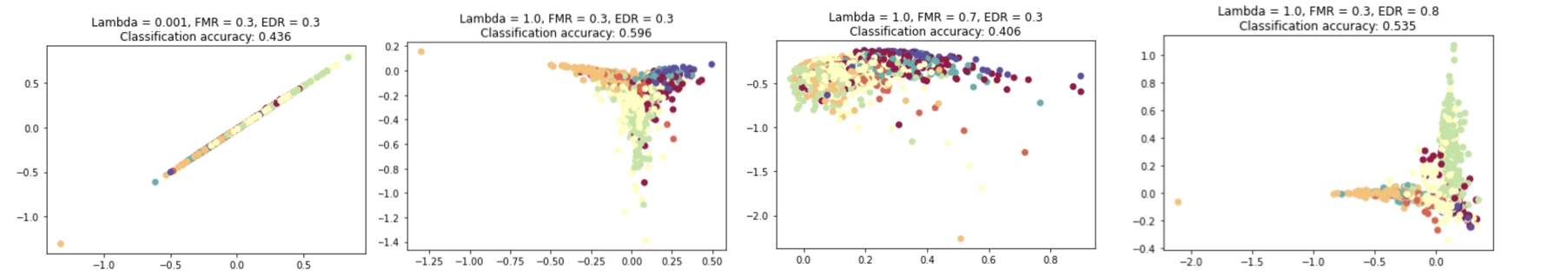}
    \caption{Node representation learning for Cora using CCA-SSG \cite{cca-ssg}. Colors represent classes. Classification accuracy was established by running a support vector machine classifier on the learned 2D node representations, using 5-fold cross-validation to fix the kernel bandwidth. We note a substantial variation in embedding quality as the parameters (regularization lambda, feature mask rate, edge drop rate)  vary.}
    \label{fig:cora}
\end{figure*}

One hypothesis explaining this gap is the heavy reliance of these "state-of-the-art" approaches on the correct choice of hyperparameters (for instance, $\lambda, \tau, p_f$ or $p_e$ in the approaches described above). We exemplify this phenomenon through an example case in Figure~\ref{fig:cora}, where we propose to deploy CCA-SSG \cite{cca-ssg} using a simple 2-layer GCN \cite{Kipf2016semi} to learn a 2D visualization of the nodes. We then assess the quality of the learned embeddings by learning a support vector classifier to classify the nodes, and evaluate the performance of the classification on held-out data. We note a substantial variation in embedding quality as the edge drop rate $p_e$, the feature mask rate $p_m$ and the regularization parameter $\lambda$ vary. This highlights the importance of selecting the ``correct'' set of hyperparameters: a wrong choice of hyperparameters could cause the method to significantly underperform or to learn uninformative embeddings (see Figure~\ref{fig:cora} left).
However, in the unsupervised context, there is no established cross-validation technique for GNNs or other principled technique for parameter selection at large. This significantly complicates the practical application of these methods in settings where there are no labels to evaluate the method on for hyperparameter selection.

\textit{(b) Trustworthiness.} 
Moreover, despite the increasing adoption of graph neural networks (GNNs), there is a notable gap in research focusing on the evaluation of the quality of GNN node representations: How effectively do these embeddings capture the structural details within the data? Are they capable of accurately encoding topological information? Hypothetically, a good embedding should preserve both local and global structure in the graph while conveying a maximum amount of information. 
%Some of those qualities have already been explored in the dimensionality reduction techniques research. In particular, Uniform Manifold Approximation and Projection (UMAP) \cite{umap} is the state-of-the-art technique that seeks to preserve local and global structure, and have been benchmarked on a number of tasks, including manifold learning and classification tasks. 
Beyond the graph setting, the need to benchmark new unsupervised approaches is underscored by the multiplication of recent publications in applied domains providing tips for performing a dimensionality reduction \cite{nguyen2019ten} or comparing existing tools \cite{xiang2021comparison,babjac2022comparison,malepathirana2022dimensionality}: with the rapidly increasing number of new methods, it becomes difficult to know which one to adopt --- particularly when these methods can be quite sensitive.
Many applications --- particularly in biology --- have a long track record of using graph-based techniques such as UMAP \cite{umap} and t-SNE \cite{tsne} for dimensionality reduction. These methods have been well established, understood, and have been vetted by the myriad of applications and benchmarking tasks, including manifold learning and classification tasks, to which they have been deployed. On the other hand, while GNNs offer the potential for richer representation learning by capturing both feature and node information, these methods have not yet attained a similar level of reliability.
This gap underscores {\it (a) the need for a comparison of GNN-based approaches to current methods}, allowing practitioners to place GNNs in the landscape of unsupervised learning methods. While GNNs offer more flexibility as they allow combining both graph data and node covariates, this comparison is nevertheless indispensable to start evaluating them as dimensionality reduction techniques; and {\it (b) the need for more in-depth analysis and validation of GNNs in a variety of settings that go beyond node classification benchmarks}, and incorporate a wider variety of tasks -- including learning manifolds.

\xhdr{Contributions} To fill this gap, we propose here a systematic comparison of learned node representations, focusing more specifically on the context of dimensionality reduction and data visualization. To bridge classical dimensionality reduction methods that do not incorporate node features with state-of-the-art GNN-based approaches, we first introduce an unsupervised learning technique inspired by UMAP \cite{mcinnes2018umap-software}, named GNUMAP, which integrates the UMAP framework for learning low-dimensional embeddings while being less parameter-intensive compared to contrastive learning methods. This approach facilitates a comparative analysis of self-supervised and reconstruction-based unsupervised embedding methods. We then propose to evaluate existing methods on two sets of tasks: (a) a suite of manifold learning tasks, thereby allowing us to compare GNN methods with more established benchmarks; and (b) classification tasks; Our goal is to provide practitioners with a clearer understanding of effective approaches in the unsupervised application of GNNs, thereby promoting their broader use across various applications.

\section{GNUMAP: Bridging classical dimensionality reduction and Graph Neural Networks}

In this section, we introduce Graph-Neural UMAP (GNUMAP), a method that allows us to bridge the Graph Neural Network (GNN) framework with established dimensionality reduction techniques proven effective on real-world data. We begin by briefly reviewing UMAP, before extending it to the analysis of graph data. 

\xhdr{UMAP} UMAP \cite{umap}, is a standard technique for dimensionality reduction of Euclidean data denoted as \(\{x_i\}_{i=1}^n \in \R^{p_h}\), and relies partly on the graph formalism. The first step of the method consists indeed in constructing a nearest-neighbor graph based on the data points $\{x_i\}_{i=1}^n$. The connection probability between nodes in this graph is given by the formula 
\begin{equation}\label{eq:umap-high-dim-prob}
    p_{ij} = p_{i|j} + p_{j|i} - p_{j|i} \times p_{i|j},
\end{equation} where 

\begin{equation}\label{eq:p_ij}
p_{j|i} = \exp\left(-\frac{d(x_i, x_j)-\rho_i}{\sigma_i}\right).
\end{equation}
In this context, $x_i \in \mathbb{R}^{p_h}$ denotes the original high-dimensional coordinates of input data point \(i\), \(\rho_i\) represents the distance to the nearest neighbor of point \(i\), and \(\sigma_i\) represents the local density factor around point $i$.
UMAP finds a low-dimensional representation $y_i \in \mathbb{R}^{p_l}, p_l <\!\!< p_h$ for each point $x_i$ that minimizes the cross-entropy between connection probabilities in the high and low-dimensional space. In the low-dimensional space, UMAP uses the family of curves \(f(x) = \frac{1}{1 + \alpha x^{2\beta}}\) to compute the connection probability analogous to \(p_{ij}\), which is expressed as 
\begin{equation}\label{eq:q_{ij}}
q_{ij} = \frac{1}{1 + \alpha \times d(y_i, y_j)^{2\beta}}.
\end{equation}
%where \(y_i \in \R^{p_l}\) represents the low-dimensional coordinates of data point \(i\) (with $p_l <<p_h$).
By default, UMAP assumes the spread of the low-dimensional embeddings is 1 and the desired minimum distance between embeddings is 0.1. In such case, UMAP notes that \(f(x) = \frac{1}{1 + \alpha x^{2\beta}}\) where \(\alpha=1.57\) and \(\beta=0.89\) best models the low-dimensional connection probability. While \(\alpha=1.57\) and \(\beta=0.89\) are the default parameters, the authors of UMAP allow customizing of \(\alpha\) and \(\beta\) based on the desired embedding spread and minimum distance.

The algorithm then computes the cross-entropy loss between the pairwise connection probabilities in both high (\(p_{ij}\)) and low (\(q_{ij}\)) dimensions. This cross-entropy loss is formulated as:
\[\mathcal{L} = -\sum_{i} \sum_{j} \left[p_{ij} \log(q_{ij}) + (1 - p_{ij}) \log(1 - q_{ij}) \right]\]

This approach enables UMAP to find a low-dimensional representation while preserving the intrinsic topological structure of the data.

\textbf{GNUMAP.} By contrast, in our proposed adaptation of UMAP to the graph setting, the input of the algorithm is already a graph with either binary or weighted edges. We will denote this given input graph as $\mathcal{G} = (\mathcal{V}, \mathcal{E}, X)$, where $\mathcal{V} \in \mathbb{R}^n$ is a set of $n$ nodes, $\mathcal{E}$ is a set of edges along with their connection probability, and $X \in \mathbb{R}^{n \times p_h}$ be the $p_h$-dimensional node feature matrix. This inherent graph structure of the input allows us to bypass the initial step of traditional UMAP, which converts data into a graph with connection probability defined as in Equation~\ref{eq:umap-high-dim-prob}. Similar to UMAP, our approach focuses on achieving dimensionality reduction by identifying a low-dimensional representation that preserves the topology of the original graph. 

\emph{(a) High Dimensional Node Connectivity.} Let $\mathbf{A} \in \mathbb{R}^{n \times n}$ be the input graph's (possibly weighted) adjacency matrix. We replace the connection probability  $p_{ij}$ in Equation~\ref{eq:p_ij} by $A_{ij}$, the {\it observed } (and potentially weighted) connection between node ${i}$ and ${j}$. This alleviates in particular any ambiguity on the choice of the parameters $\rho_i$ and $\sigma_i$ in the original method. Note here that compared to the original UMAP algorithm, if the adjacency matrix is binary, the $p_{ij}$s take values in $\{0,1\}$.%$\mathbf{P}$ is a sparse matrix since ${p_{ij} = 0}$ when there is no edge between node ${i}$ and ${j}$ in the input graph.

\emph{(b) Low Dimensional Node Connectivity.} Let $y_{i}$ denote the GNN embedding of datapoint $i$ ($y_i = \text{GNN}(x_i, \mathcal{N}(i))$, where $\mathcal{N}(i)$ denote the neighborhood of point $i$. We further feed the GNN outputs into a differentiable batch normalization layer that scales and decorrelates the input features to reduce internal covariate shift. The effect of the DBN layer is demonstrated in Table~\ref{table:dbnchoice} and Figure~\ref{fig:dbn} in the Appendix. 

Then, $$q_{ij} = \frac{1}{1 + \alpha \times d(y_{i}, y_{j})^{2\beta}}$$ denotes the weight of the low-dimensional node connections between $y_i$ and $y_j$. This way of defining the low-dimensional connection strength $q_{ij}$ draws significant inspiration from UMAP. As our DBN objective ensures the embedding distances are normalized, the default UMAP minimum distance of 0.1 allows interpretability in our setting, and therefore we incorporated UMAP's default constants \(\alpha=1.57\) and \(\beta=0.89\). These values allow indeed smaller tails (so more spread-out embeddings) than a baseline choice of $\alpha=\beta=1$ (see Figure~\ref{fig:tails} in the Appendix). For a demonstration of GNUMAP performance with varying \(\alpha\) and \(\beta\), see Table~\ref{table:ab} in the Appendix.

\emph{(c) Loss Calculation.} GNUMAP employs a cross-entropy loss between high and low dimensional node connectivity, so the loss becomes
\begin{equation}\label{eq:gnumap}
\mathcal{L} = -\sum_{i} \sum_{j} \left[p_{ij} \log(q_{ij}) + (1 - p_{ij}) \log(1 - q_{ij}) \right]
\end{equation}

For scalability purposes, GNUMAP downsamples the loss corresponding to absent edges, and samples as many negatives as positives to compute the loss. The procedure is summarized in Algorithm~\ref{alg:cap}.\\

\begin{algorithm}
\caption{GNUMAP Algorithm for Output Dimension $p_l$}\label{alg:cap}
\begin{algorithmic}[1]
    \State \textbf{Inputs:} Adjacency matrix \(\mathbf{A} \in \mathbb{R}^{n \times n}\)
    \State \(n_{\text{pos}} = \text{count}(p_{ij} > 0 \text{ in } \mathbf{P})\)
    \State \(E_{\text{pos}} = \{ (i, j) \in \mathcal{V} \times \mathcal{V} : p_{ij} > 0 \}\)
    \For {epochs = 1 to 400}
        \State Sample \(n_{\text{neg}}\) negative edges, where \(n_{\text{neg}} = n_{\text{pos}}\)
        \State Compute node embeddings through a graph convolution network:  $\mathcal{Y}_{\text{GCN}}^{n \times p_l} = \text{GCN(Features, Edge Index)}$
        \State Apply differentiable batch normalization: 
        $\mathcal{Y}_{\text{DBN}}^{n \times p_l} = \text{DBN}(\mathcal{Y}_{\text{GCN}})$
        \State Compute \(d(y_i, y_j)\) across all \((i, j) \in E_{\text{neg}} \cup E_{\text{pos}}\), then append to create \(D\)
        \State Compute the corresponding low-dimensional connection probability:     ${Q} = \frac{1}{1 + \alpha D^{2\beta}} \in \mathbb{R}^{1 \times (n_{\text{neg}} + n_{\text{pos}})}$
        \State Access \(\mathbf{A}\) at positive edge indices and negative sampled indices to get high-dimensional connection probability
        \State \hspace{0.5cm} \(P \in \mathbb{R}^{1 \times (n_{\text{neg}} + n_{\text{pos}})}\)
        \State Compute \(\mathcal{L} = \) Cross-Entropy(\(P, Q\)) and Backpropagate \(\mathcal{L}\)
    \EndFor
\end{algorithmic}
\end{algorithm}

\xhdr{Remark 1} \textbf{GNUMAP is an auto-encoder}. Using the reconstruction objective in Equation~\ref{eq:gnumap}, it is is obvious that GNUMAP is simply an auto-encoder \cite{zhai2018autoencoder}. Auto-encoders are a class of reconstruction-based techniques that aim to find a low-dimensional representation of the data such that the distance or similarity between learned node embeddings is predictive of the existence of an edge between nodes in the original graph. The best known version of this type of approach, GAE, along with its variational version, VGAE, were first suggested in \cite{vgae}. However, our proposed adaptation of UMAP differs slightly from these methods in several aspects: (i) the probability $q_{ij}$ is defined differently.
Indeed, in GAE, $q_{ij}$ is a function of the inner product between node representations: $$q^{GAE}_{ij} = \big(1  + e^{-X_i^TX_j}\big)^{-1} = \big( 1+ e^{ \frac{d(X_i,X_j)^2 - \| X_i\|^2 -  \| X_j\|^2}{2}} \big)^{-1}.$$ 

Edge probabilities therefore depend both of the distance $d(X_i,X_j)$ and on the norms $\|X_i\|_2$ and $\|X_j\|_2$ of the embeddings themselves. Consequently, embeddings close to the origin will typically exhibit a higher number of connections than embeddings farther from the origin. This allows the model to  potentially build in degree heterogeneity, and places all high degree nodes towards the origin.   However, in our proposed construction, this probability is solely a function of the distance. This can be beneficial, particurlarly when we expect the graph to have several cores, whose center (high degree nodes) should be placed far from one another. We will come back to this point in our synthetic experiments in section 3. (ii) GNUMAP uses a whitening step that allows to regularize the objective function. GNUMAP's differentiable batch normalization design closely follows that of CLGR \cite{hong2023simplified} since the effectiveness of batch whitening was already demonstrated in the previous work. \\
\xhdr{Remark 2} Contrary to self-supervised techniques for node embeddings, this proposed method does not rely on the choice of specific hyperparameters to perform well.\\
\xhdr{Remark 3} The driving hypothesis that GNUMAP leverages is that connected nodes should be close. GNUMAP is therefore amenable to homophilic networks (where adjacent nodes are presumed to be similar), rather than heterophilic networks --- a limitation of our framework compared to self-supervised learning techniques.\\
\xhdr{Remark 4} Another method related to this direct extension of UMAP is SpaGCN \cite{hu2021spagcn}. Developed for transcriptomics applications, SpaGCN is a GCN algorithm that incorporates histological data to identify spatial domains and the associated gene expressions. From the method perspective, while SpaGCN is also an adaptation of UMAP to the graph setting, it differs from our approach in two significant ways: (i) SpaGCN directly targets creating a pre-specified number of clusters in the embedding space. The associated clusters are selected by minimizing the  KL divergence between their distance in low-dimensional space, and their distance in high-dimensional space. The definitions of $q$ and $p$ are also quite different: SpaGCN's $q$ is assigned to a lower value when the embeddings are further from the cluster centers, and $p$ is updated as the twice-normalized square of $q$ at every three epochs. 
GNUMAP, by contrast, is amenable to learning a wider variety of manifolds, as it does not explicitly target the clustering of embeddings, but simply seeks to reconstruct a low-dimensional representation of the nodes that aligns with the original graph.
%there is no notion of distance to cluster centers in the $q$ assignment, and p is defined once as the weighted adjacency matrix of the input data. Also, GNUMAP employs cross-entropy loss, whereas SpaGCN utilizes  Kullback-Leibler.

Overall, GNUMAP is a simple auto-encoder structure that leverages the success of UMAP in the Euclidean setting to perform dimensionality reduction. While this method is not conceptually novel (it is just an autoencoder), as we will show in our experiments section, the choice of the loss, as well as the addition of a whitening step are key to its success. This simple approach can perform well on a variety of tasks, and, as previously argued, is more interpretable and closer to many practioners' requirements in dimensionality reduction.
%minimum distance between nodes in embedding space

\section{Evaluating GNNs' manifold learning abilities}

Having established a natural extension of UMAP to the GNN setting, we propose to benchmark these approaches on a set of toy examples evaluating their ability to correctly learn an underlying manifold structure. While this is a standard test for any dimensionality reduction technique, GNNs have not yet been evaluated in this specific context.

\begin{figure*}[h!]
    \centering
    \begin{tabular}{|p{1.2cm}|c|c|c|c|c|c|}
        \hline
        Dataset & Original Data & SPAGCN & GAE & CCA-SSG & GRACE & GNUMAP \\
        \hline
        Blobs &
        \includegraphics[width= 1.7cm, height=1.7cm, trim={1.4cm 1.25cm 1.2cm 1.35cm},clip]{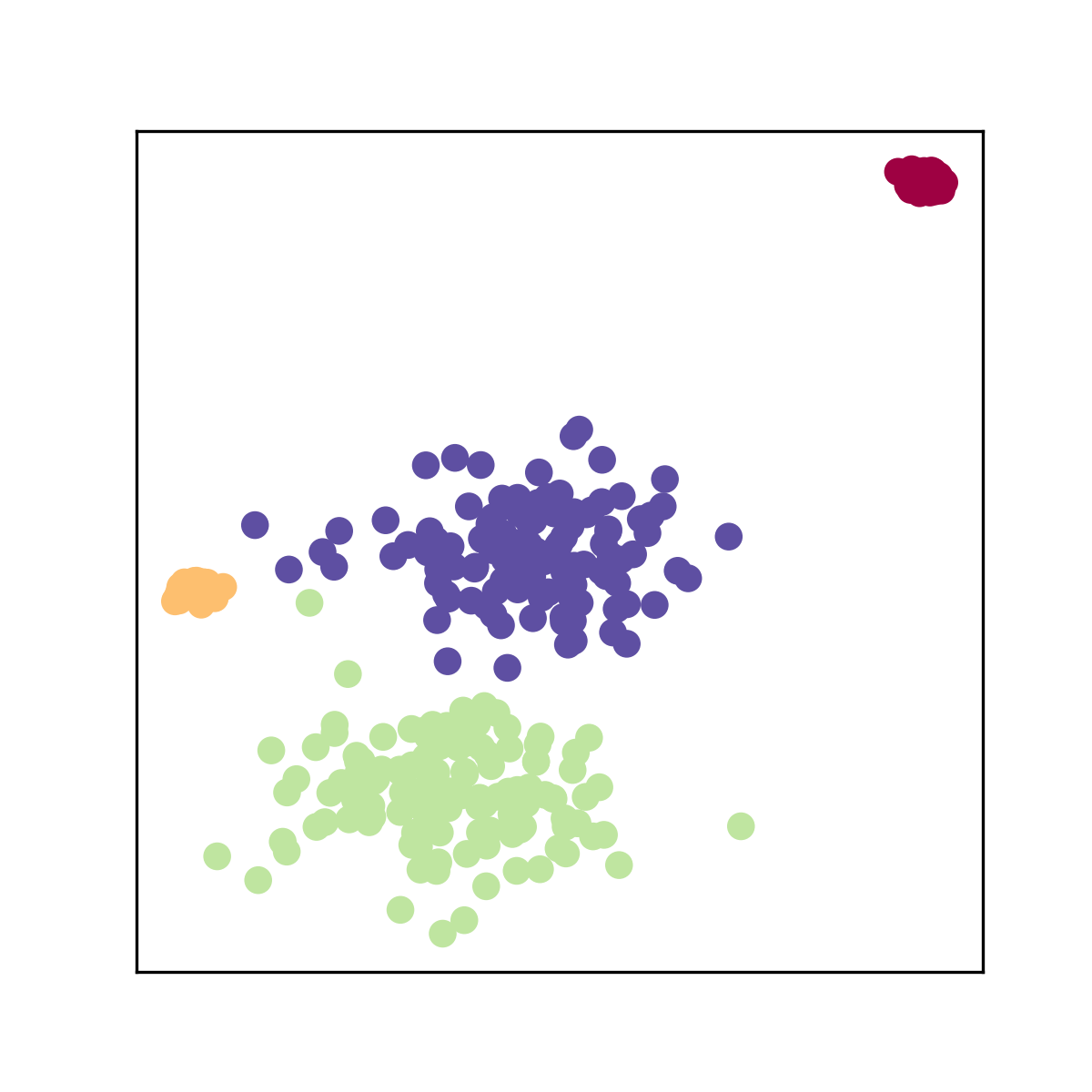} &
        \includegraphics[width= 1.7cm, height=1.7cm, trim={1.4cm 1.25cm 1.2cm 1.35cm},clip]{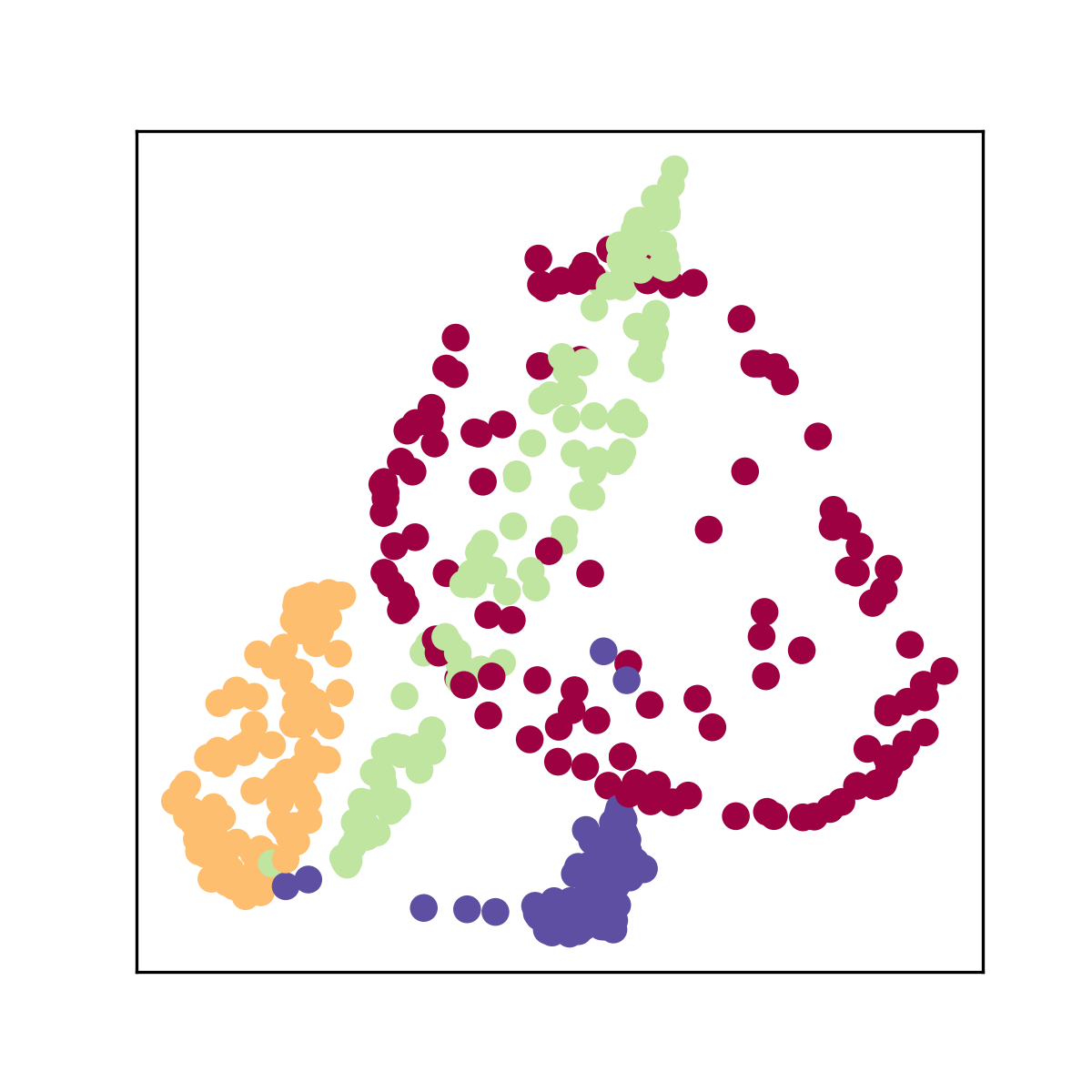} &
        \includegraphics[width= 1.7cm, height=1.7cm, trim={1.4cm 1.25cm 1.2cm 1.35cm},clip]{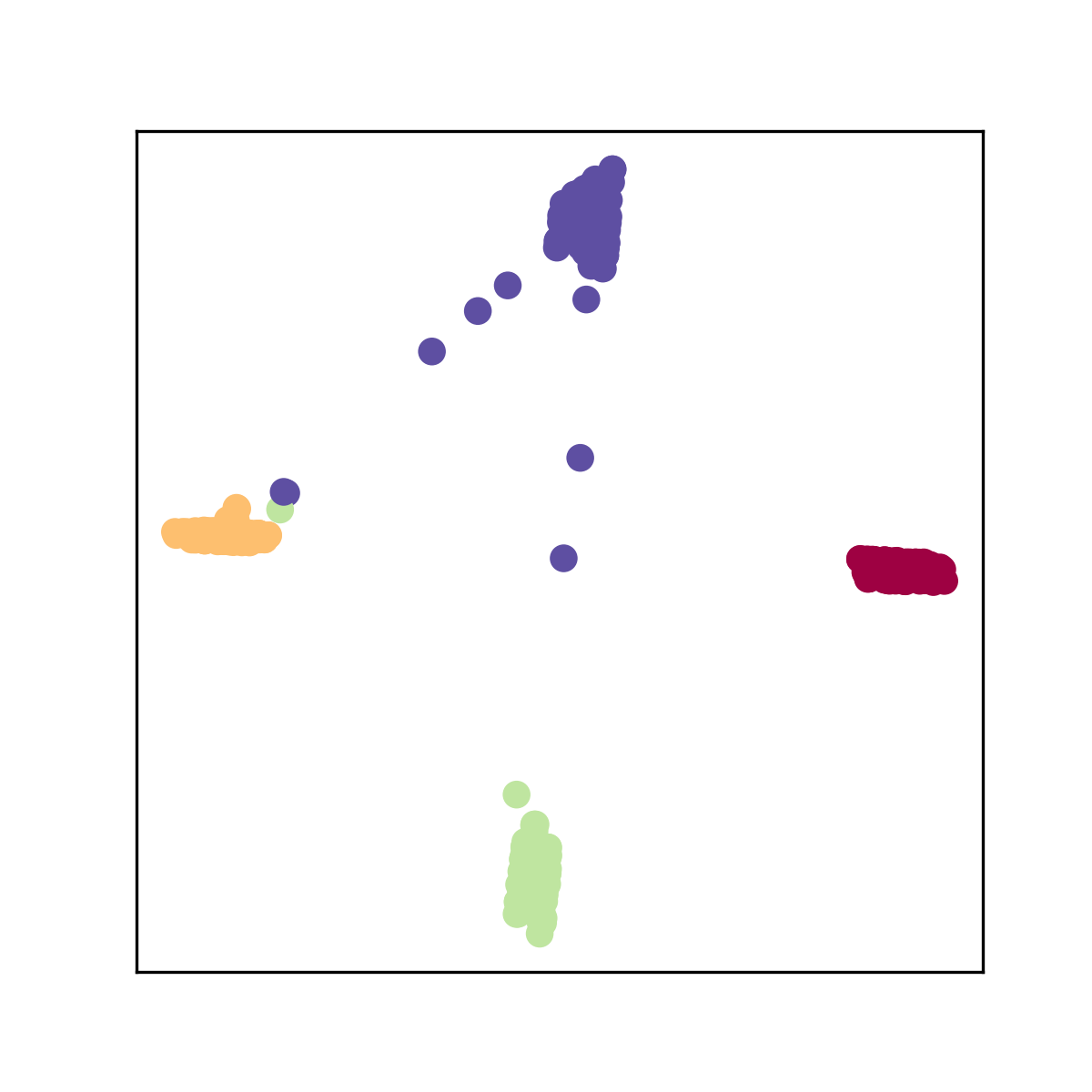} &
        \includegraphics[width= 1.7cm, height=1.7cm, trim={1.4cm 1.25cm 1.2cm 1.35cm},clip]{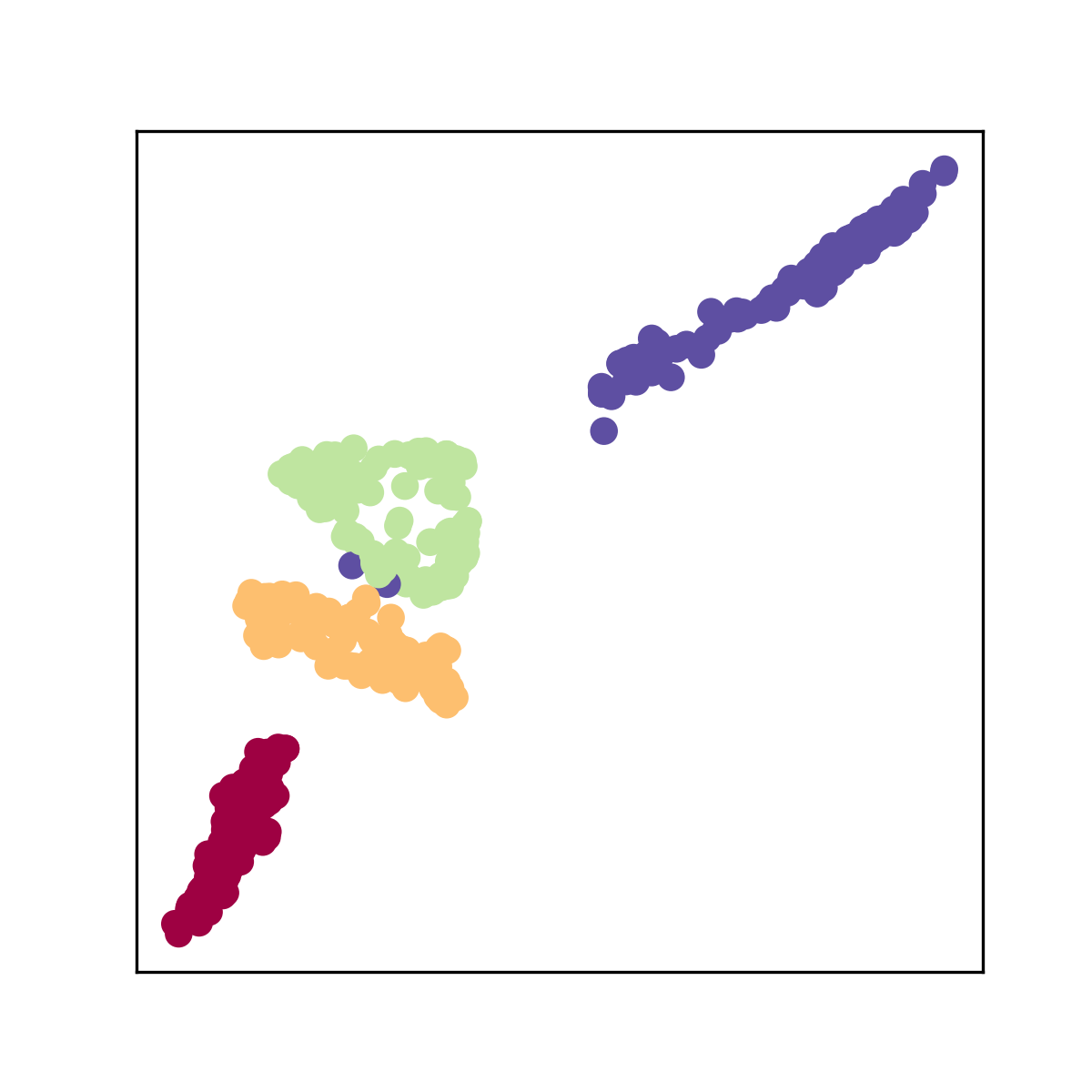} &
        \includegraphics[width= 1.7cm, height=1.7cm, trim={1.4cm 1.25cm 1.2cm 1.35cm},clip]{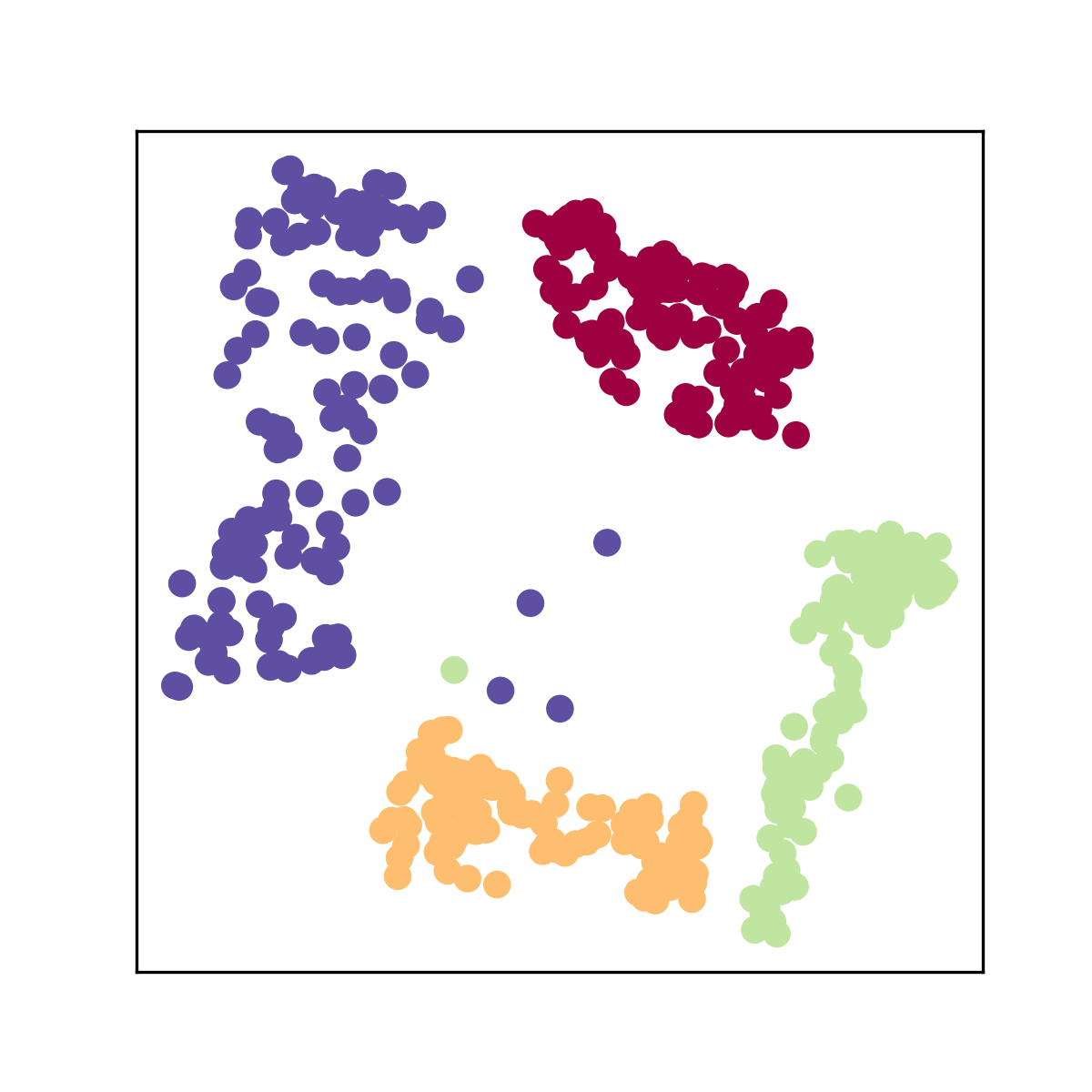} &
        \includegraphics[width= 1.7cm, height=1.7cm, trim={1.4cm 1.25cm 1.2cm 1.35cm},clip]{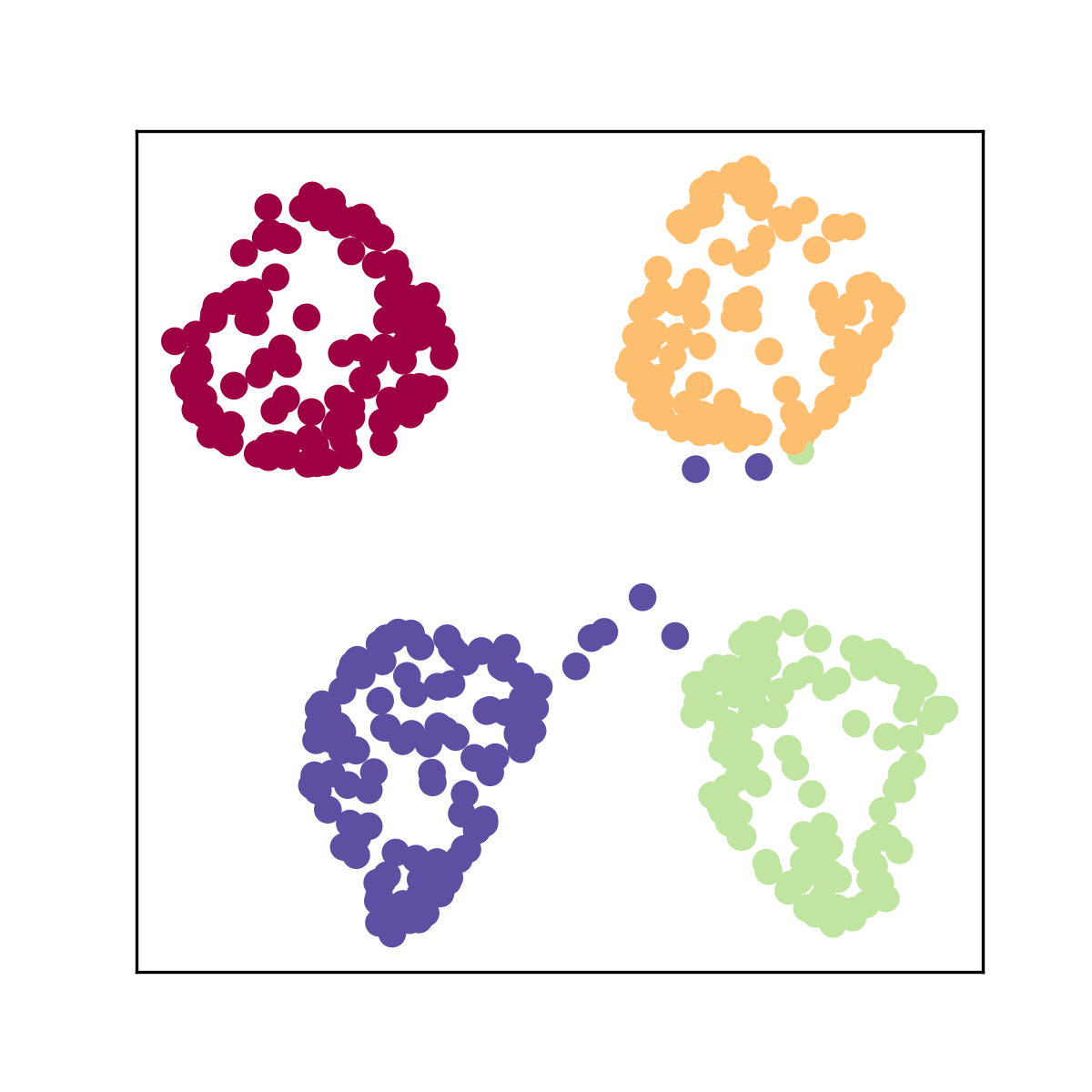} \\
        \hline
        Swissroll &
        \includegraphics[width=1.7cm, height=1.7cm, trim={1.4cm 1.25cm 1.2cm 1.35cm},clip]{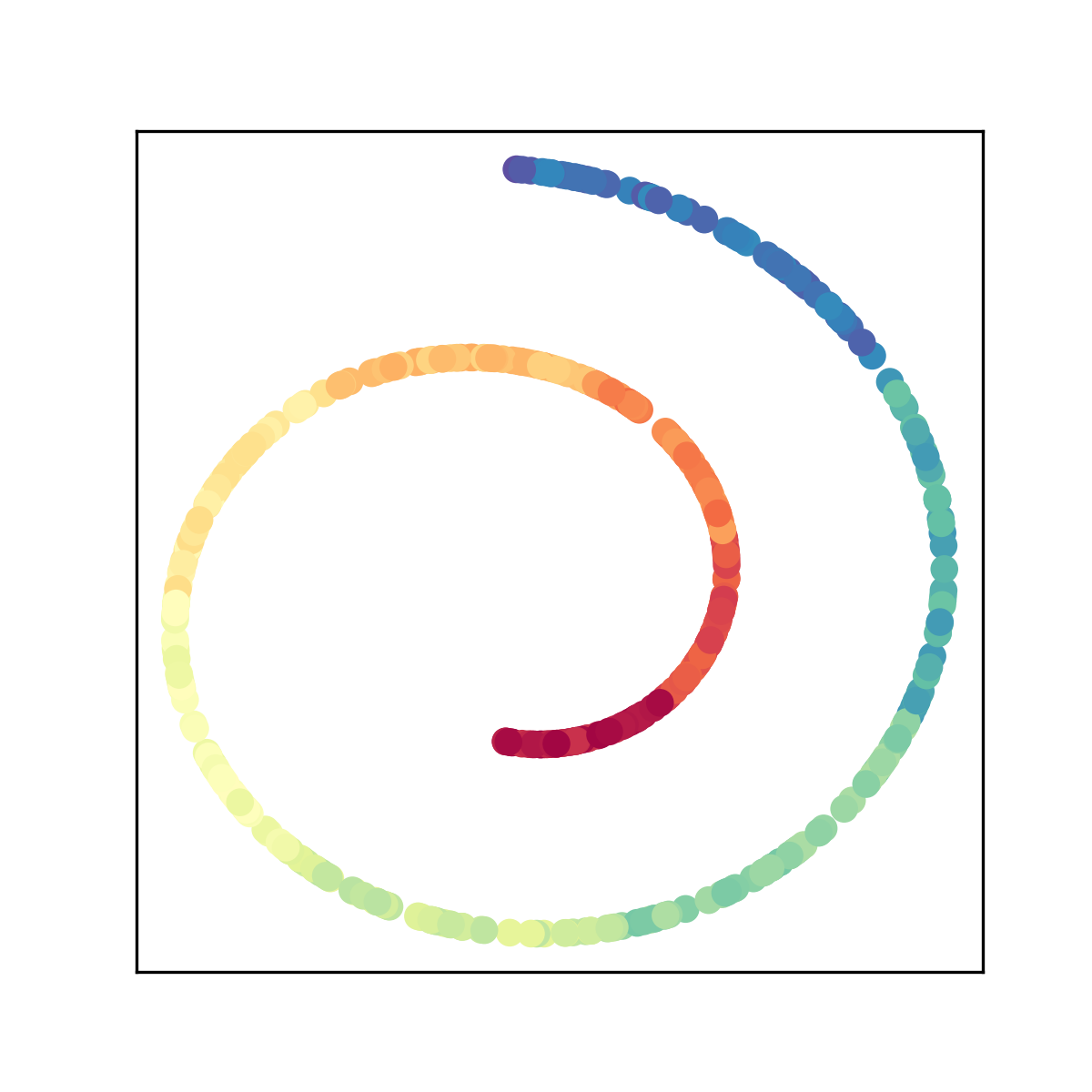} &
        \includegraphics[width=1.7cm, height=1.7cm, trim={1.4cm 1.25cm 1.2cm 1.35cm},clip]{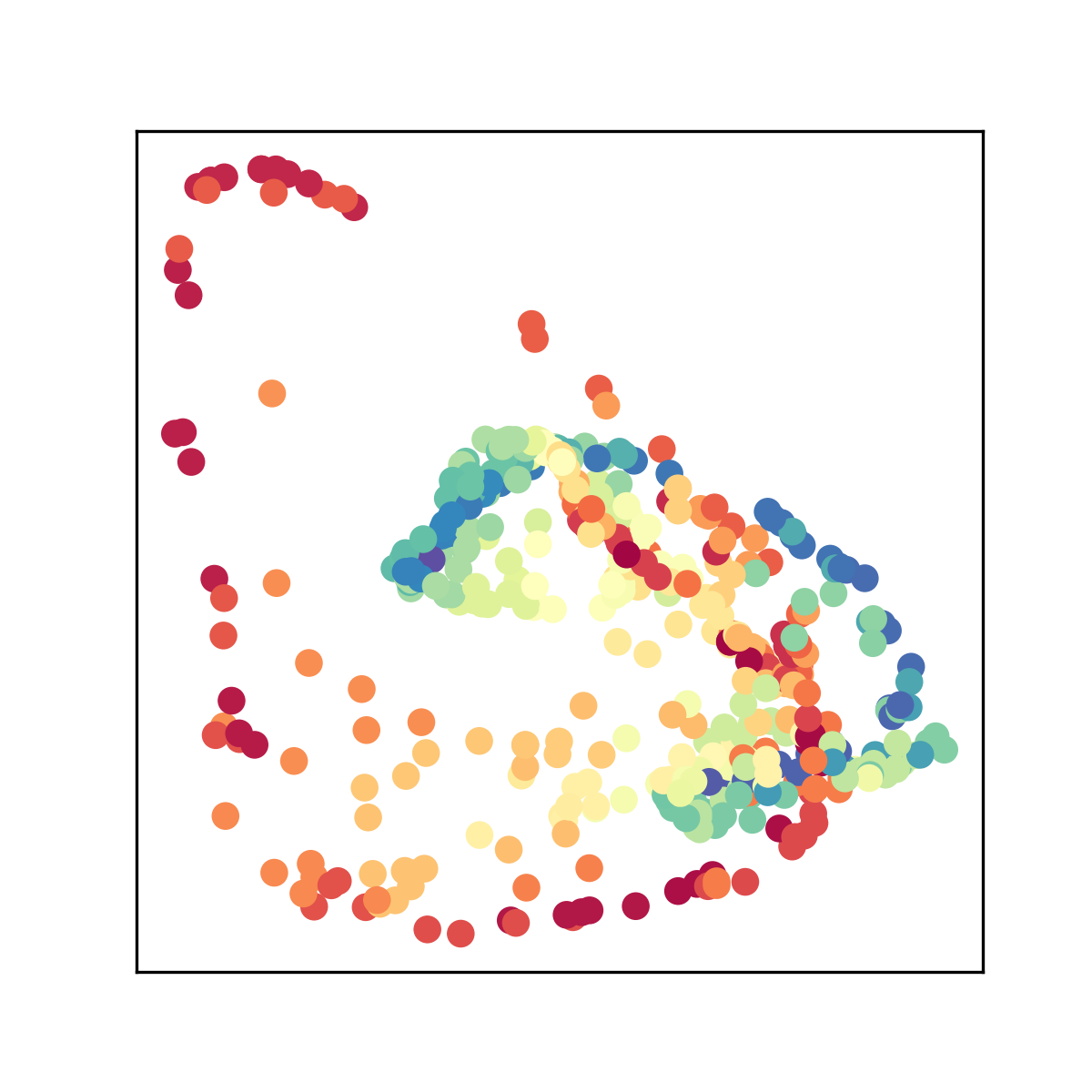} &
        \includegraphics[width=1.7cm, height=1.7cm, trim={1.4cm 1.25cm 1.2cm 1.35cm},clip]{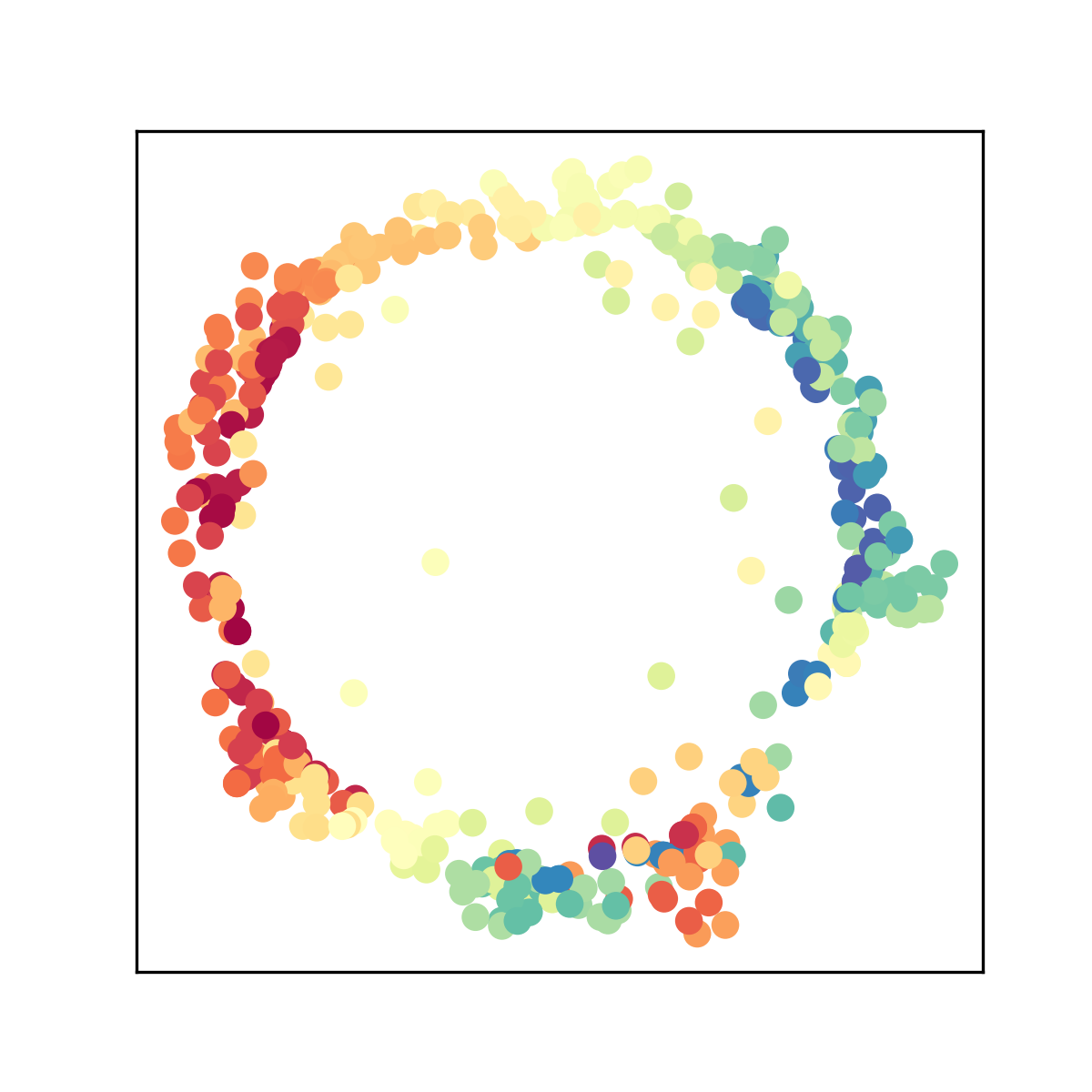} &
        \includegraphics[width=1.7cm, height=1.7cm, trim={1.4cm 1.25cm 1.2cm 1.35cm},clip]{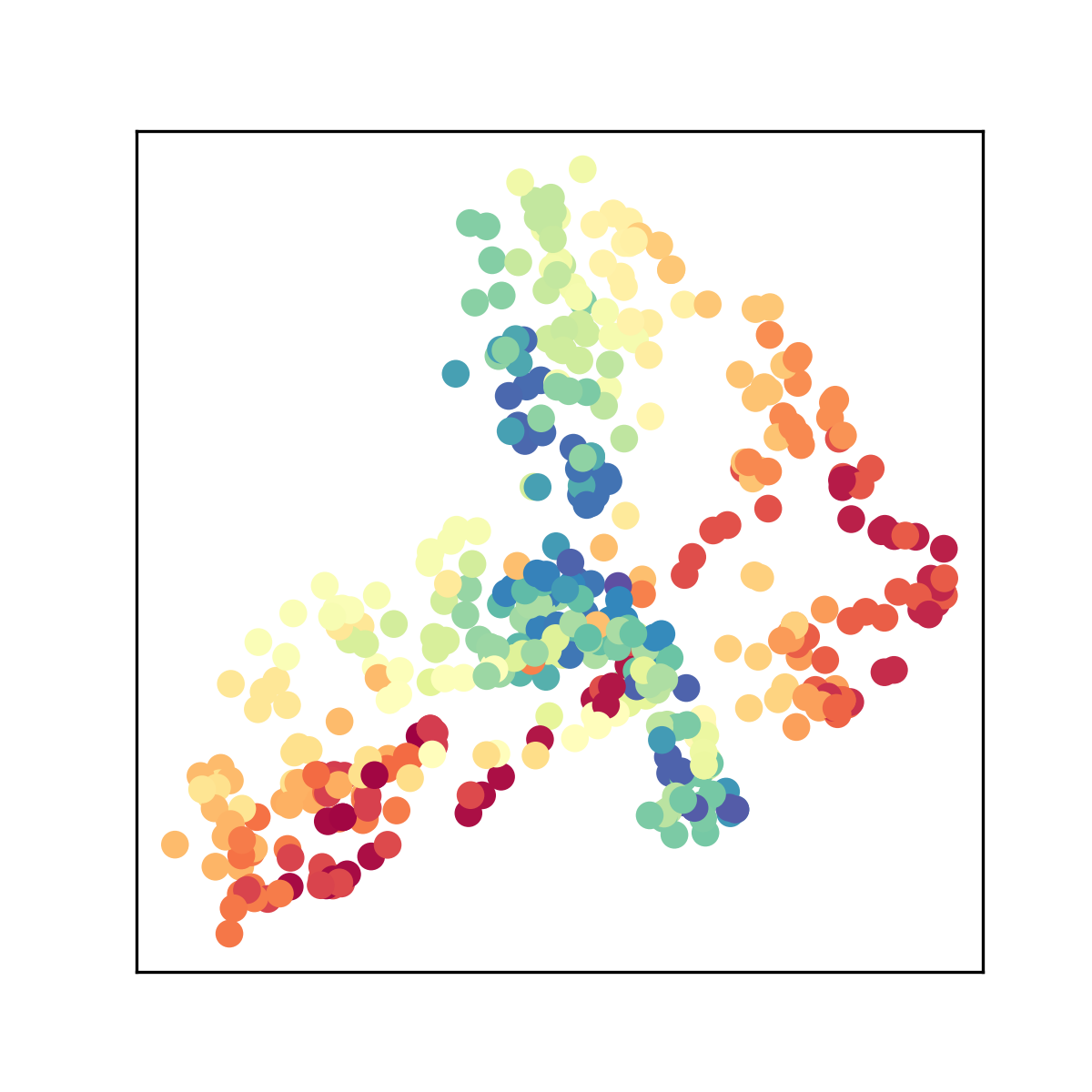} &
        \includegraphics[width=1.7cm, height=1.7cm, trim={1.4cm 1.25cm 1.2cm 1.35cm},clip]{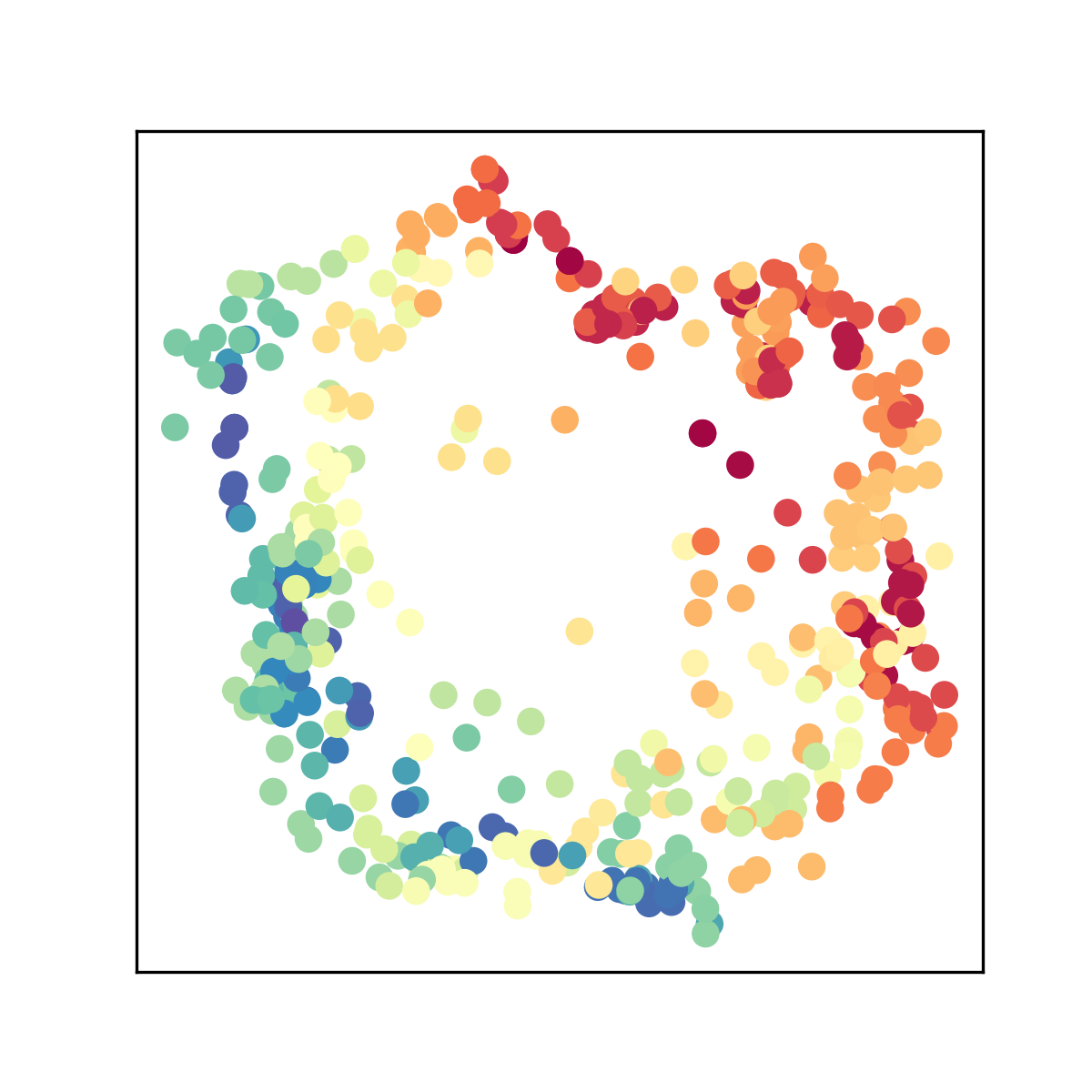} &
        \includegraphics[width=1.7cm, height=1.7cm, trim={1.4cm 1.25cm 1.2cm 1.35cm},clip]{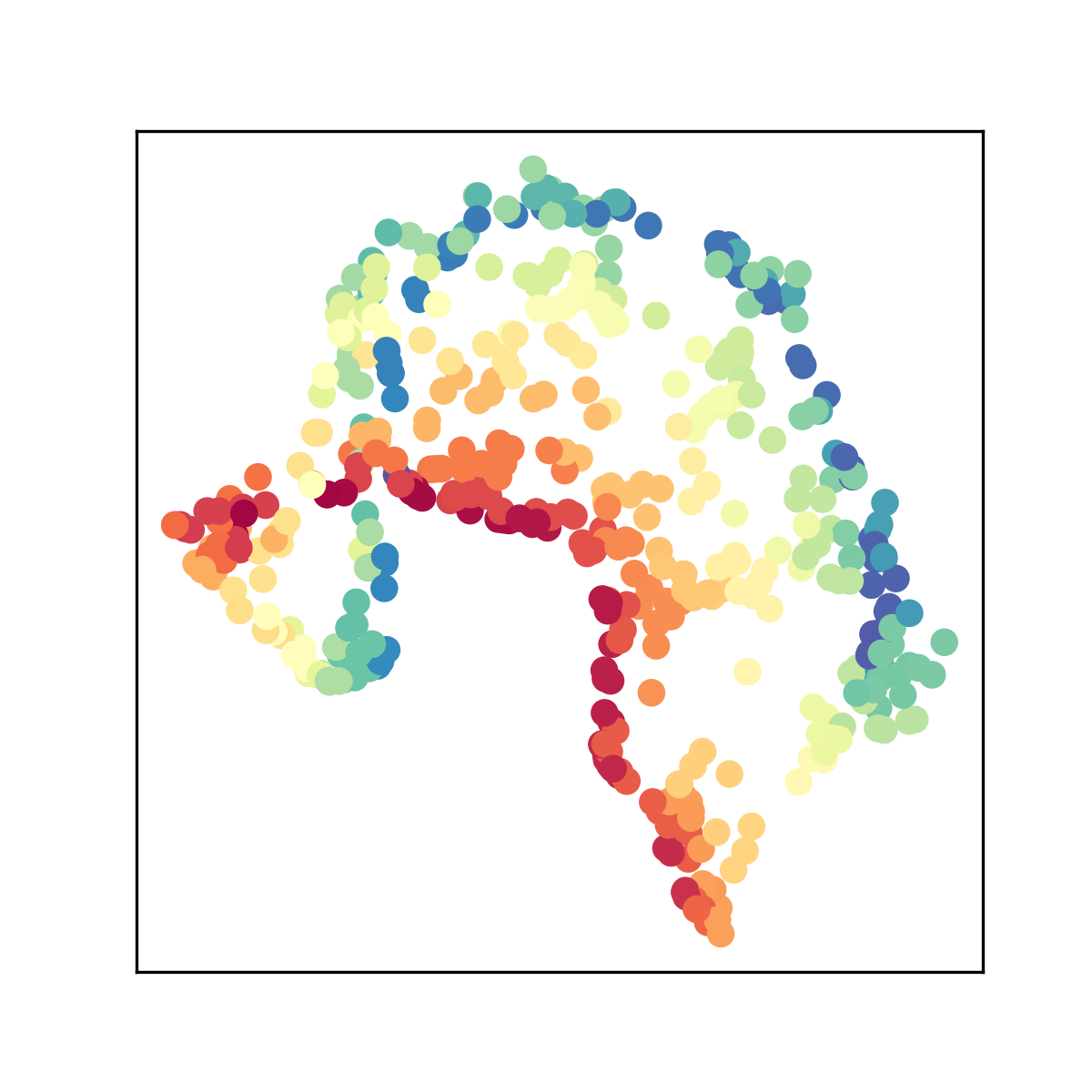} \\
        \hline
        Circles &
        \includegraphics[width=1.7cm, height=1.7cm, trim={1.4cm 1.25cm 1.2cm 1.35cm},clip]{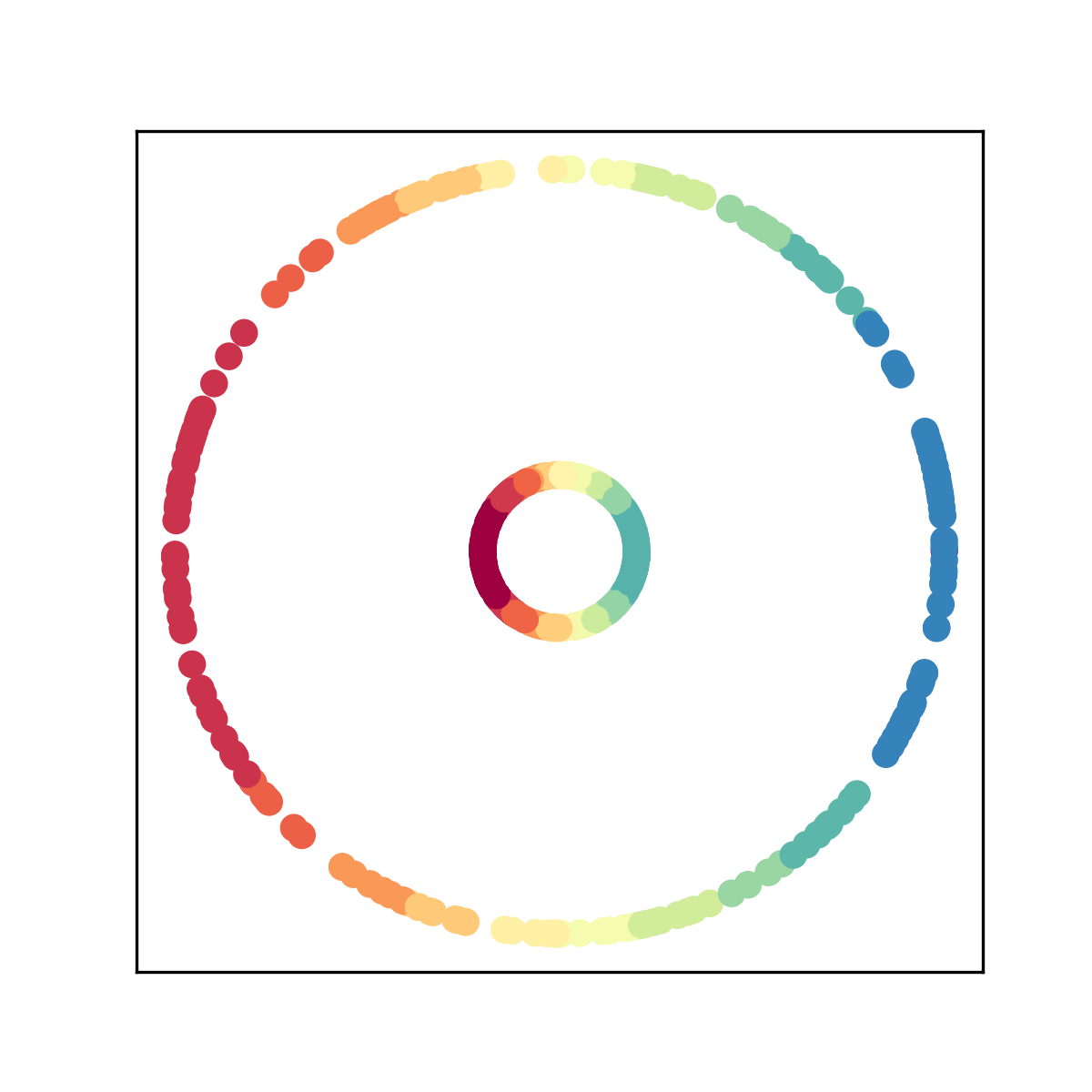} &
        \includegraphics[width=1.7cm, height=1.7cm, trim={1.4cm 1.25cm 1.2cm 1.35cm},clip]{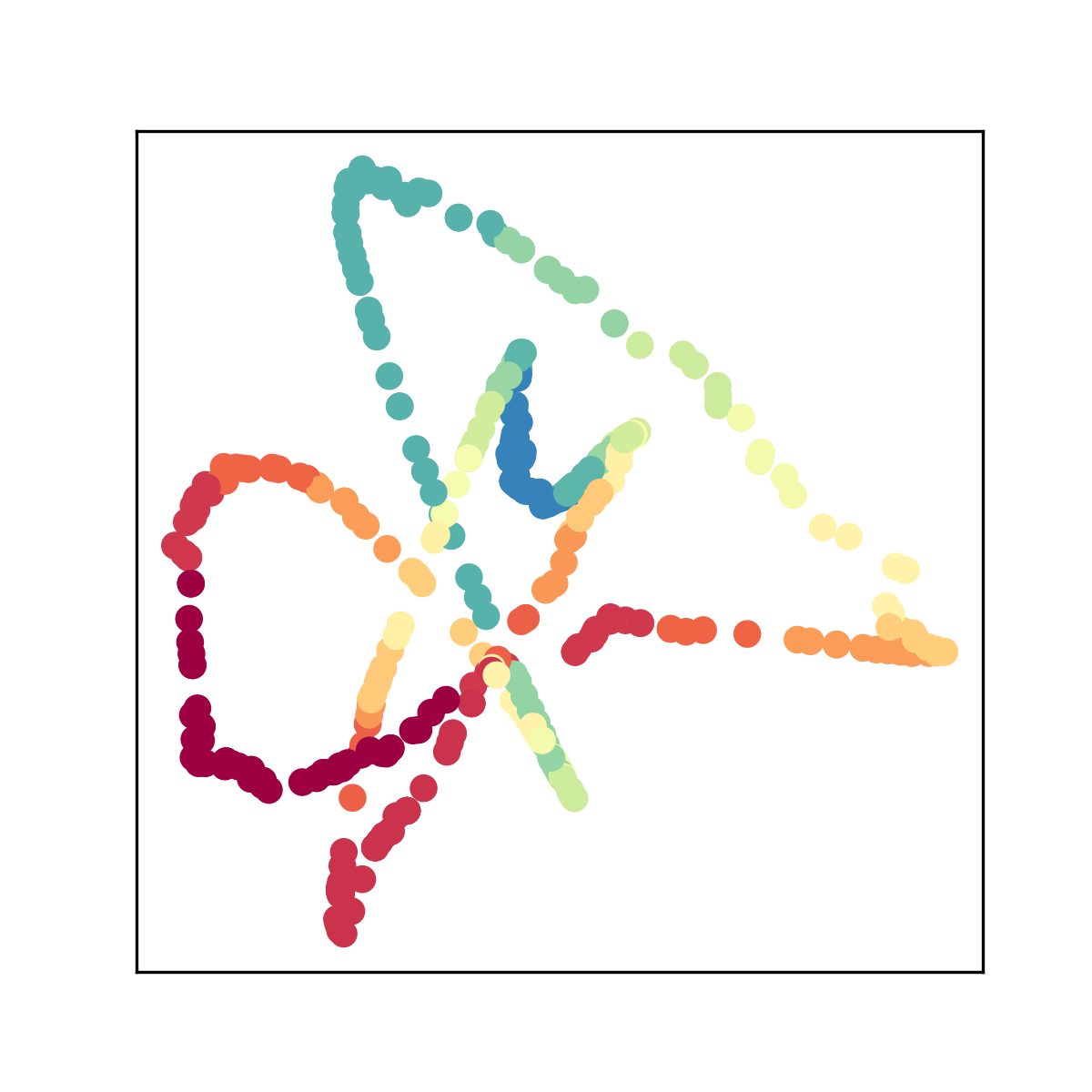} &
        \includegraphics[width=1.7cm, height=1.7cm, trim={1.4cm 1.25cm 1.2cm 1.35cm},clip]{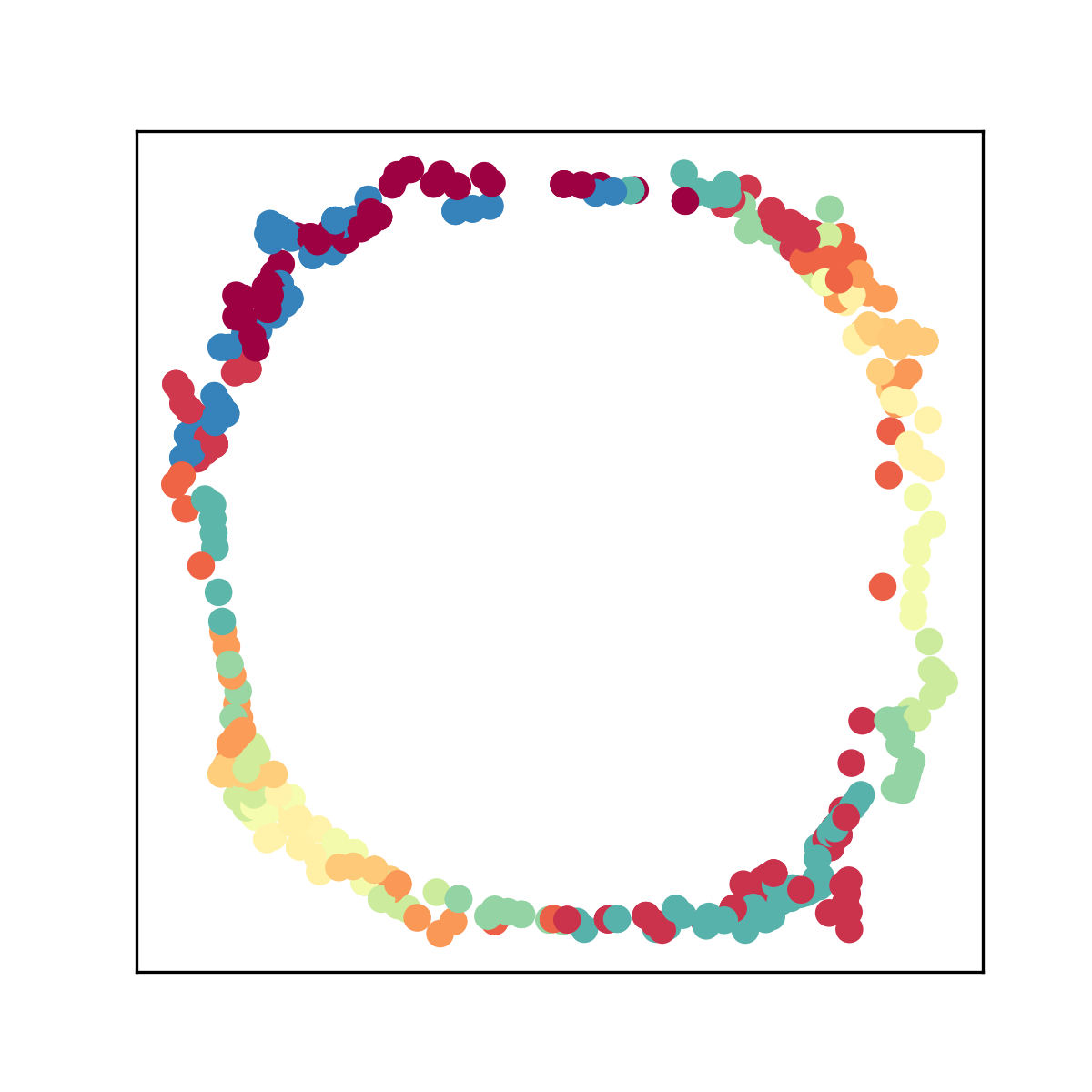} &
        \includegraphics[width=1.7cm, height=1.7cm, trim={1.4cm 1.25cm 1.2cm 1.35cm},clip]{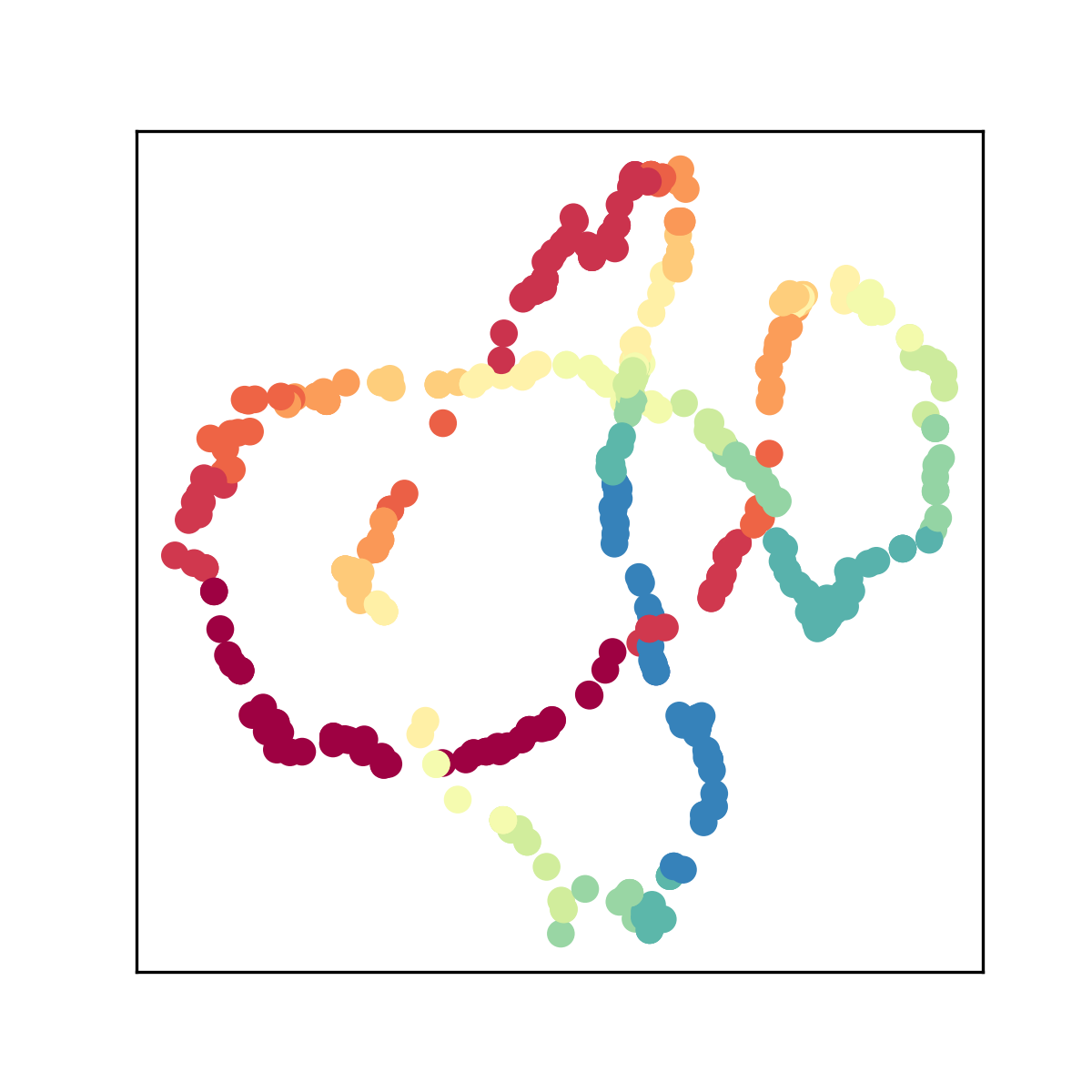} &
        \includegraphics[width=1.7cm, height=1.7cm, trim={1.4cm 1.25cm 1.2cm 1.35cm},clip]{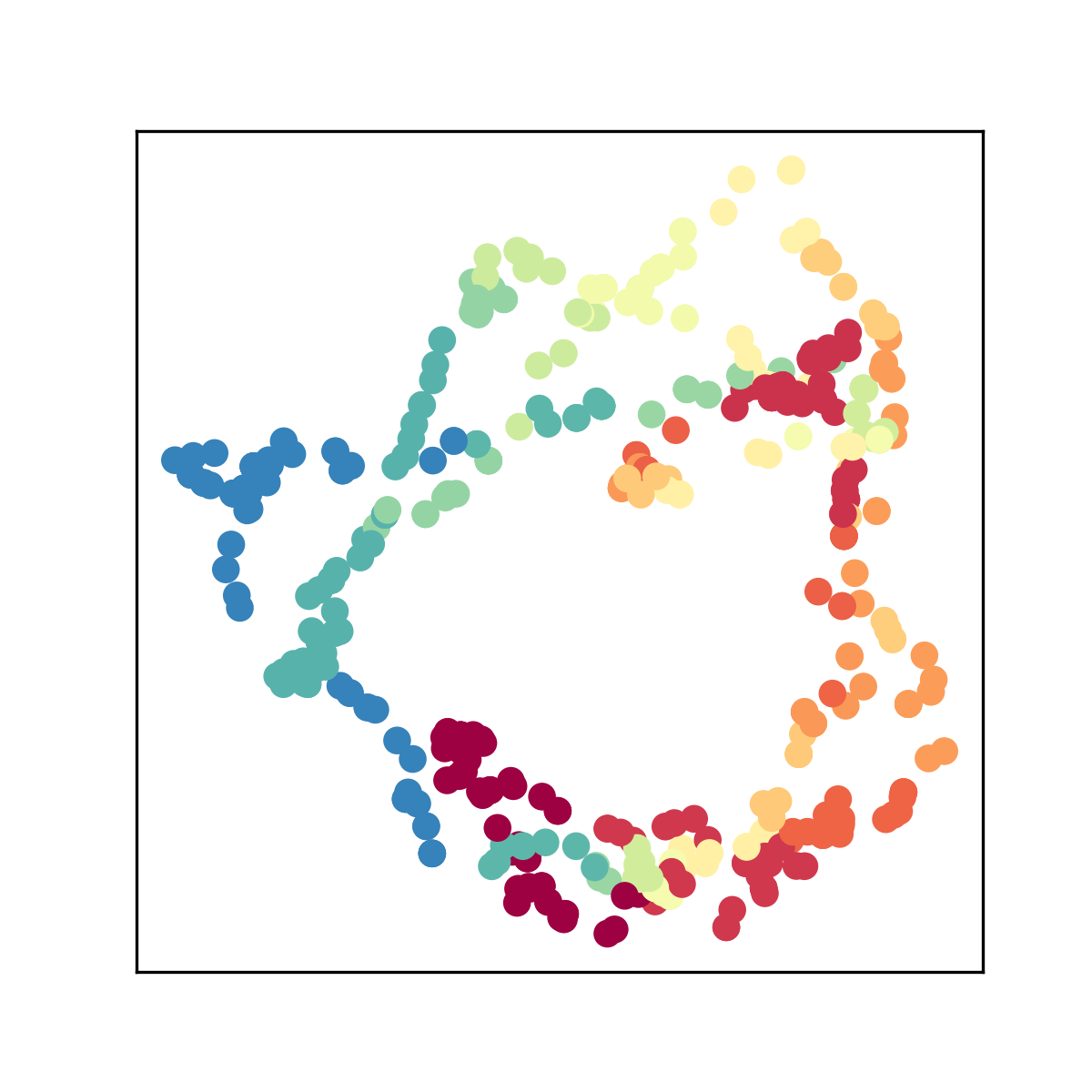} &
        \includegraphics[width=1.7cm, height=1.7cm, trim={1.4cm 1.25cm 1.2cm 1.35cm},clip]{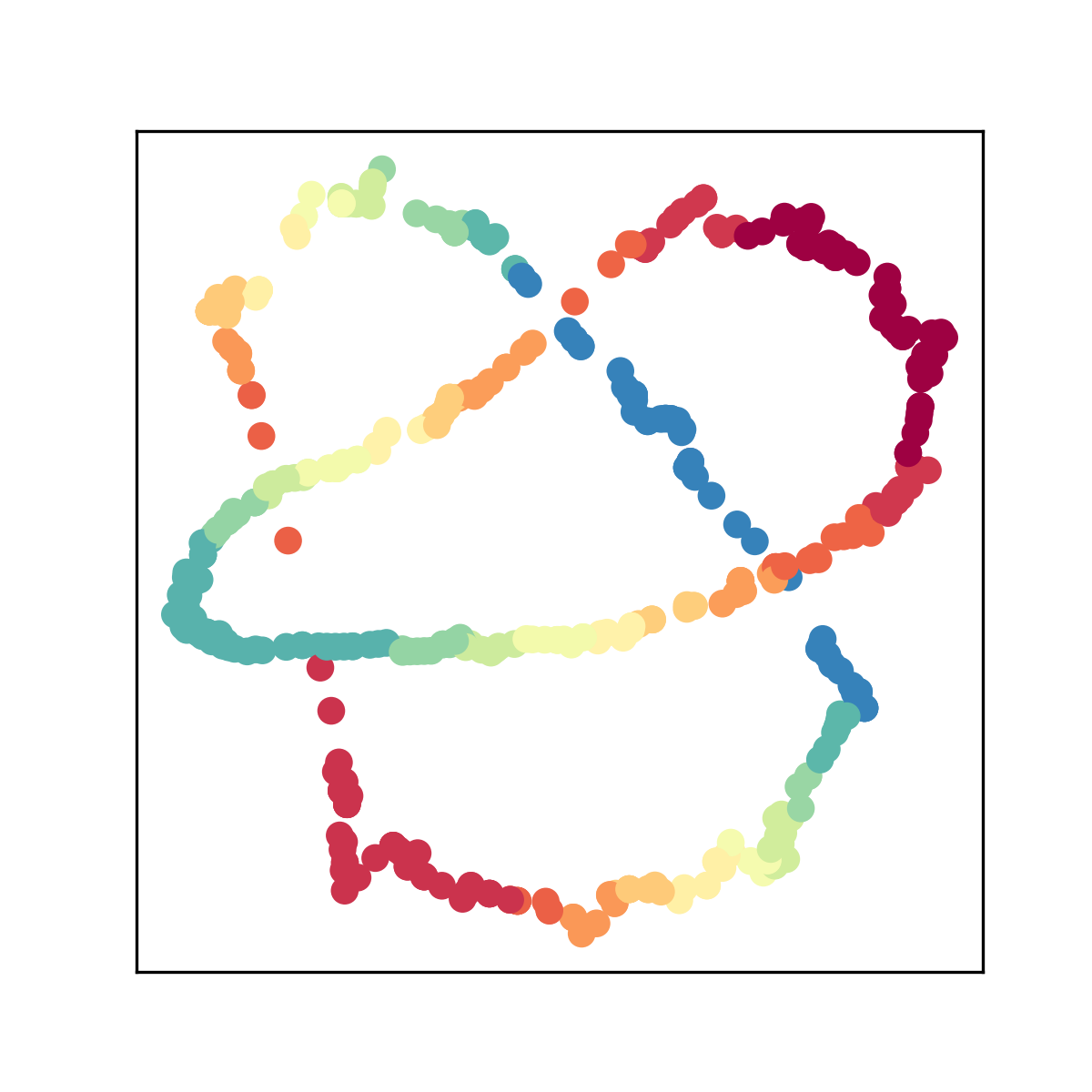} \\
        \hline   
        Moons &
        \includegraphics[width=1.7cm, height=1.7cm, trim={1.4cm 1.25cm 1.2cm 1.35cm},clip]{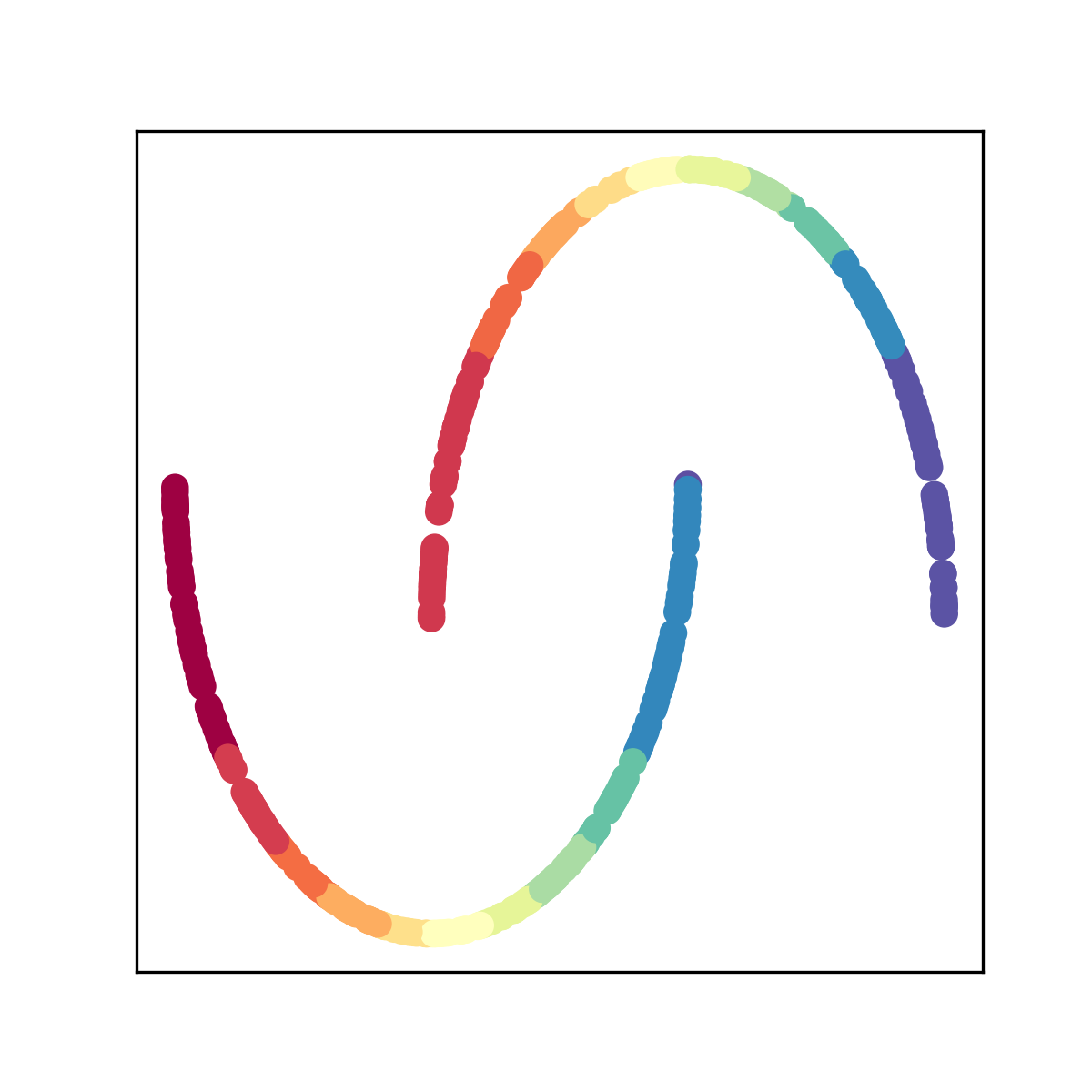} &
        \includegraphics[width=1.7cm, height=1.7cm, trim={1.4cm 1.25cm 1.2cm 1.35cm},clip]{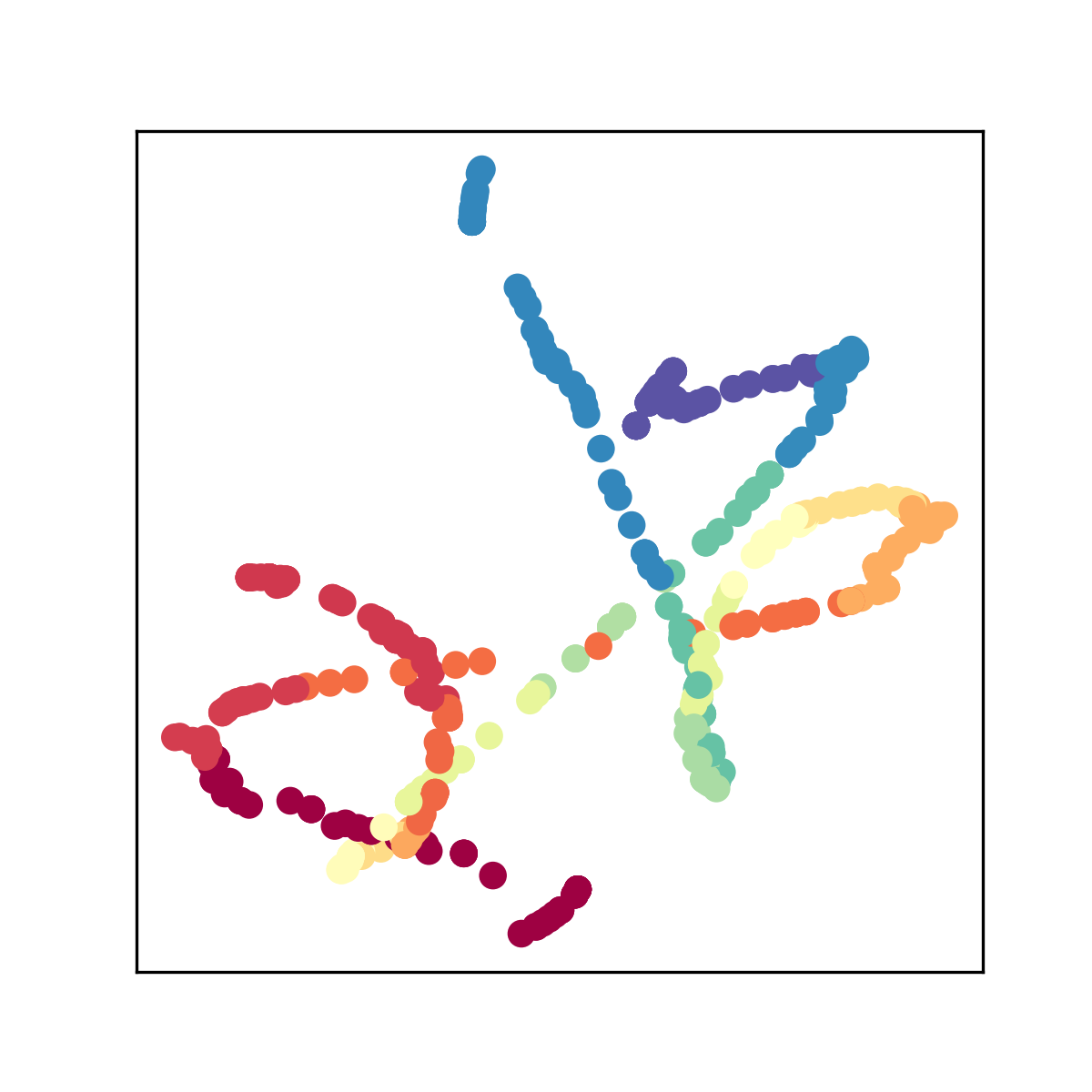} &
        \includegraphics[width=1.7cm, height=1.7cm, trim={1.4cm 1.25cm 1.2cm 1.35cm},clip]{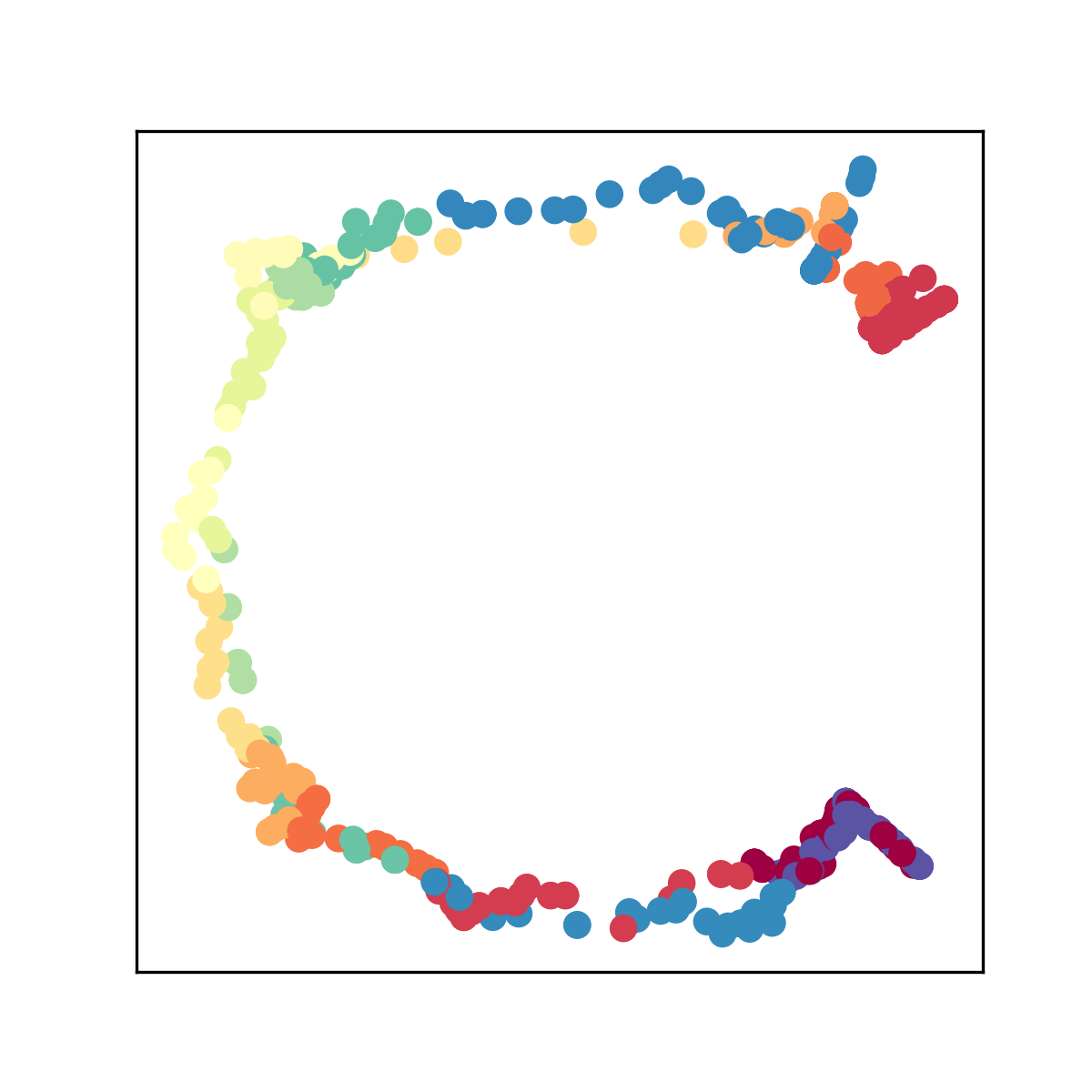} &
        \includegraphics[width=1.7cm, height=1.7cm, trim={1.4cm 1.25cm 1.2cm 1.35cm},clip]{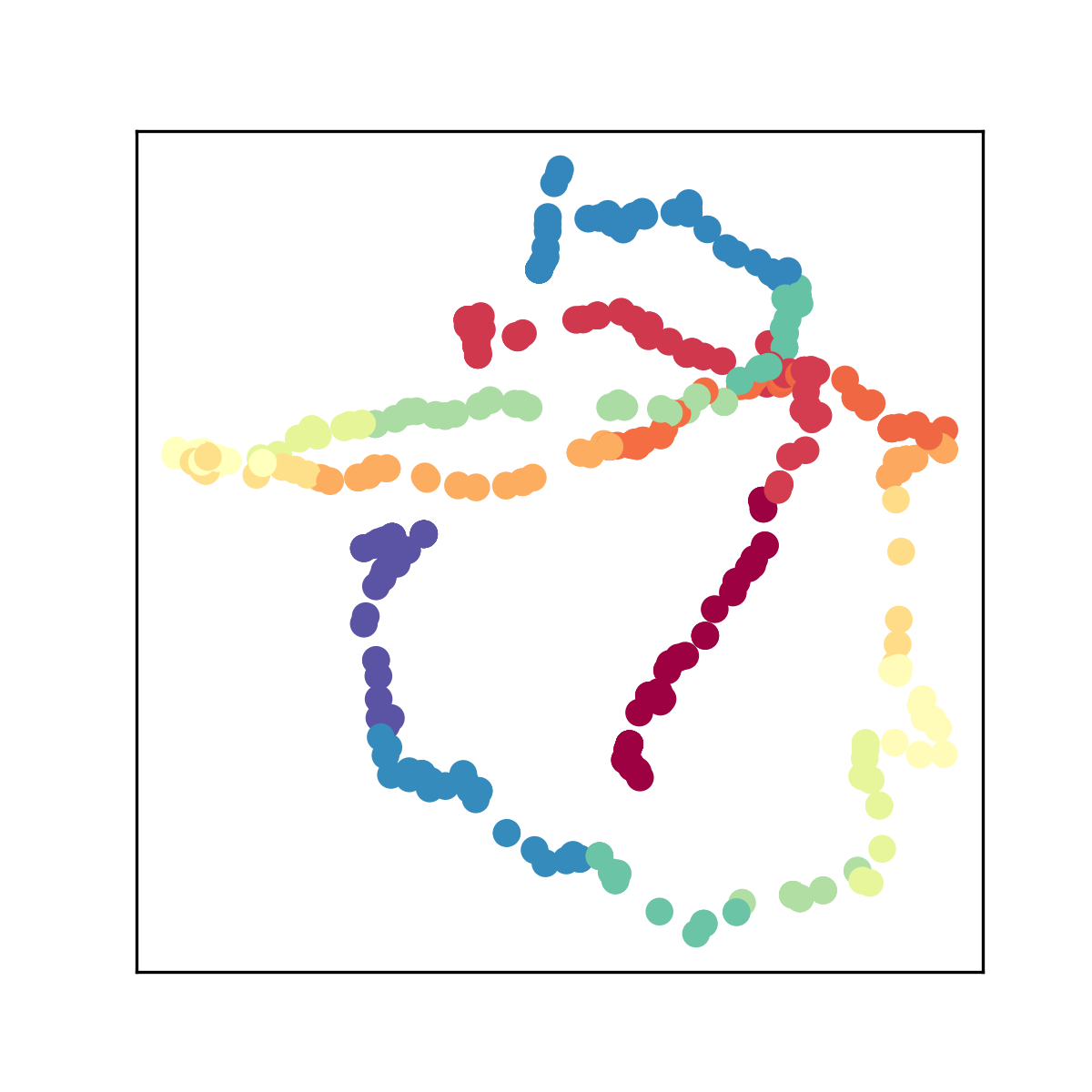} &
        \includegraphics[width=1.7cm, height=1.7cm, trim={1.4cm 1.25cm 1.2cm 1.35cm},clip]{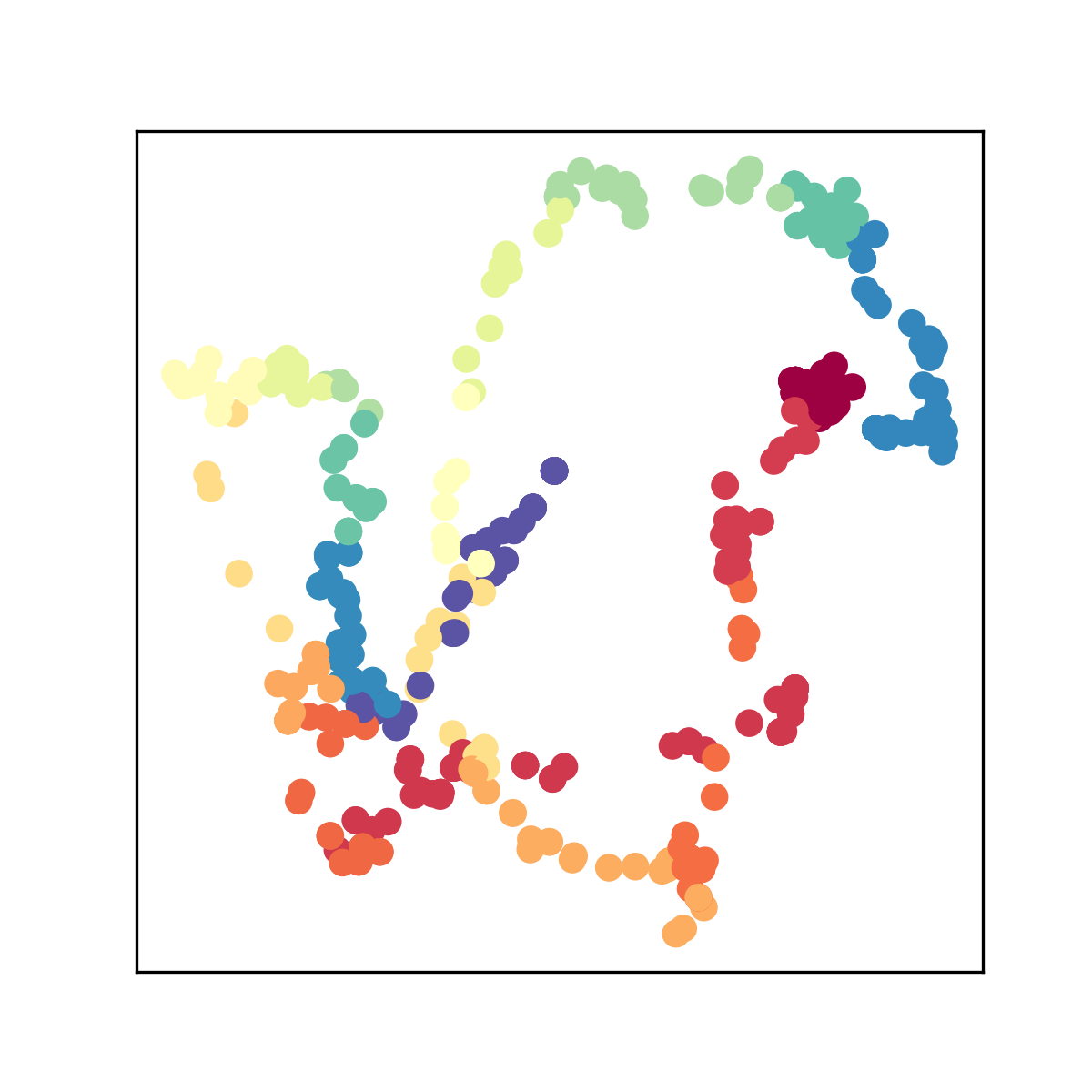} &
        \includegraphics[width=1.7cm, height=1.7cm, trim={1.4cm 1.25cm 1.2cm 1.35cm},clip]{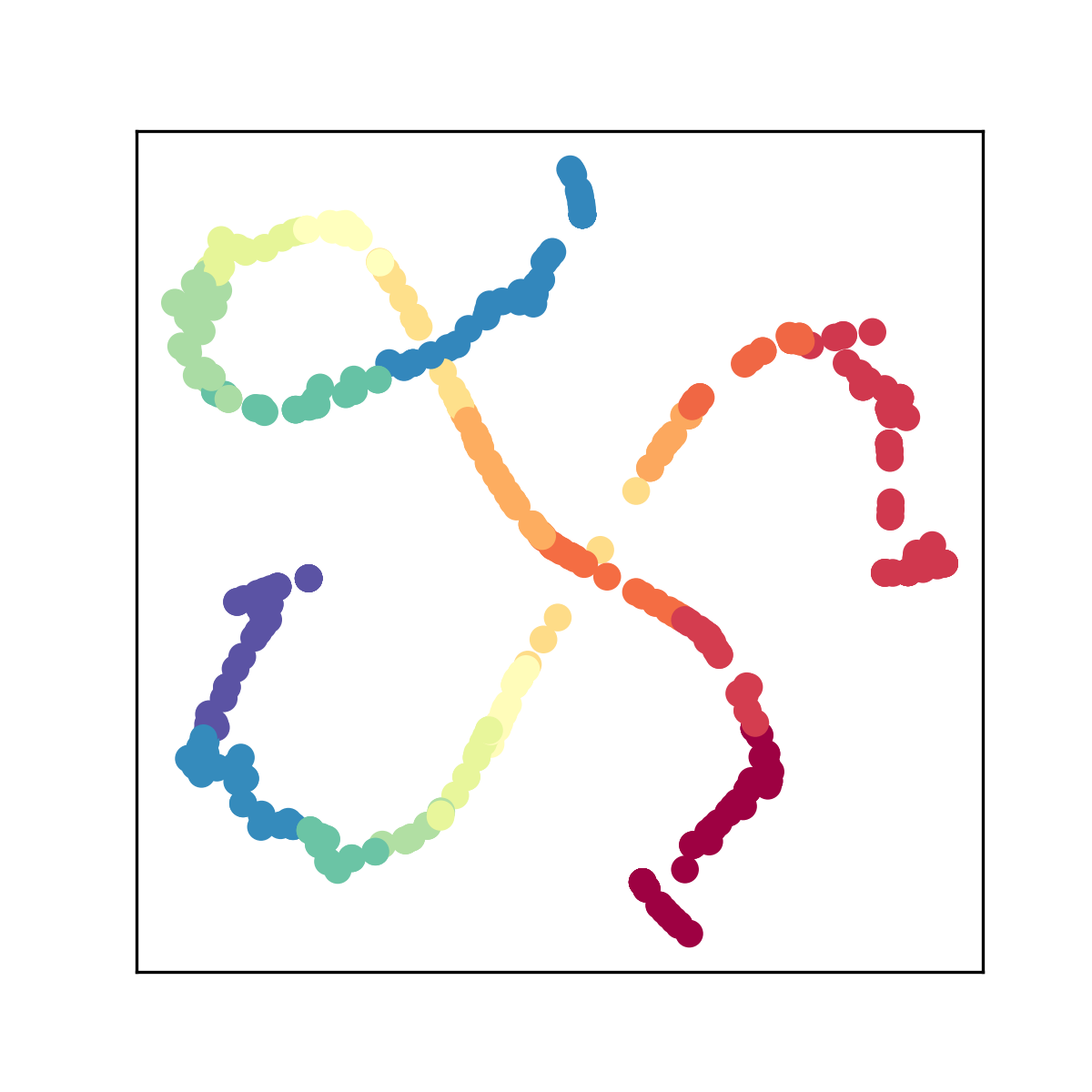} \\
        \hline   
    \end{tabular}
    \caption{Node representation learning for 4 synthetic datasets : Blobs, Swissroll, Circles and Moons. Each image represents a different method of visualization.}
    \label{fig:synthetic_visual}
\end{figure*}

%\subsection{Synthetic Geometric Datasets}
We first evaluate GNUMAP performance on four synthetic geometric graph datasets considered as benchmarks in the realm of unsupervised dimensionality reduction methods: the Blobs, Swissroll, Moons, and Circles datasets (see Figure~\ref{fig:synthetic_visual}). The datasets were generated with ground-truth low-dimensional coordinates and cluster labels, which allows us to calculate our proposed metrics and also to inspect if our metrics are coherent with the embedding visualization. The four datasets were generated with 500 nodes, each connected to its 20 nearest neighbors. For simplicity, the experiments in this work have assigned a value of 1 to positive edges and 0 to negative edges. However, the GNUMAP algorithm is applicable to any input graph preprocessed such that weaker edges are mapped closer to 0 and stronger edges closer to 1. The features were instantiated as the embeddings of a 10-component Laplacian eigenmap decomposition of the induced 20-nearest graph. The latter is indeed an established procedure to instantiate features on a graph \cite{hu2021spagcn}. We compare GNUMAP with state-of-the-art GNN embedding methods tailored for dimensionality reduction: DGI \cite{dgi}, BGRL \cite{bgrl}, CCA-SSG \cite{ccassg}, GAE \cite{vgae}, VGAE \cite{vgae} and SpaGCN \cite{hu2021spagcn}. We also compare with well-established dimensionality reduction methods PCA, t-SNE \cite{tsne}, Isomap \cite{isomap}, Laplacian Eigenmap \cite{LaplacianEigenmap}, UMAP \cite{umap}, and DenseMAP \cite{densmap}. Note all methods mentioned above are Euclidean methods, which enable a consistent assessment with respect to our proposed metrics. To evaluate embedding quality, we introduce metrics including but not limited to: agreement in local geometry, Spearman correlation, and classification accuracy. We refer the reader to Appendix~\ref{app:metrics} for a more detailed description of these metrics.

We implement a standard 2-layer GCN architecture for all GNN-based methods. Note that SpaGCN, CCA-SSG, GRACE, and BGRL are hyperparameter-heavy methods. SpaGCN requires the louvain resolution and number of neighbours for cluster initialization, as well as the penalty coefficient for embedding distance from cluster centers. Similarly, GRACE, BGRL and CCA-SSG require the specification of the regularization $\lambda$, an edge drop rate, as well as a feature mask rate. Autoencoder methods such as GNUMAP and GAE, on the other hand, are completely parameter free.

Since it is unclear which set of hyperparameters would be optimal for each dataset when deploying contrastive-learning techniques, we ran a full search for all possible hyperparameter combinations.  {\it Note that this provides the most favorable comparison of these methods to GNUMAP, but deviates from the original unsupervised learning setting, since we are choosing the hyperparameters of the methods based on the labels --- therefore making the methods more supervised. In real-world experiments, however, we would not benefit from such supervision. Despite this selection procedure, as we will show, GNUMAP still manages to outperform most methods.}

We visualize some of the embedding from the best hyperparameter combination in Figure \ref{fig:synthetic_visual}. In addition to GNUMAP, we visualized GAE for comparisons with explicitly learned distance between low-dimensional embeddings, CCA-SSG and GRACE for comparisons with contrastive learning methods. We also show results from related method SpaGCN. For a complete illustration including DGI, BGRL, VGAE, as well as PCA, t-SNE, Isomap, Laplacian Eigenmap, UMAP, and DenseMAP, see Figure~\ref{fig:synthetic_appendix} in the appendix. Note the dimensionality reduction methods were not fitted on spectral features like the GNN-based models.

\renewcommand{\arraystretch}{2}
\renewcommand{\arraystretch}{1}

Furthermore, we compute proposed metrics over 100 experiments and summarize the results in Table~\ref{table:synthetic_metric}.

% Furthermore, we compute proposed metrics over 100 swissroll experiments and summarize the results in Table \ref{table:synthetic_metric}. Metrics for other synthetic datasets can be found in the Appendix.

\begin{table*}
    \centering
        \caption{Evaluated mean and standard deviation of proposed metrics for synthetic datasets blobs, circles, moons, and swissroll over 100 experiments with GNN-based methods. The upward arrows next to metric name denote a better performance with higher metrics, whereas the downward arrows denote better performance with lower metrics. The best result is bolded, and the next best result is underlined.}
        \vspace{1em}
        \begin{tabular}{|p{1.4cm}|p{1.6cm}|p{2cm}|p{3cm}|p{2cm}|p{1.9cm}|p{1.9cm}|}
\hline
\rowcolor[HTML]{C0C0C0} 
Dataset   & Model   & Classification Accuracy ↑ & Calinski Harabasz Score ↑    & Davies Bouldin Score ↓ & Spearman Correlation w/ Original Graph ↑ & Overlap \% of 50 Neighbours ↑ \\ \hline
Blobs     & DGI     & 0.65 ± 0.06               & 95.17 ± 90.13                & 5.53 ± 4.73            & 0.22 ± 0.09                              & 0.52 ± 0.06                   \\
          & BGRL    & 0.70 ± 0.07               & 79.26 ± 57.93                & 5.28 ± 7.56            & 0.23 ± 0.09                              & 0.59 ± 0.05                   \\
          & CCA-SSG & 0.90 ± 0.08               & 1139.31 ± 1258.86            & 1.91 ± 3.87            & 0.52 ± 0.18                              & \underline{0.76 ± 0.06}             \\
          & SPAGCN  & 0.74 ± 0.07               & 61.07 ± 37.80                & 5.89 ± 7.41            & 0.22 ± 0.08                              & 0.70 ± 0.04                   \\
          & GAE     & \underline{0.93 ± 0.10}         & \textbf{10456.43 ± 12149.96} & \underline{1.14 ± 2.25}      & \underline{0.61 ± 0.13}                        & 0.58 ± 0.07                   \\
          & VGAE    & 0.91 ± 0.10               & \underline{5523.10 ± 6633.58}      & 1.33 ± 2.12            & \underline{0.61 ± 0.13}                        & 0.55 ± 0.09                   \\
          & GRACE   & 0.84 ± 0.12               & 561.59 ± 631.46              & 2.59 ± 3.32            & 0.48 ± 0.18                              & 0.73 ± 0.06                   \\
          & GNUMAP  & \textbf{0.99 ± 0.02}      & 1086.79 ± 173.32             & \textbf{0.58 ± 0.11}   & \textbf{0.69 ± 0.10}                     & \textbf{0.97 ± 0.01}          \\ \hline
Circles   & DGI     & 0.47 ± 0.06               & 34.19 ± 25.52                & 12.37 ± 7.42           & 0.23 ± 0.10                              & 0.72 ± 0.06                   \\
          & BGRL    & 0.49 ± 0.05               & 29.51 ± 16.63                & 10.90 ± 6.39           & 0.20 ± 0.07                              & 0.76 ± 0.05                   \\
          & CCA-SSG & 0.53 ± 0.06               & \textbf{72.81 ± 92.22}       & \textbf{10.69 ± 6.74}  & \underline{0.40 ± 0.21}                        & \underline{0.88 ± 0.04}             \\
          & SPAGCN  & \underline{0.54 ± 0.05}         & 31.00 ± 17.62                & 11.60 ± 8.73           & 0.21 ± 0.06                              & 0.87 ± 0.03                   \\
          & GAE     & 0.42 ± 0.05               & 40.56 ± 40.21                & 13.56 ± 15.43          & 0.38 ± 0.16                              & 0.88 ± 0.07                   \\
          & VGAE    & 0.42 ± 0.05               & 39.22 ± 21.40                & 11.52 ± 5.89           & 0.39 ± 0.16                              & 0.83 ± 0.07                   \\
          & GRACE   & 0.49 ± 0.05               & \underline{54.22 ± 60.21}          & 10.94 ± 8.63           & 0.36 ± 0.22                              & 0.84 ± 0.04                   \\
          & GNUMAP  & \textbf{0.66 ± 0.03}      & 46.68 ± 30.84                & \underline{10.88 ± 7.70}     & \textbf{0.48 ± 0.15}                     & \textbf{0.99 ± 0.01}          \\ \hline
Moons     & DGI     & 0.57 ± 0.06               & 104.58 ± 51.66               & 3.80 ± 2.12            & 0.21 ± 0.10                              & 0.71 ± 0.06                   \\
          & BGRL    & 0.62 ± 0.05               & 125.24 ± 50.80               & 3.08 ± 2.02            & 0.20 ± 0.07                              & 0.76 ± 0.05                   \\
          & CCA-SSG & 0.70 ± 0.07               & 308.89 ± 178.11              & \underline{2.02 ± 1.05}      & 0.34 ± 0.18                              & \underline{0.90 ± 0.04}             \\
          & SPAGCN  & \underline{0.70 ± 0.05}         & 160.19 ± 73.17               & 2.32 ± 1.42            & 0.21 ± 0.06                              & 0.89 ± 0.03                   \\
          & GAE     & 0.57 ± 0.07               & \textbf{768.68 ± 516.56}     & 3.01 ± 1.70            & 0.33 ± 0.13                              & 0.85 ± 0.05                   \\
          & VGAE    & 0.56 ± 0.06               & \underline{620.11 ± 426.56}        & 3.36 ± 1.78            & \underline{0.36 ± 0.17}                        & 0.80 ± 0.06                   \\
          & GRACE   & 0.65 ± 0.06               & 238.53 ± 132.55              & 2.51 ± 1.60            & 0.31 ± 0.19                              & 0.86 ± 0.04                   \\
          & GNUMAP  & \textbf{0.88 ± 0.02}      & 559.49 ± 163.03              & \textbf{0.91 ± 0.45}   & \textbf{0.45 ± 0.13}                     & \textbf{0.99 ± 0.01}          \\ \hline
Swissroll & DGI     & 0.18 ± 0.03               & 53.80 ± 17.87                & 6.59 ± 1.42            & 0.20 ± 0.11                              & 0.56 ± 0.06                   \\
          & BGRL    & 0.20 ± 0.03               & 53.69 ± 11.91                & 5.95 ± 1.48            & 0.25 ± 0.10                              & 0.61 ± 0.05                   \\
          & CCA-SSG & 0.21 ± 0.03               & 120.78 ± 43.25               & \underline{4.55 ± 1.26}      & 0.38 ± 0.16                              & \underline{0.76 ± 0.06}             \\
          & SPAGCN  & \underline{0.22 ± 0.03}         & 93.13 ± 18.85                & 4.91 ± 2.55            & 0.27 ± 0.08                              & 0.72 ± 0.04                   \\
          & GAE     & 0.16 ± 0.02               & \underline{216.94 ± 72.76}         & 5.74 ± 1.99            & \underline{0.66 ± 0.12}                        & 0.74 ± 0.03                   \\
          & VGAE    & 0.15 ± 0.02               & 173.35 ± 42.33               & 6.41 ± 2.22            & 0.64 ± 0.12                              & 0.72 ± 0.03                   \\
          & GRACE   & 0.20 ± 0.02               & 102.91 ± 26.44               & 4.82 ± 1.32            & 0.37 ± 0.14                              & 0.70 ± 0.04                   \\
          & GNUMAP  & \textbf{0.28 ± 0.03}      & \textbf{314.90 ± 86.80}      & \textbf{2.53 ± 0.50}   & \textbf{0.88 ± 0.08}                     & \textbf{0.96 ± 0.02}          \\ \hline
\end{tabular}
    \label{table:synthetic_metric}
\end{table*}

 % make it easy to edit the table later

\xhdr{Discussion} Overall, GNUMAP consistently outperforms other self-supervised GNN-based methods in these small examples. GNUMAP successfully embeds blobs, swissroll, circles, and moons in 2-dimensions in a way that differentiates ground truth labels more clearly than state-of-the-art GNN-based methods. In Table~\ref{table:synthetic_metric}, our method outperforms state-of-the-art GNN-based methods in all five of our proposed metrics and the GNUMAP embedding visualization in Figure~\ref{fig:synthetic_visual} is coherent with the metrics. Note that GNUMAP successfully unrolls the Swissroll to closely resemble the original data, which is a standard synthetic dataset where linear methods fail to correctly embed.

% \textbf{Synthetic Blobs}
% GNUMAP outperforms other GNN methods in all metrics except for Calinski Harabasz score, which is outperformed by CCA-SSG, but the standard deviation for CCA-SSG is very large. Such findings are coherent with the embedding visualizations, as GNUMAP clearly distinguishes the four clusters. CCA-SSG embeddings also had clearly established blobs, but embeddings were concentrated in almost a line, implying a potential oversmoothing problem. On the other hand, when compared with traditional dimensionality reduction methods, GNUMAP is on par or outperformed on all metrics. Blobs appear to be a dataset better captured by traditional dimensionality reduction, and the expressivity of graph neural networks does not seem to improve embedding quality in this case.

\section{Evaluating unsupervised GNNs' embedding performance on real-world data}

To assess GNUMAP's performance across real-world data, we conducted a comparative analysis using well-established graph benchmark datasets: Cora, Citeseer, and Pubmed. They are standard citation network benchmark datasets \cite{yang2016revisiting}. In these networks, nodes represent scientific publications, and edges denote citation links between publications. Node features are the bag-of-words representation of papers, and node label is the academic topic of a paper.

We also incorporated the Mouse Spleen cell data \cite{goltsev2018deep} to assess GNUMAP performance in biological applications. The dataset was originally generated by ``co-detection by indexing"(CODEX) techniques, which is to take the image of a tissue section, and at each of the tissue locations, the relevant biomarkers information is measured and recorded \cite{goltsev2018deep}. For our experiments, we created a 5-nearest-neighboring-cell graph from the CODEX data. While the 3-topic cluster labels in the Mouse Spleen dataset are generated by the spatial LDA model, previous work \cite{spatial-lda} has established that such topic annotations are biologically consistent with immunologist-labeled topics. Therefore, we will regard the 3-topic cluster labels as ground truth labels in the Mouse Spleen dataset for the subsequent analysis.
% In addition, we implemented the California Road Network dataset\cite{road_cal} to demonstrate GNUMAP performance with a more realistic data with a small-world property. While the original dataset did not have class labels, we assigned 10 labels based on the node's latitude information.
For all experiments with real-world datasets, GNUMAP was compared against CCA-SSG, SPAGCN, and GAE.

\begin{figure*}[h!]
    \centering
        \begin{tabular}{|p{1.4cm}|c|c|c|c|c|}
        \hline
        Dataset & SPAGCN & GAE & CCA-SSG & GNUMAP \\
        \hline
        Cora &
        \includegraphics[height=2cm, width=2cm, trim={1.35cm 1.2cm 1.2cm 1.2cm},clip]{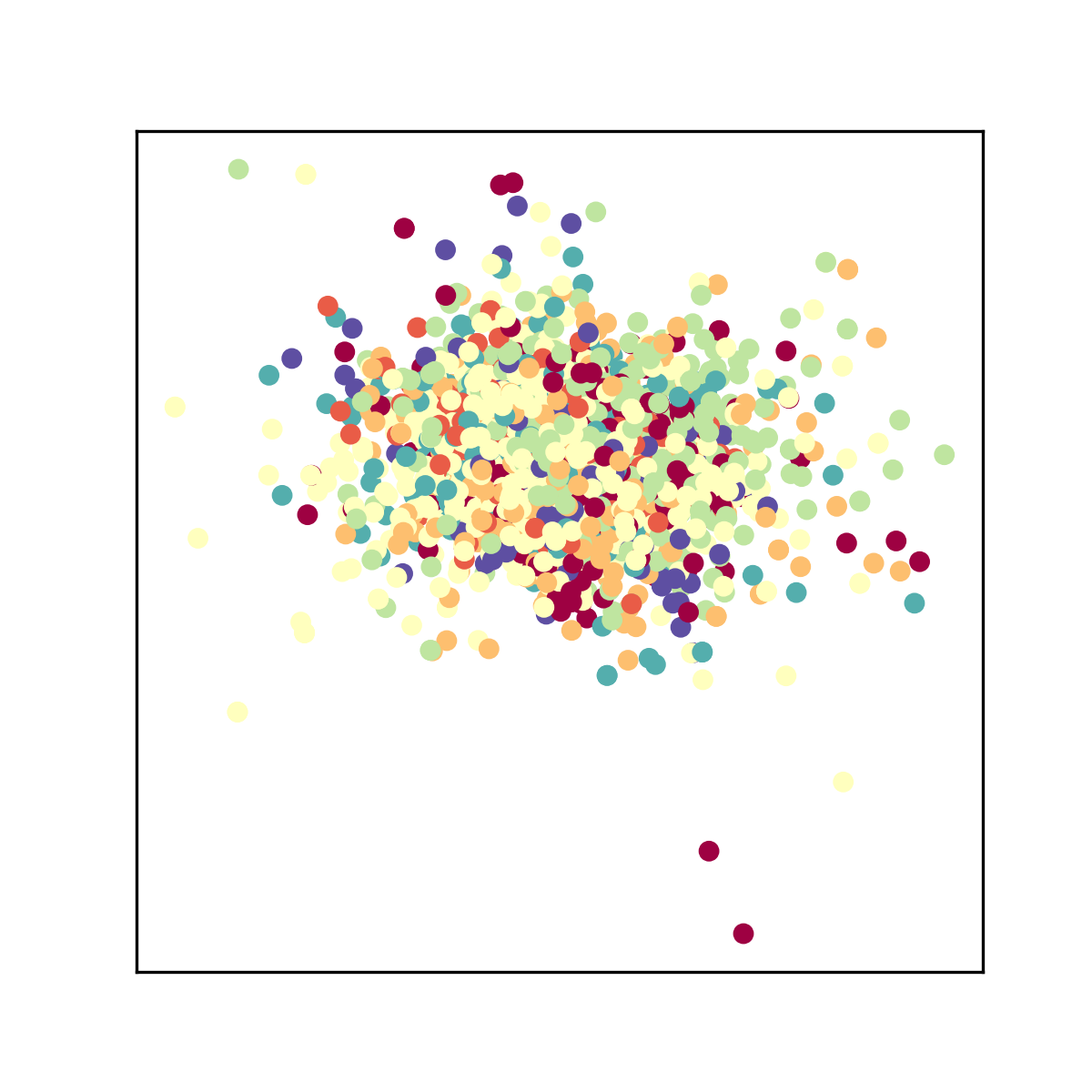} &
        \includegraphics[height=2cm, width=2cm, trim={1.35cm 1.2cm 1.2cm 1.2cm},clip]{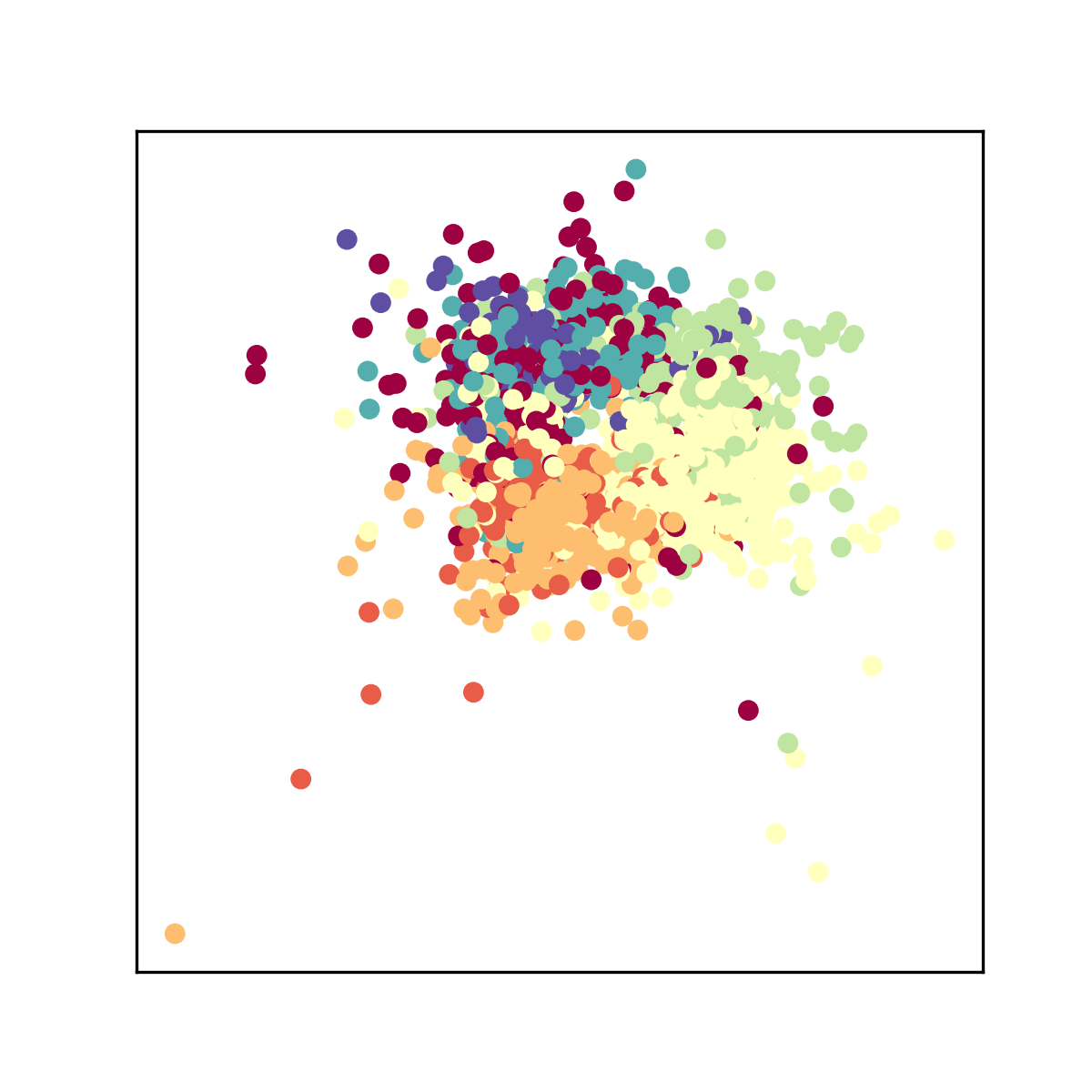} &
        \includegraphics[height=2cm, width=2cm, trim={1.35cm 1.2cm 1.2cm 1.2cm},clip]{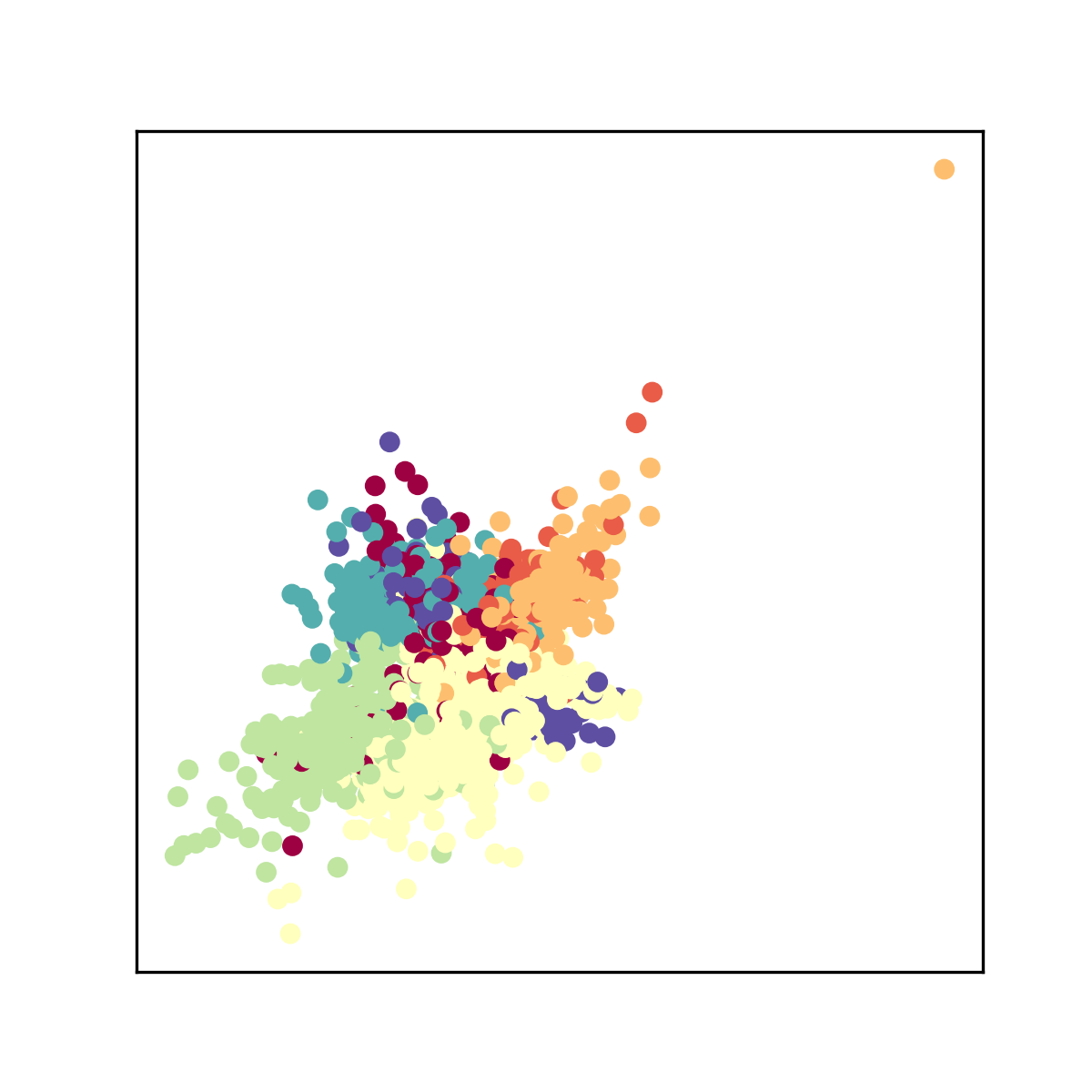} &
        \includegraphics[height=2cm, width=2cm, trim={1.35cm 1.2cm 1.2cm 1.2cm},clip]{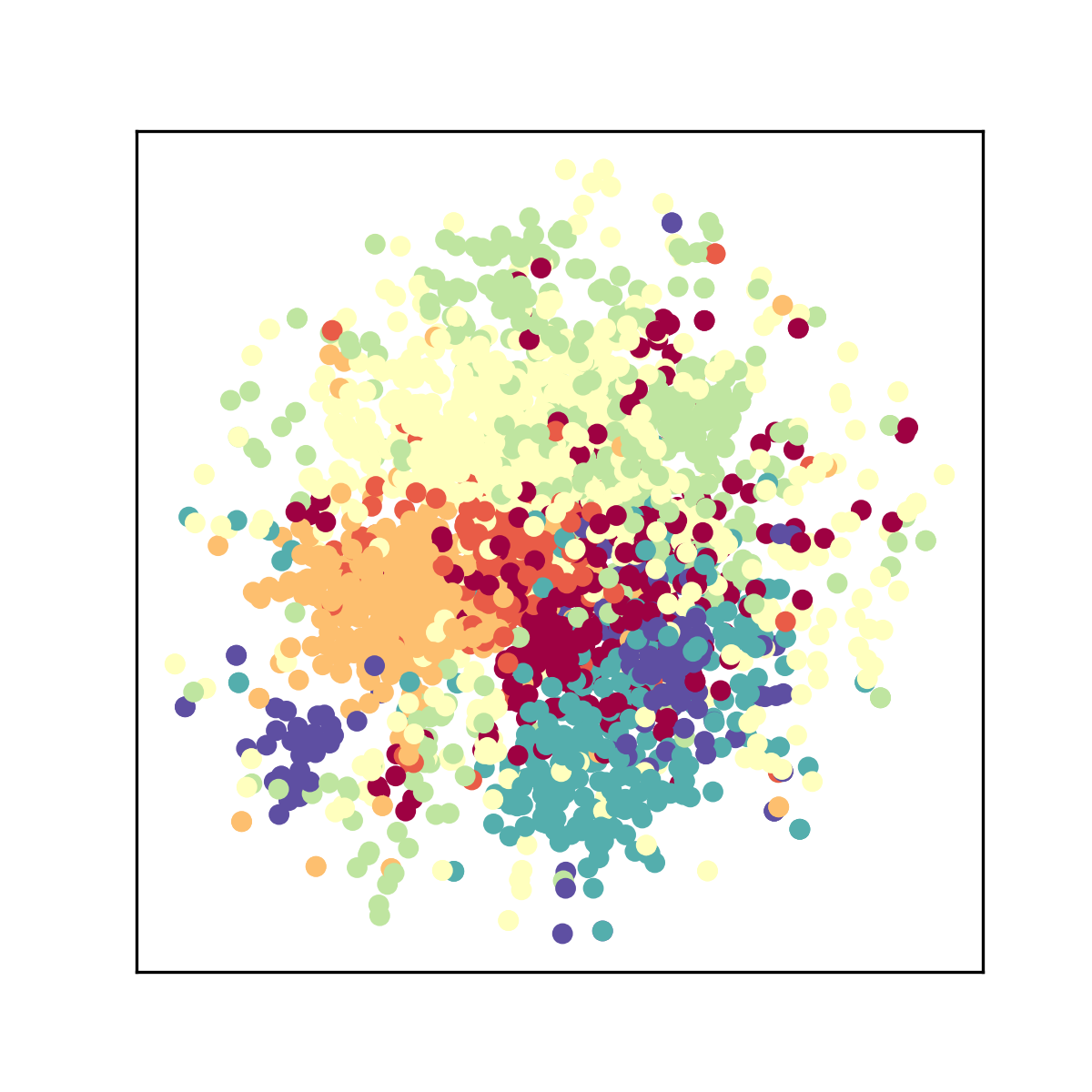} \\
        \hline
        Pubmed &
        \includegraphics[height=2cm, width=2cm, trim={1.35cm 1.2cm 1.2cm 1.2cm},clip]{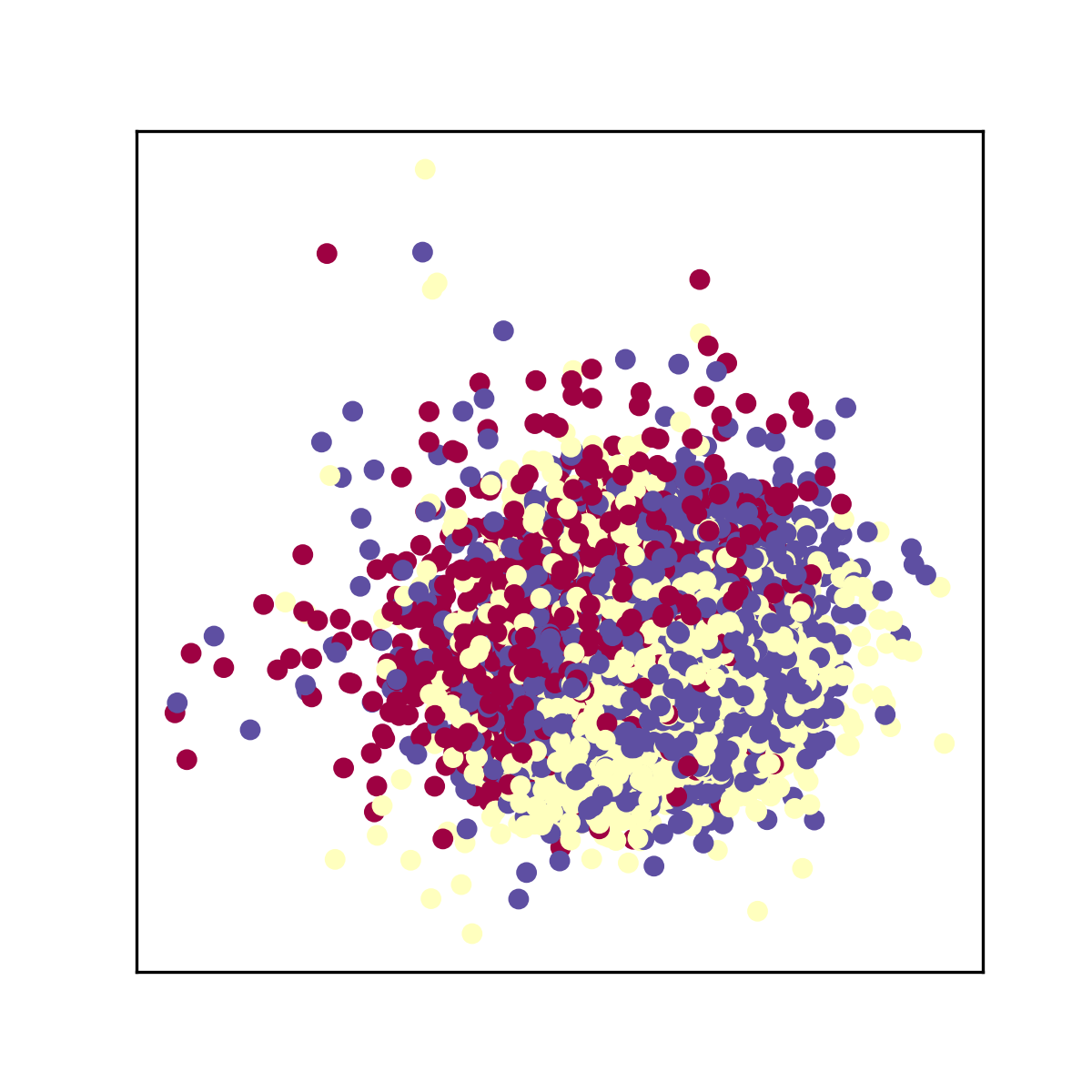} &
        \includegraphics[height=2cm, width=2cm, trim={1.35cm 1.2cm 1.2cm 1.2cm},clip]{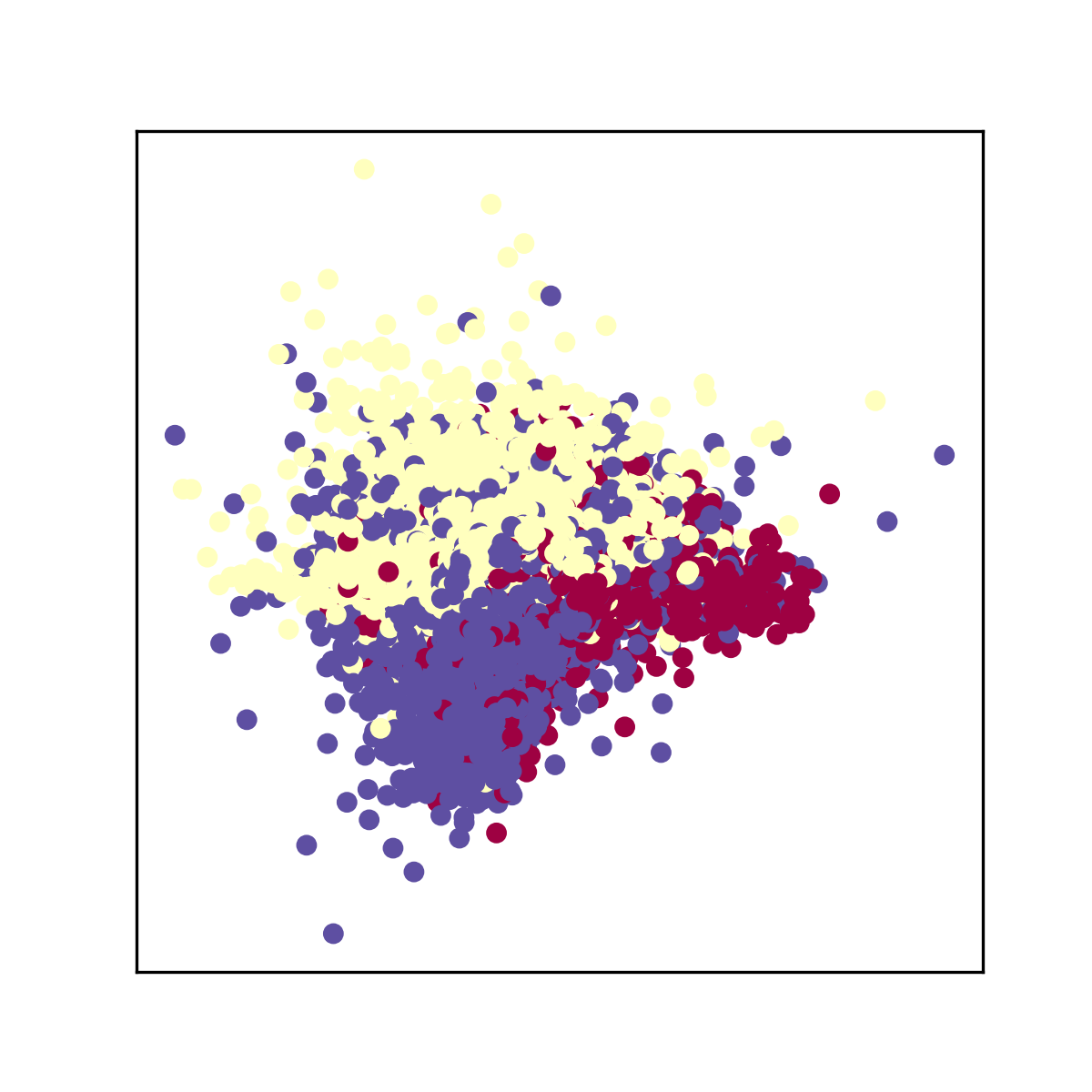} &
        \includegraphics[height=2cm, width=2cm, trim={1.35cm 1.2cm 1.2cm 1.2cm},clip]{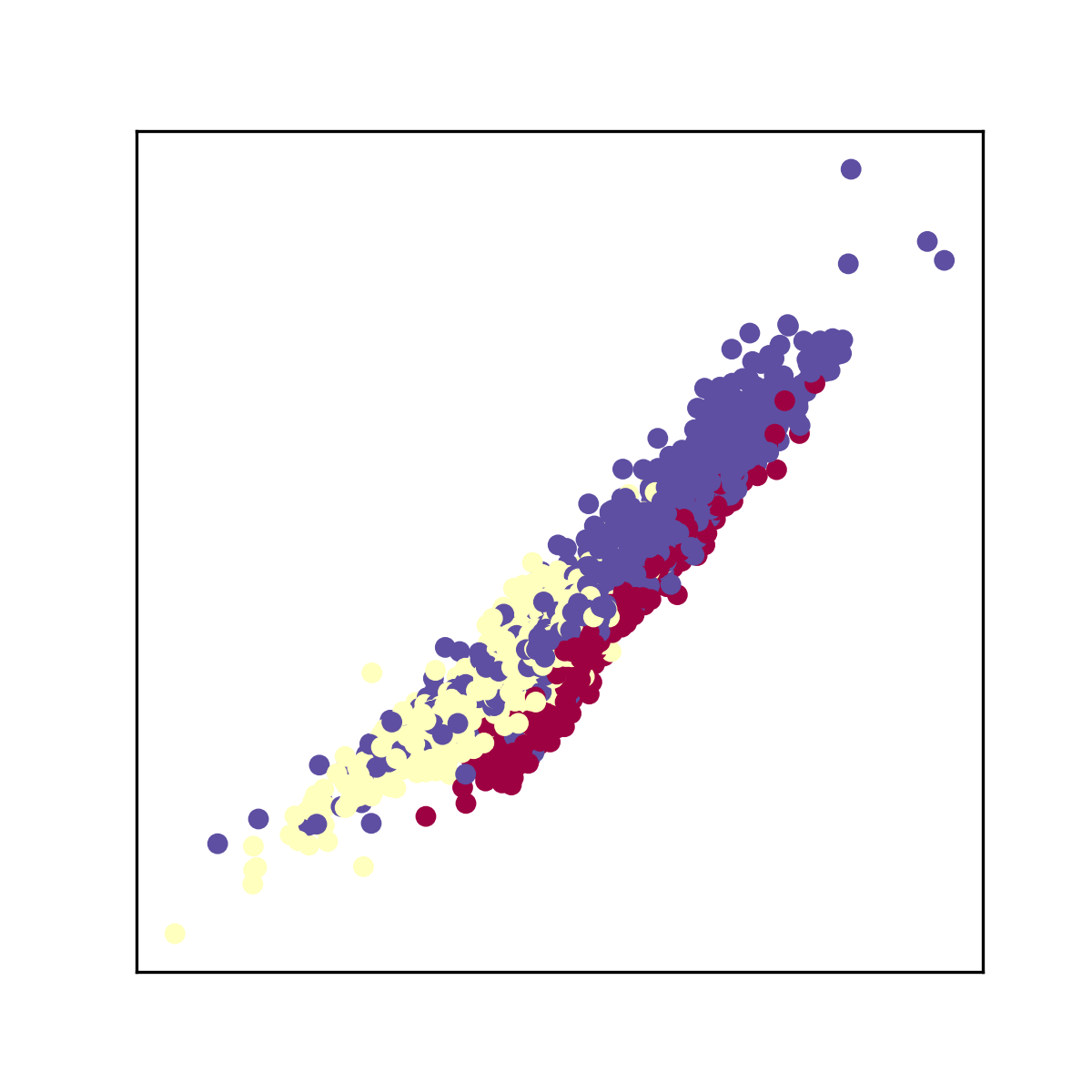} &
        \includegraphics[height=2cm, width=2cm, trim={1.35cm 1.2cm 1.2cm 1.2cm},clip]{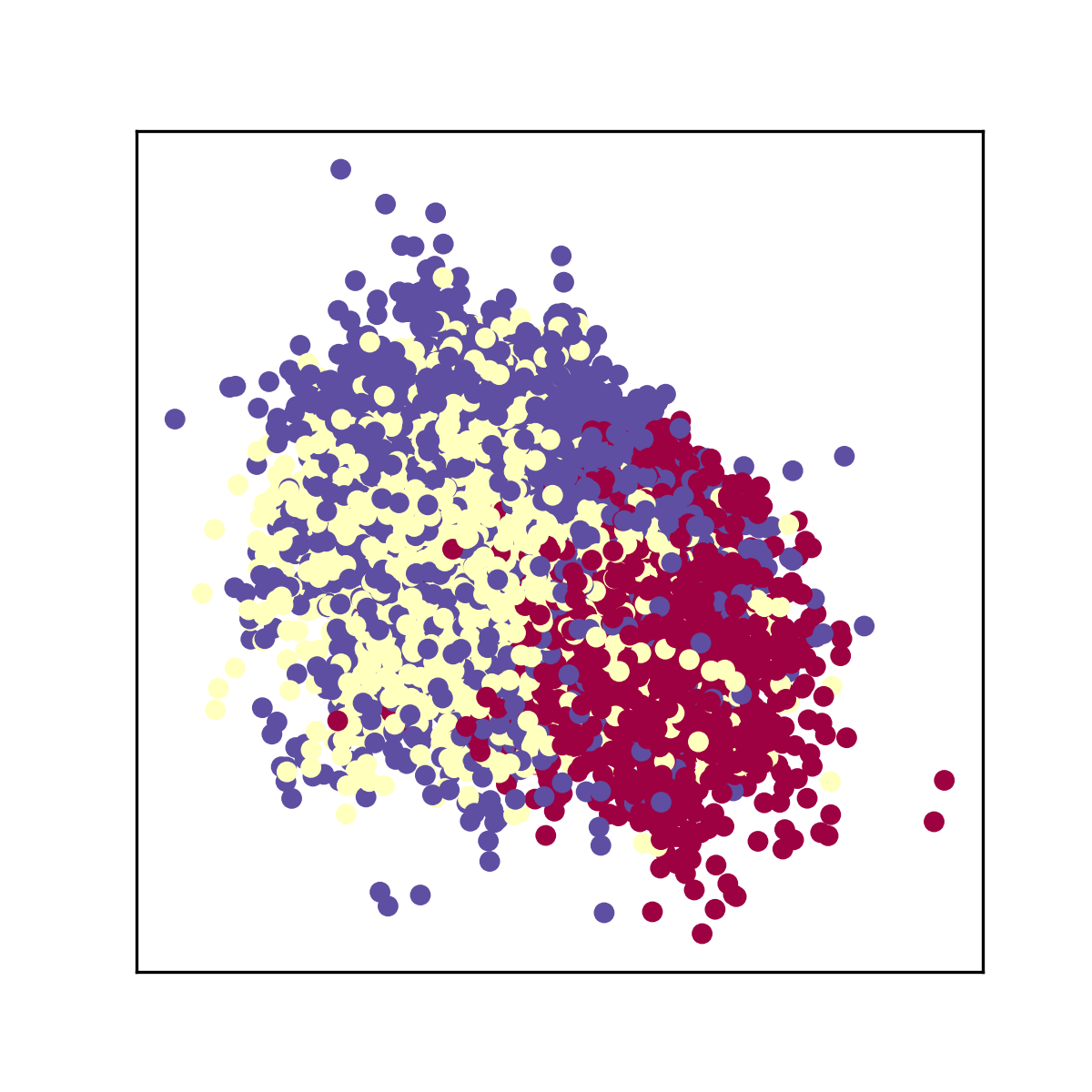} \\
        \hline   
        Citeseer &
        \includegraphics[height=2cm, width=2cm, trim={1.35cm 1.2cm 1.2cm 1.2cm},clip]{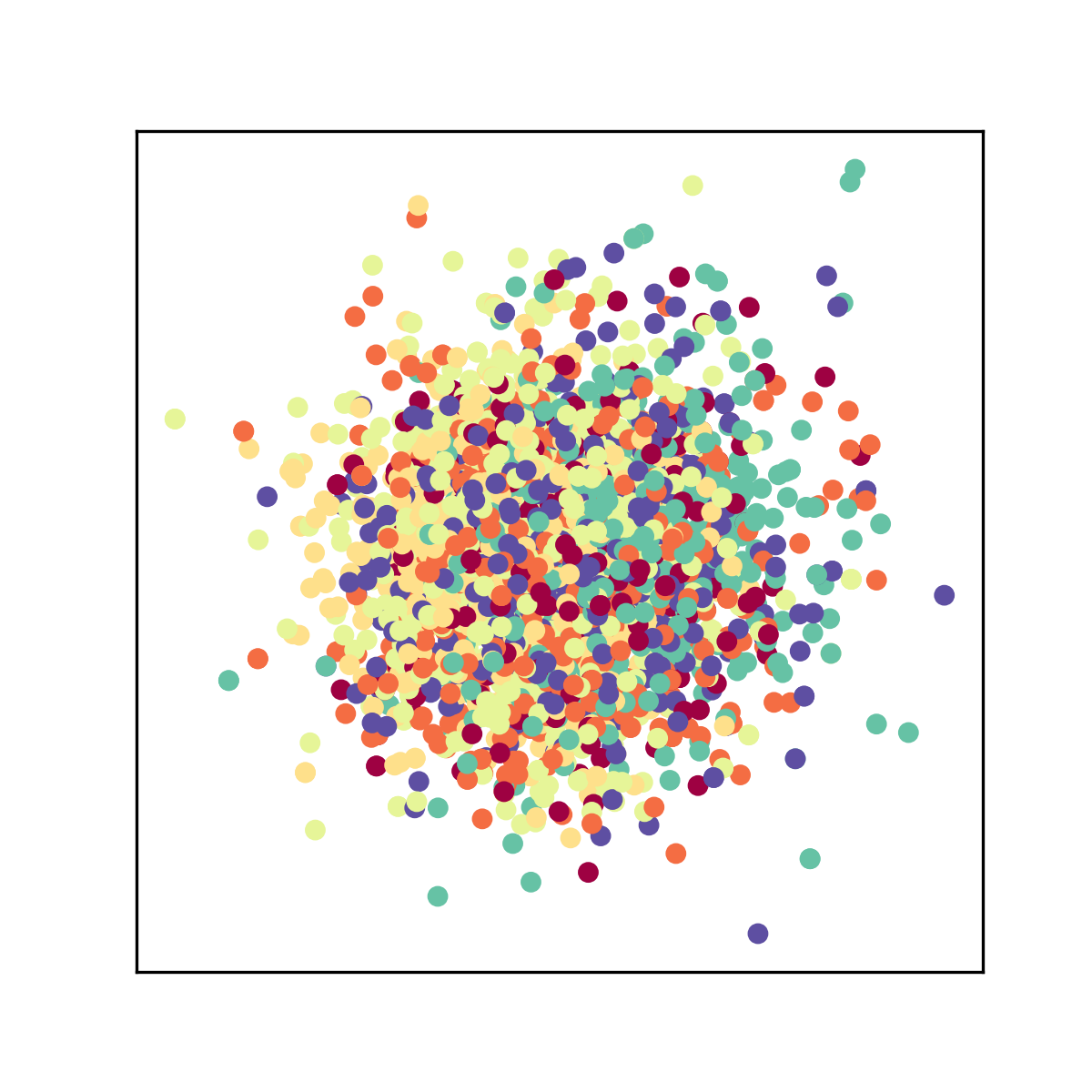} &
        \includegraphics[height=2cm, width=2cm, trim={1.35cm 1.2cm 1.2cm 1.2cm},clip]{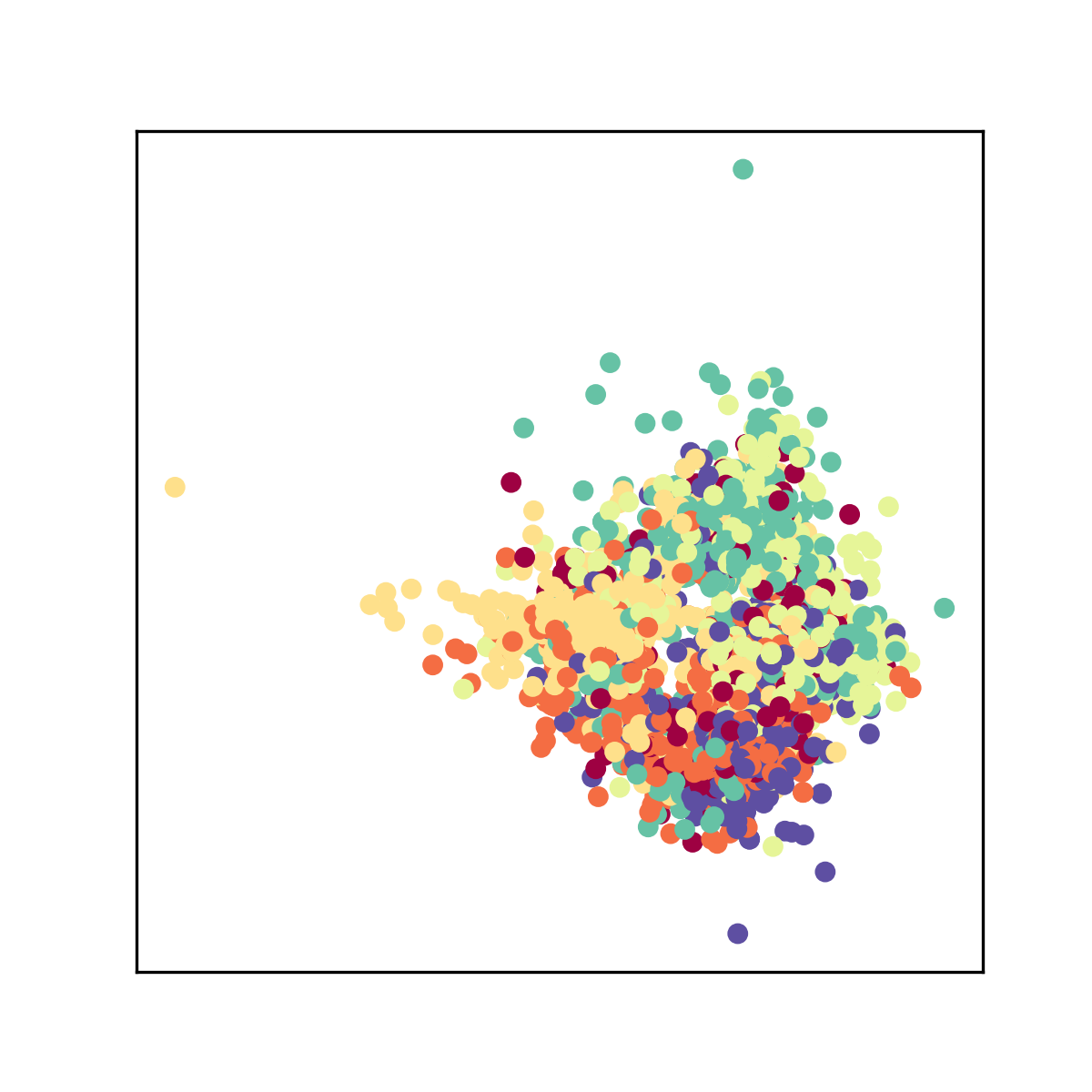} &
        \includegraphics[height=2cm, width=2cm, trim={1.35cm 1.2cm 1.2cm 1.2cm},clip]{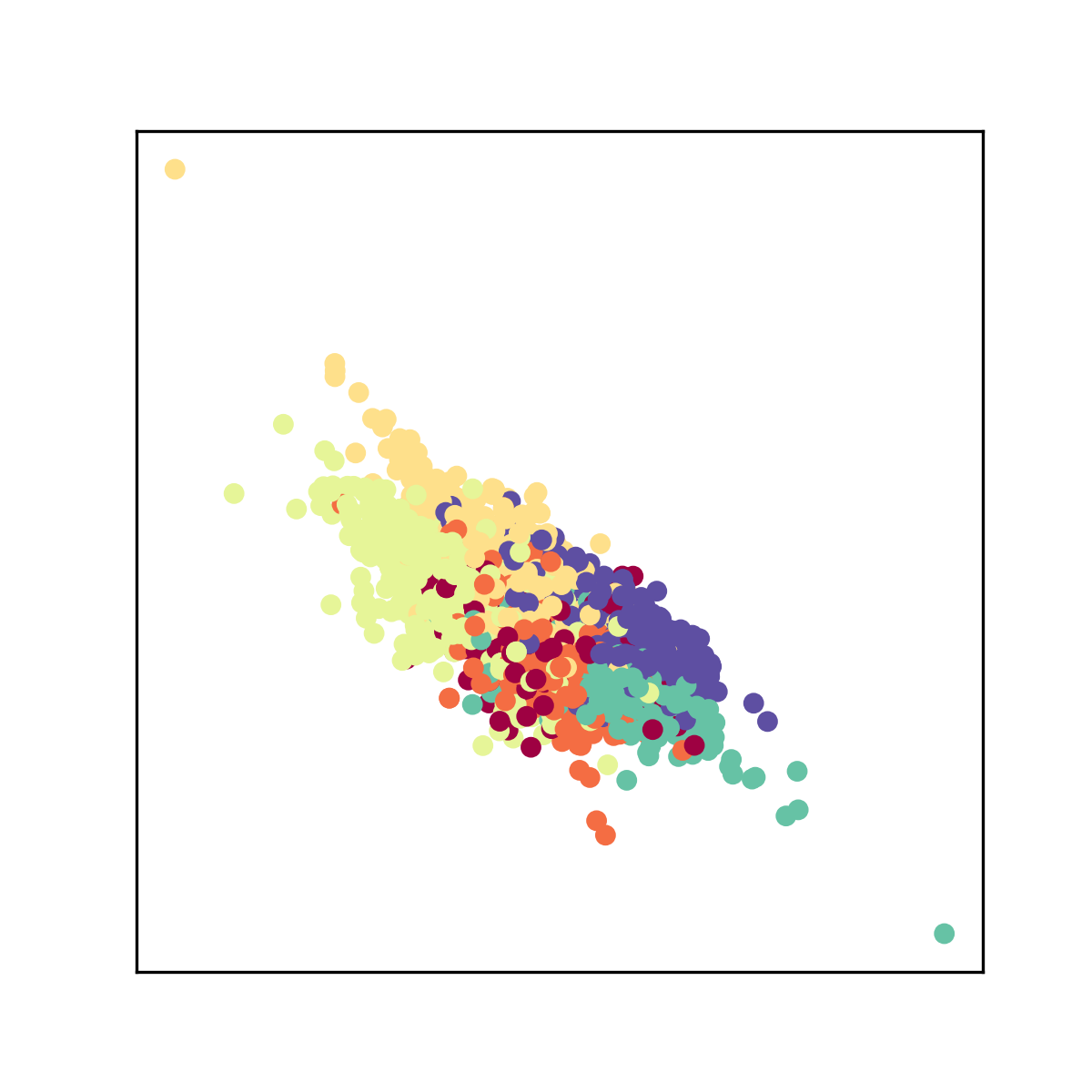} 
        &
        \includegraphics[height=2cm, width=2cm, trim={1.35cm 1.2cm 1.2cm 1.2cm},clip]{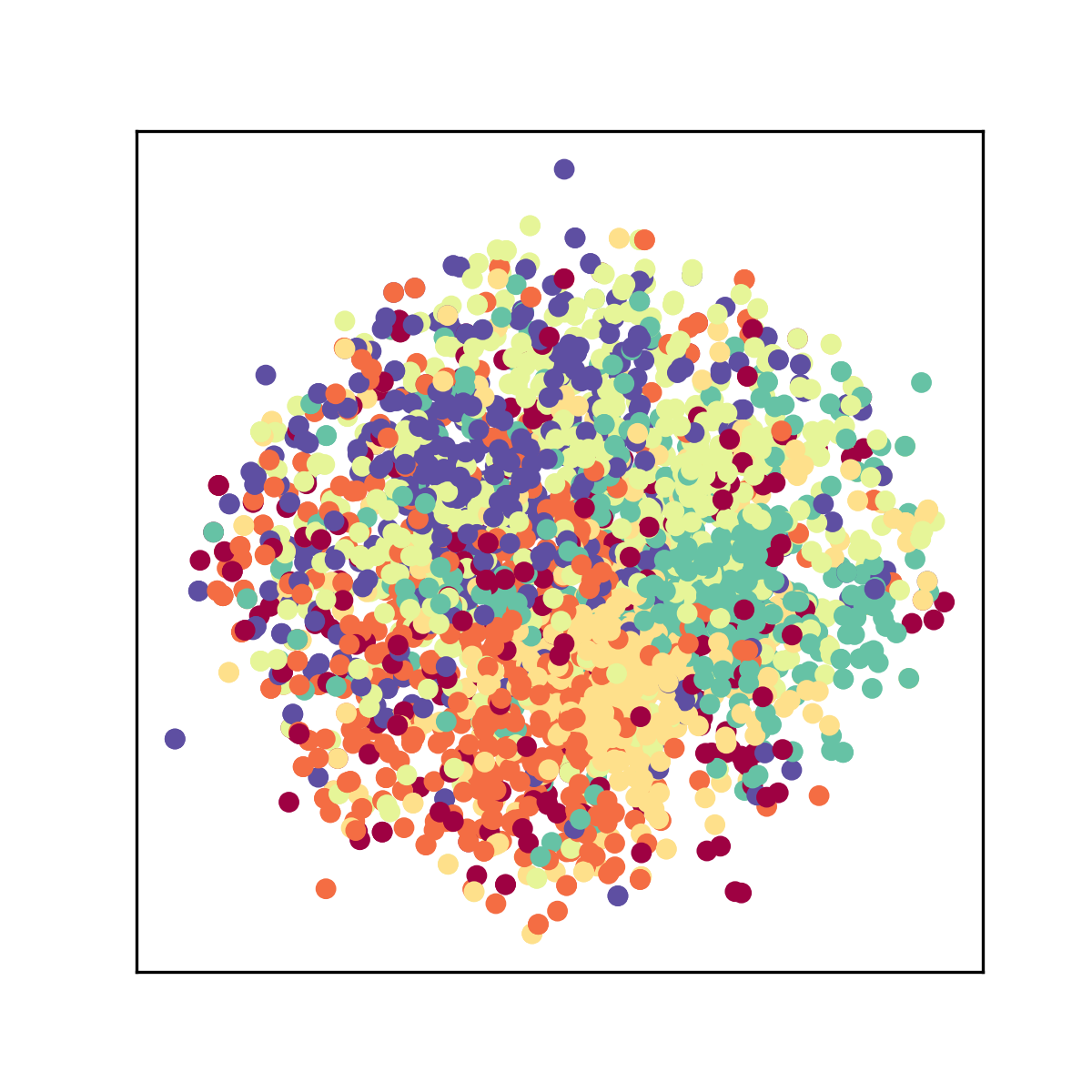} \\
        \hline     
    \end{tabular}
    \caption{Node representation learning for real-world datasets Cora, Citeseer, Pubmed.}
    \label{fig:real_visual}
\end{figure*}

All datasets come with or were assigned class labels, enabling a comparison between embedding visualization and our calculated metrics. For scalable metric evaluation, we randomly sampled 1000 datapoints and calculated the following metrics: classification accuracy, Calinski Harabasz score, and Davies Bouldin score. These metrics convey the general degree of embedding informativity and the quality of clustering, which is appropriate for evaluating Cora, Citeseer, Pubmed, and Mouse Spleen data as one expects some clustering among the same classes. The results are presented in Figure~\ref{fig:real_visual} and Table~\ref{table:real}.

\begin{figure*}[h!]
    \centering
        \begin{tabular}{|p{1.4cm}|c|c|c|c|c|c|}
        \hline
        Dataset & Original Data & SPAGCN & GAE & CCA-SSG & GNUMAP \\
        \hline
        Mouse Spleen &
        \includegraphics[height=2cm, width=2cm, trim={1.35cm 1.2cm 1.2cm 1.2cm},clip]{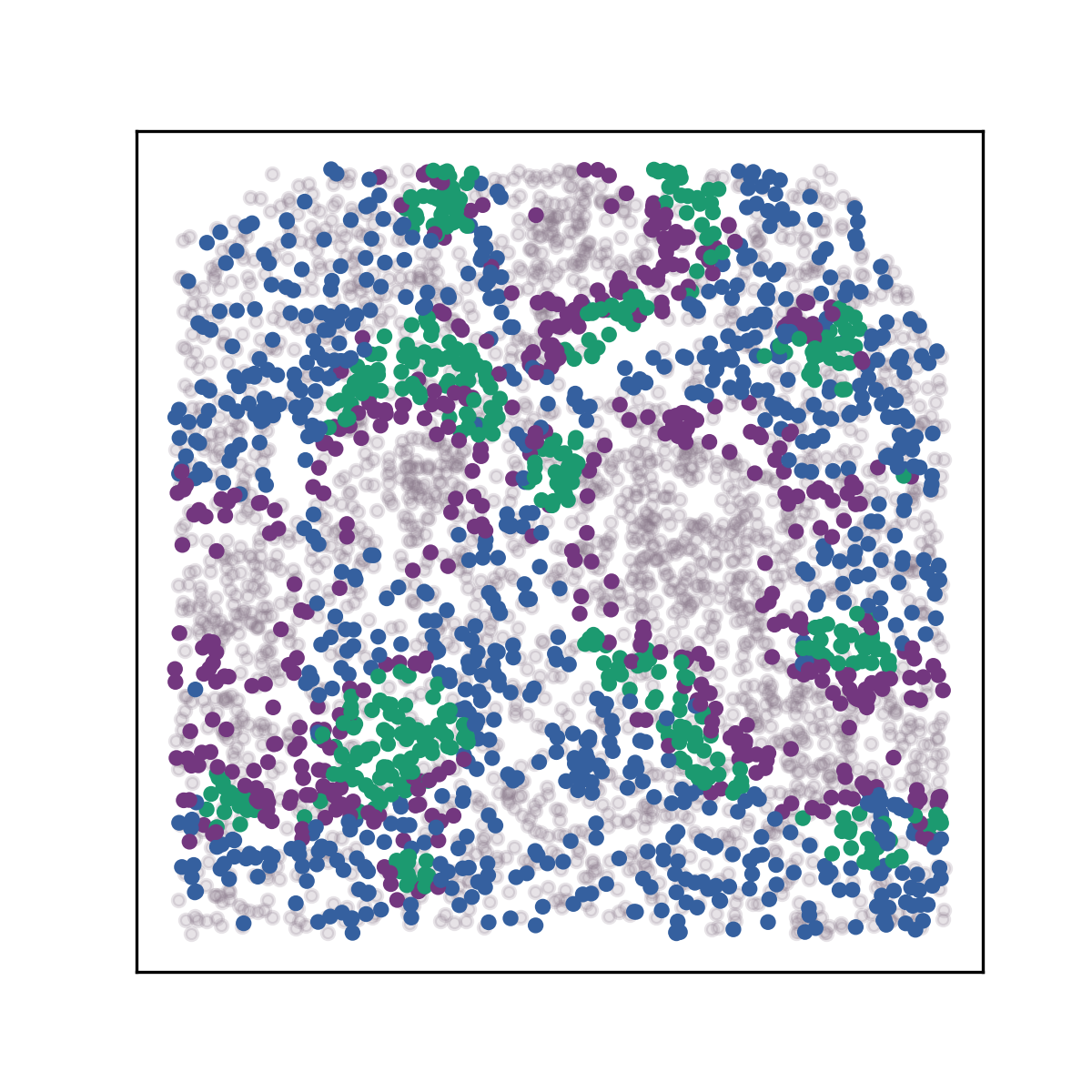} &
        \includegraphics[height=2cm, width=2cm, trim={1.35cm 1.2cm 1.2cm 1.2cm},clip]{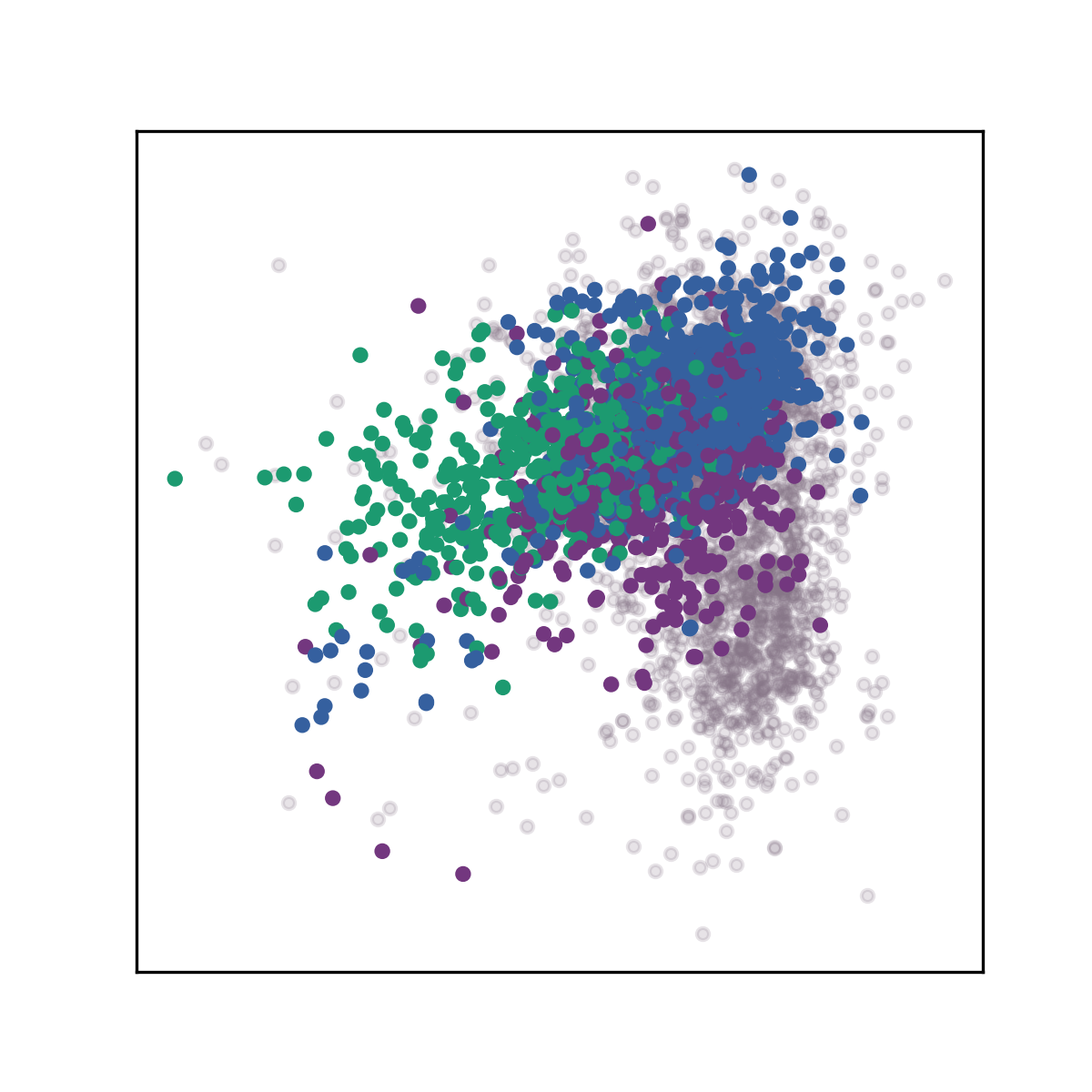} &
        \includegraphics[height=2cm, width=2cm, trim={1.35cm 1.2cm 1.2cm 1.2cm},clip]{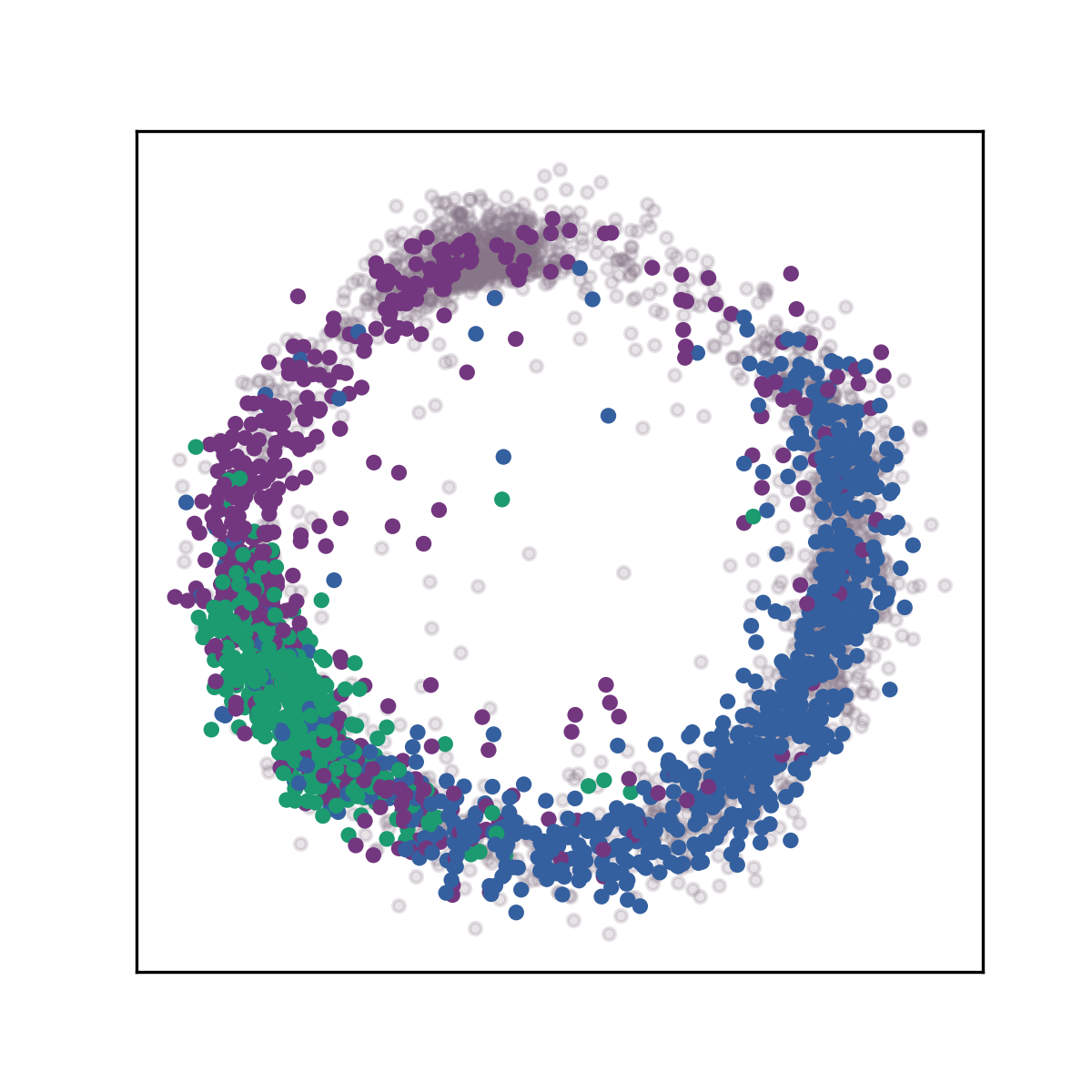} &
        \includegraphics[height=2cm, width=2cm, trim={1.35cm 1.2cm 1.2cm 1.2cm},clip]{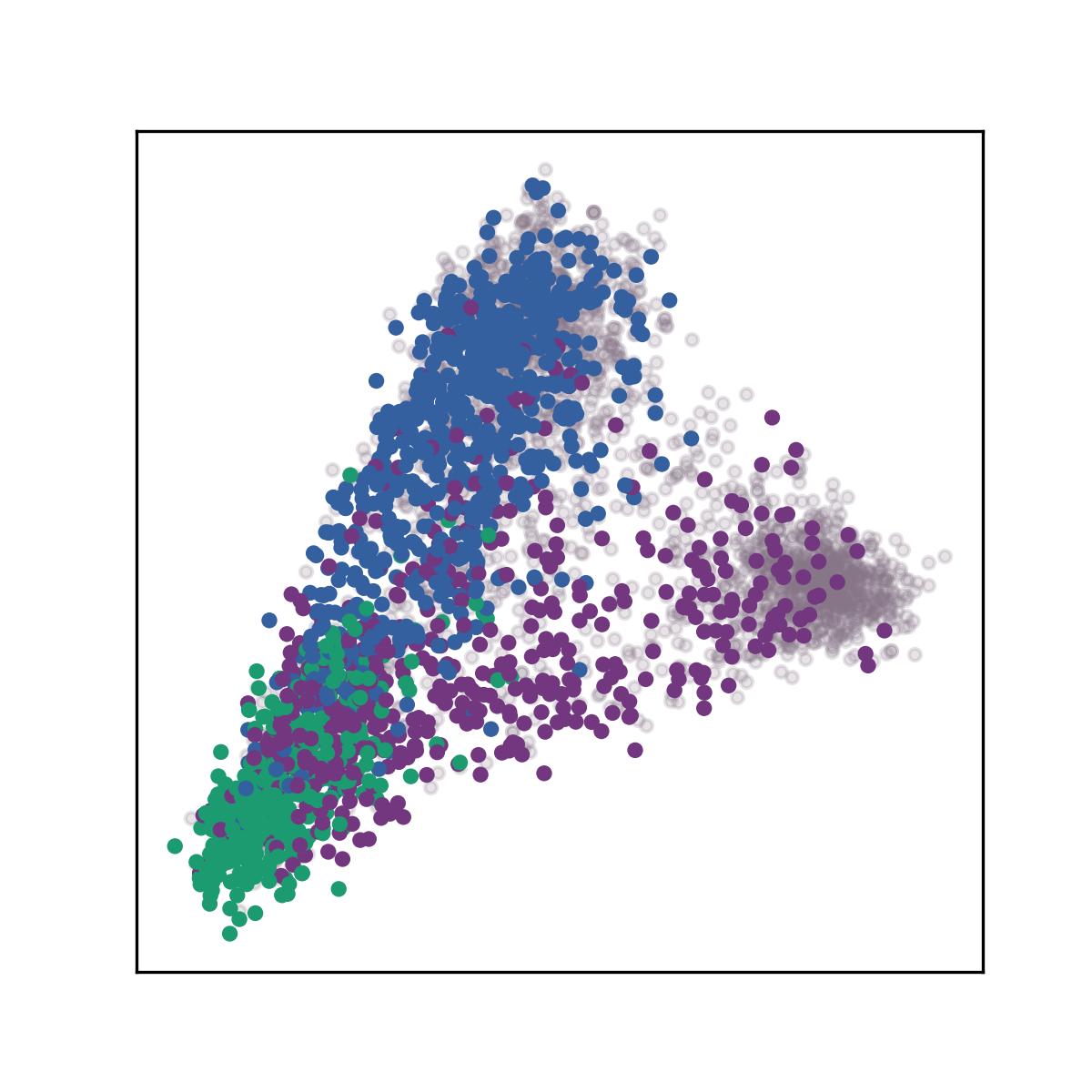} &
       \includegraphics[height=2cm, width=2cm, trim={1.35cm 1.2cm 1.2cm 1.2cm},clip]{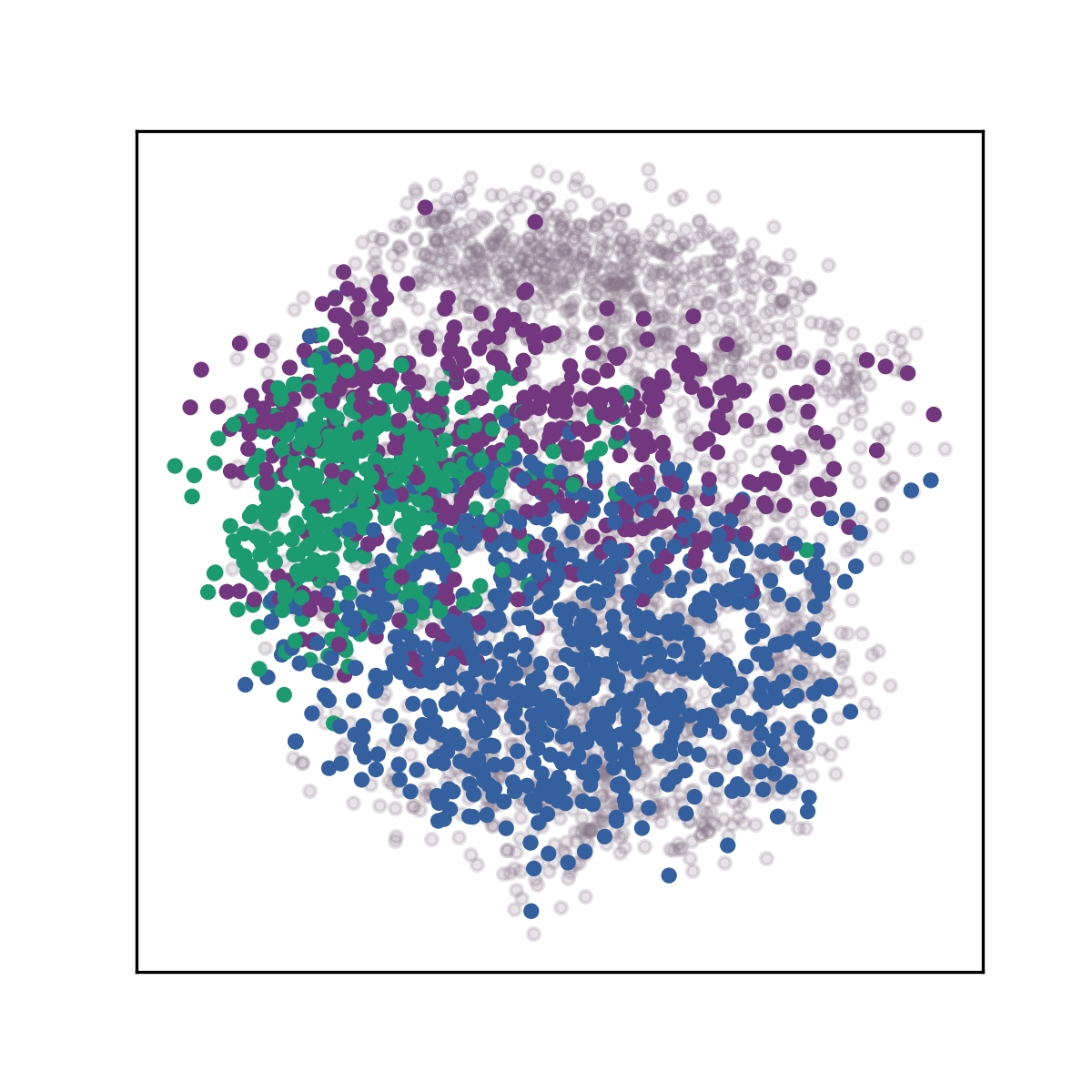} \\
        \hline   
    \end{tabular}

    \caption{Node representation learning for Mouse Spleen dataset. Colours represent assigned ground truth cluster label. Blue denotes B-cells, purple denotes marginal zone B-cells, gray denotes non-B cells, and green denotes red pulp~\cite{spatial-lda}.}
    \label{fig:mouse}
\end{figure*}

\begin{table*}[h!]
    \centering
        \caption{Summary mean and standard deviation of proposed metrics over multiple experiments with real-world datasets Cora, Citeseer, Pubmed, and Mouse Spleen. The upward arrows next to the metric name denote a better performance with higher metrics, whereas the downward arrows denote a better performance with lower metrics. The best result is bolded, and the next best result is underlined.}
        \vspace{1em}
        \begin{tabular}{|p{2.1cm}|p{2.2cm}|p{2.5cm}|p{2.5cm}|p{2.5cm}|}
        \hline
 \rowcolor[HTML]{C0C0C0} 
Dataset      & Model   & Classification \newline Accuracy ↑ & Calinski Harabasz Score ↑                      & Davies Bouldin Score ↓                      \\ \hline
Cora         & CCA-SSG & 0.47 ± 0.11                                 & \textbf{107.12 ± 86.14} & 16.31 ± 25.76                               \\
             & SPAGCN  & 0.31 ± 0.01                                 & 10.15 ± 4.25                                   & 17.24 ± 13.18                               \\
             & GAE    & \underline{0.50 ± 0.03}                        & \underline{104.19 ± 20.61}                               & \underline{6.08 ± 6.18}                        \\
             & GNUMAP  & \textbf{0.64 ± 0.04}& 87.90 ± 21.08                        & \textbf{4.58 ± 3.19} \\ \hline 
Pubmed       & CCA-SSG & 0.59 ± 0.09                                 & 67.36 ± 69.68                                  & 14.66 ± 18.52                               \\
             & SPAGCN  & 0.45 ± 0.03                                 & 16.99 ± 12.01                                  & 12.69 ± 14.60                               \\
             & GAE     & \textbf{0.70 ± 0.01} & \textbf{144.36 ± 10.58} & \textbf{2.20 ± 0.17} \\
             & GNUMAP & \underline{0.67 ± 0.01}                       & \underline{142.50 ± 19.38}                       & \underline{2.81 ± 0.30}                        \\ \hline
Citeseer     & CCA-SSG & \underline{0.50 ± 0.09}                                 & \textbf{159.64 ± 77.50} & 8.56 ± 17.91                                \\
             & SPAGCN  & 0.26 ± 0.02                                 & 7.45 ± 3.52                                    & 23.18 ± 20.15                               \\
             & GAE    & 0.47 ± 0.03                       & \underline{81.11 ± 12.58}                         & \textbf{5.78 ± 2.57} \\
             & GNUMAP  & \textbf{0.52 ± 0.03} & 51.81 ± 12.05                                 & \underline{7.64 ± 4.38}                        \\ \hline
Mouse Spleen & CCA-SSG & \textbf{0.66 ± 0.02} & \textbf{138.08 ± 37.43}& \underline{2.30 ± 1.19}                        \\
             & SPAGCN  & 0.60 ± 0.03                                 & 75.92 ± 40.20                                  & 4.72 ± 4.63                                 \\
             & GAE    & \underline{0.65 ± 0.02}                       & \underline{126.91 ± 9.52}                         & \textbf{2.11 ± 0.11} \\
             & GNUMAP  & 0.62 ± 0.02                                 & 79.78 ± 12.22                                  & 2.74 ± 0.32    \\     \hline
        \end{tabular}
    \label{table:real}
\end{table*}

%\subsection{Analysis of the Results}
\xhdr{Discussion}
On homophilic citation networks, GNUMAP is consistently the best or second-best for two of the three metrics: classification accuracy and Davies Bouldin score. For Cora, GNUMAP outperforms all other models in classification accuracy and Davies Bouldin score, and GNUMAP produces the second best Clinski Harabasz score after CCA-SSG. On Pubmed, GAE was the best-performing model on all three metrics, and GNUMAP was a close second. The autoencoder framework appears especially effective in this data set. GAE was the best-performing model on all three metrics for Citeseer, and GNUMAP was a close second. Autoencoder framework appears especially effective on this dataset. On the Mouse Spleen dataset, CCA-SSG and GAE perform best or second-best over all three evaluation metrics. GNUMAP is the third-best method across all metrics, and SpaGCN is the worst. However, since SpaGCN was designed specifically for biological application, such poor performance was unexpected, and this suggests the 5-nearest-neighboring-cells graph preprocessing method was unable to reflect the mouse spleen's biological properties. This allows for future exploration for a separate preprocessing pipeline for converting biological data to graph. We further emphasize again hat for these examples, we chose the best performing hyperparameters for SPAGCN and CCA-SSG, thereby conferring these methods a significant advantage over the autoencoder setting. 

\section{Conclusion}
With the exponential growth in data complexity, statisticians and computer scientists have recognized graphs as a modeling framework for high-dimensional data types. Therefore, modeling with graphs, predominantly through GNNs, is gaining traction in various areas, especially the biological \cite{spatial-lda, hu2021spagcn} domain. Through this work, we contribute to the ongoing research on unsupervised graph learning by proposing GNUMAP: a GNN inspired by theoretical research on the dimensionality reduction technique UMAP. Through our unique metrics that evaluate informativeness (classification accuracy), preservation of original graph information (agreement in local geometry, spearman correlation, frechet distance), and cluster quality (Davies-Bouldin score, Calinski-Harabasz score, silhouette score), we establish GNUMAP as a robust and expressive GNN embedding method that performs outperforms many existing GNN embedding methods, as well as dimensionality reduction methods depending on the task. Clear clustering of GNUMAP embeddings across synthetic geometric datasets as well as Cora, Citeseer, Pubmed, and Mouse Spleen further highlights the adaptive power of GNUMAP. GNUMAP performance can be characterized as ``data-agnostic"--unlike state-of-the-art GNNs, there is minimal hyperparameter tuning required to enhance GNUMAP performance, allowing an easy application to any dataset.

\textbf{Limitations and Future Work}
The reconstruction-based approach that we propose here in this paper, GNUMAP, is by design extremely simple, and its objective is interpretable. While it performs well on the examples that we provide here, this method is only valid for homophilic datasets -- where edges encode similarities between nodes. However, this method also lends itself to natural extensions, such as accounting for cluster densities using the density coefficient of DenseMAP \cite{densmap}, to better capture the original data characteristics in low-dimensional embeddings.

\bibliographystyle{unsrt}
\bibliography{main}

\appendix
\onecolumn
\begin{center}
{\LARGE\bf Appendix}
\end{center}
\section{Metrics}\label{app:metrics} 

Current approaches to evaluating embeddings generally fall into one of two types: (a) visual methods, where individuals visually inspect the learned embeddings to assess if they align with their expectations, and (b) classification-based methods. In the classification-based approach, it is presumed that there exists a set of labels, not used during training, which effectively represent the structure of the underlying data. In our work, we extend these latter metrics for application to manifold scenarios (which involve continuous structures rather than discrete ones) and quantify the performance of the learned representations using the following metrics.
\begin{description}[leftmargin=0em]
    \item [General Metrics] To assess the embedding quality, we propose measuring the agreement between the learned embedding space and the original data. This agreement can in particular be measured using:
    \begin{itemize}
        \item \textbf{Agreement in Local Geometry}
        We compute the percentage of overlap between the original graph's k-nearest neighbors and knn graphs on the embedding space. A higher percentage denotes better local geometry preservation. 
        $$\frac{1}{n}\sum_{i=1}^{n}\frac{\text{\# of overlapping neighborhood on } x_i}{\text{node degree}(x_i)}$$
        \item \textbf{Spearman Correlation}
        Inspired by a methodology introduced in \cite{spearman_graph}, we evaluate the Spearman correlation between the pairwise distance in high-dimensional embedding space and embedding space from the model. Inspired by Isomap\cite{isomap}, the pairwise distance in the original space is evaluated by the shortest path distance, assuming that the geodesic distance on the manifold can be reasonably approximated by the shortest path distance. The pairwise distance in the low-dimensional space is evaluated by standard Euclidean distance. This metric evaluates the relationship between learned embedding space and the original graph using a monotonic function. The metric spans $[-1,1]$ where -1 denotes a relationship modeled by a perfectly decreasing monotonic function, and vice versa. A higher Spearman correlation implies a similarity between the learned embedding and the original data.
        \item \textbf{Classification Accuracy}
        While our embeddings are unsupervised, we assume quality embeddings would form well-separated and helpful features for the support vector machine(SVM) multi-class classification problem. Consequently, we divide our manifold into different clusters (in intrinsic space) and assess the accuracy of the embedding space in recovering the different clusters through SVM with radial basis function(RBF) kernel. We record the average accuracy in predicting cluster labels from embedding space coordinates in a 10-fold. Higher accuracy thus implies the informativity of the learned embeddings.
        \item \textbf{Frechet Distance}
        Finally, we implemented the Frechet inception distance \cite{frechet}, a metric originally proposed to measure the similarity between images, to evaluate the 2d Frechet distance between the learned embedding coordinates and the original data. Smaller Frechet distance denotes better similarity between embedding and the original data coordinates.
    \end{itemize}
    \item[Metrics for Clustered Data] For data that are expected to form clusters (e.g. synthetic blobs, citation networks, Mouse Spleen gene expression prediction), we propose metrics evaluating the quality of cluster separation in the embedding space.
    \begin{itemize}
        \item \textbf{Davies Bouldin Score}
        This metric is the average ratio of inter-cluster distance to intra-cluster distance to the most similar cluster\cite{davies_bouldin}. Davies Bouldin score is lower when clusters are concentrated and far apart. A lower score denotes better cluster separation, and the minimum score is zero.
        \item \textbf{Calinski Harabasz Score}
        This metric is the ratio of the sum of intra-cluster dispersion and of inter-cluster dispersion\cite{calinski_harabasz}. Calinski Harabasz score is higher when the intra-cluster variability is low and inter-cluster variability is high. A higher score indicates better score cluster separation.
        \item \textbf{Silhouette Score}
        This metric is the mean silhouette coefficient\cite{silhouette} of all data points. Let $a$ be the mean intra-cluster distance and $b$ be the mean nearest-cluster distance. Then, the silhouette coefficient for each data point is computed by $\frac{b - a}{\max(a, b)}$. Silhouette score is concerned with how similar a data point is to its assigned cluster compared to other clusters. The best value, 1, indicates well-separated and correctly assigned clusters, while the worst value, -1, often indicates that a sample is more similar to a different cluster. A silhouette score of 0 indicates overlapping clusters.
    \end{itemize}

\end{description}

\section{Effect of $\alpha$ and $\beta$ in Low Dimensional Representation}

 We will investigate the effect of $\alpha$ and $\beta$ in eq-\ref{eq:q_{ij}}. Recall the formula defining the edge connection probability in the low-dimensional space.

\begin{equation*}
q_{ij} = \frac{1}{1 + \alpha \times d(y_i, y_j)^{2\beta}}\
\end{equation*}

\begin{figure}[ht]
\centering
\includegraphics[width=\textwidth/2]{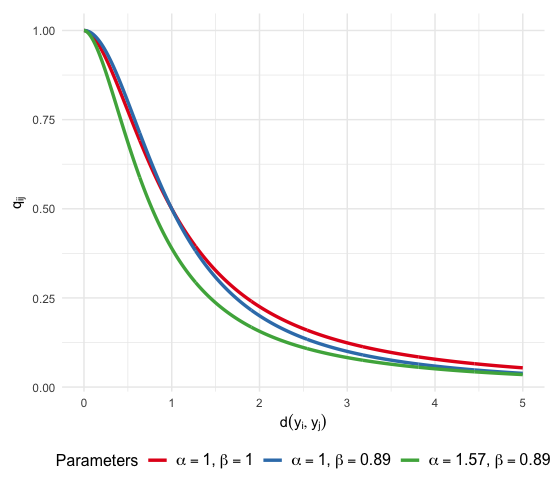}
    \caption{Comparisons of the effects of $\alpha$ and $\beta$ on the probability $q_{ij}$ according to eq-\ref{eq:q_{ij}}}
    \label{fig:tails}
\end{figure}

\begin{table*}
    \centering
    \caption{Evaluated mean and standard deviation of proposed metrics for GNUMAP alterations for 100 synthetic swissroll experiments. The upward arrows next to metric names denote better performance with higher metrics, whereas the downward arrows denote better performance with lower metrics.}
    \vspace{1em}

    \begin{tabular}{|>{\centering\arraybackslash}p{2.6cm}|>{\centering\arraybackslash}p{0.5cm}|>{\centering\arraybackslash}p{0.5cm}|>{\centering\arraybackslash}p{1.7cm}|>{\centering\arraybackslash}p{2cm}|>{\centering\arraybackslash}p{1.6cm}|>{\centering\arraybackslash}p{2.2cm}|>{\centering\arraybackslash}p{1.6cm}|}
        \hline
        \rowcolor[HTML]{C0C0C0} 
        Note & a & b & Classification Accuracy $\uparrow$ & Calinski Harabasz Score $\uparrow$ & Davies Bouldin Score $\downarrow$ & Spearman Correlation w/ Original Graph $\uparrow$ & Overlap \% of 50 Neighbours $\uparrow$ \\ \hline
        Baseline choice of $\alpha=\beta=1$ & 1 & 1 & 0.28 ± 0.03 & 329.94 ± 82.75 & 2.44 ± 0.51 & 0.88 ± 0.07 & 0.96 ± 0.02 \\ \hline
        Smaller minimum distance & 1.92 & 0.79 & 0.27 ± 0.02 & 373.00 ± 75.28 & 2.50 ± 0.69 & 0.93 ± 0.06 & 0.97 ± 0.02 \\ \hline
        Larger spread & 0.13 & 0.81 & 0.28 ± 0.02 & 319.46 ± 84.10 & 2.56 ± 0.78 & 0.88 ± 0.07 & 0.96 ± 0.02 \\ \hline
        Larger spread, smaller minimum distance & 0.15 & 0.79 & 0.28 ± 0.03 & 281.57 ± 97.08 & 2.70 ± 0.84 & 0.80 ± 0.16 & 0.95 ± 0.04 \\ \hline
        UMAP default & 1.57 & 0.89 & 0.28 ± 0.03 & 314.90 ± 86.90 & 2.53 ± 0.50 & 0.88 ± 0.08 & 0.96 ± 0.02 \\ \hline
    \end{tabular}

    \label{table:ab}
\end{table*}

UMAP uses the family of curves \(f(x) = \frac{1}{1 + \alpha x^{2\beta}}\) to compute the low-dimensional connection probability. If x, the distance between embeddings, is close to zero, connection probability becomes closer to 1. Else, the probability becomes closer to zero. UMAP's low-dimensional connection probability calculation is dependent on $\alpha$ and $\beta$ values. Figure~\ref{fig:tails} illustrates low-dimensional connection probability with respect to distance between embeddings for different  $\alpha$ and $\beta$ combinations.

\section{Effects of DBN}
\xhdr{Decorrelation Batch Normalization (DBN)}

\begin{figure}[h!]
    \centering
    \begin{tabular}{|p{4cm}|p{4cm}|}
        \hline
        With DBN & Without DBN \\
        \hline
        \includegraphics[width=4cm, height=4cm, trim={1.4cm 1.25cm 1.2cm 1.35cm}, clip]{images/final_swissroll/GNUMAP2.png} &
        \includegraphics[width=4cm, height=4cm, trim={1.4cm 1.25cm 1.2cm 1.35cm}, clip]{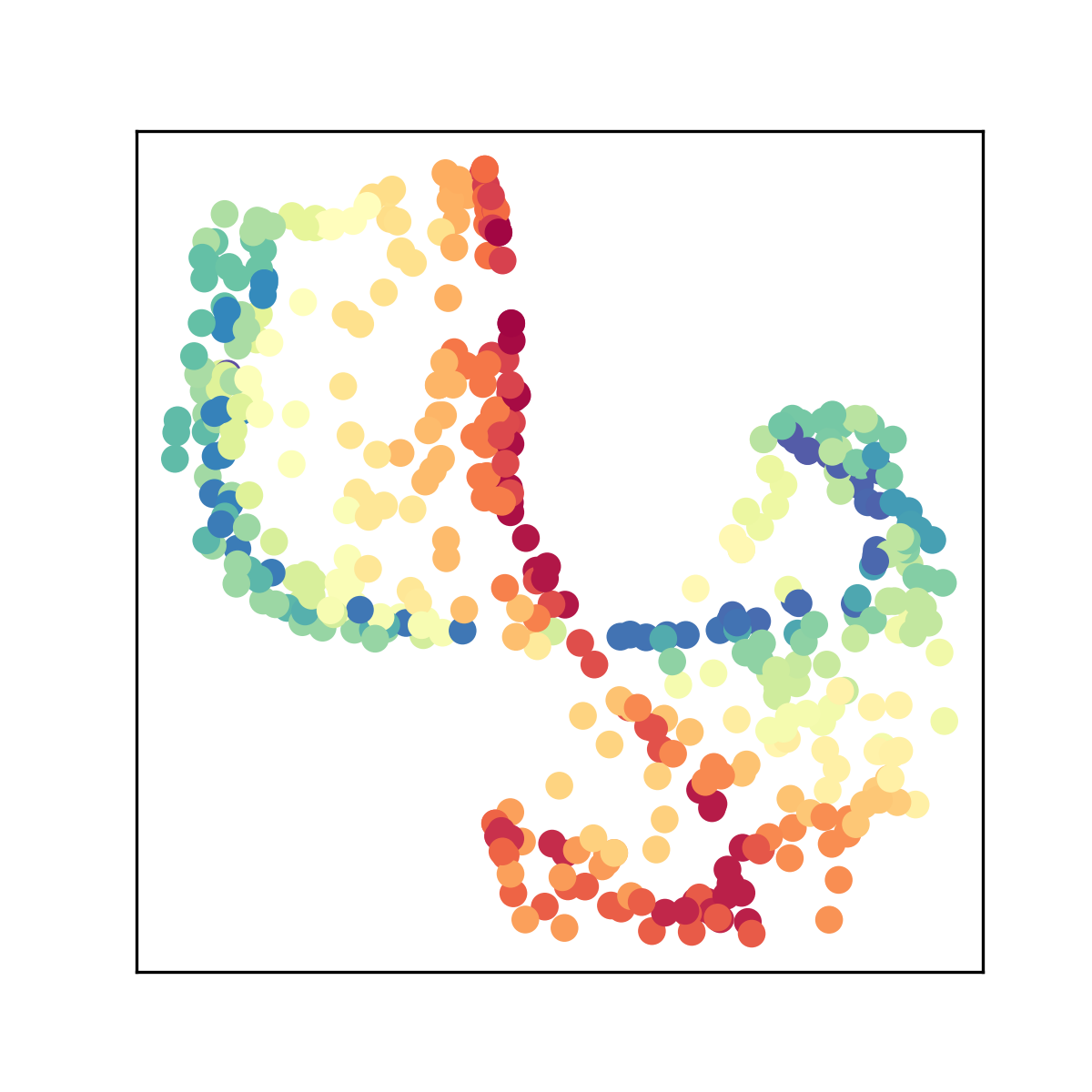} \\
        \hline
    \end{tabular}
    \caption{GNUMAP synthetic Swissroll embeddings with and without DBN layer.}
    \label{fig:dbn}
\end{figure}

DBN layer performs batch whitening, which removes linear correlation between input channels and, therefore, stabilizes GNN model convergence. DBN layer is effective in unrolling the higher-dimensional graph object into 2-dimensions, which can be confirmed both by visual inspection in Figure~\ref{fig:dbn} and by metric evaluation in Table~\ref{table:dbnchoice}.

\begin{table}[h!]
    \centering
        \caption{Evaluated mean and standard deviation of proposed metrics for GNUMAP alterations for 100 synthetic Swissroll experiments. The upward arrows next to the metric name denote a better performance with higher metrics, whereas the downward arrows denote a better performance with lower metrics.}
        \vspace{1em}
\begin{tabular}{|p{2.3cm}|p{2.5cm}|p{2cm}|p{2cm}|p{2cm}|p{2.2cm}|}
\hline
\rowcolor[HTML]{C0C0C0} 
Architecture Choice & Calinski Harabasz Score ↑ & Davies Bouldin Score ↓ & Spearman Correlation w/ Original Graph ↑ & Overlap \% of 50 Neighbours ↑ & Classification Accuracy ↑ \\
Without DBN         & 305.39 ± 73.33            & 2.74 ± 0.76            & 0.87 ± 0.11                              & 0.95 ± 0.03                   & 0.26 ± 0.03               \\
With DBN            & \textbf{314.90 ± 86.90}   & \textbf{2.53 ± 0.50}   & \textbf{0.88 ± 0.08}                     & \textbf{0.96 ± 0.02}          & \textbf{0.28 ± 0.03}  \\ \hline
\end{tabular}
    \label{table:dbnchoice}
\end{table}

\newpage
\section{Additional Figures}
\renewcommand{\arraystretch}{2}
\begin{figure*}[h!]
    \centering
    \begin{tabular}{|p{1.5cm}|c|c|c|c|c|c|c|c|c|c|}
        \hline
        Method & Blobs & Swissroll & Circles & Moons \\
        \hline
        Ground Truth &
        \includegraphics[height=1.7cm, trim={1.4cm 1.25cm 1.2cm 1.35cm},clip]{images/final_blobs/manifold_Blobs.png} &
        \includegraphics[height=1.7cm, trim={1.4cm 1.25cm 1.2cm 1.35cm},clip]{images/final_swissroll/manifold_Swissroll.png} &
        \includegraphics[height=1.7cm, trim={1.4cm 1.25cm 1.2cm 1.35cm},clip]{images/final_circles/manifold_Circles.png} &
        \includegraphics[height=1.7cm, trim={1.4cm 1.25cm 1.2cm 1.35cm},clip]{images/final_moons/manifold_Moons.png} \\
        \hline
        DGI &
        \includegraphics[height=1.7cm, trim={1.4cm 1.25cm 1.2cm 1.35cm},clip]{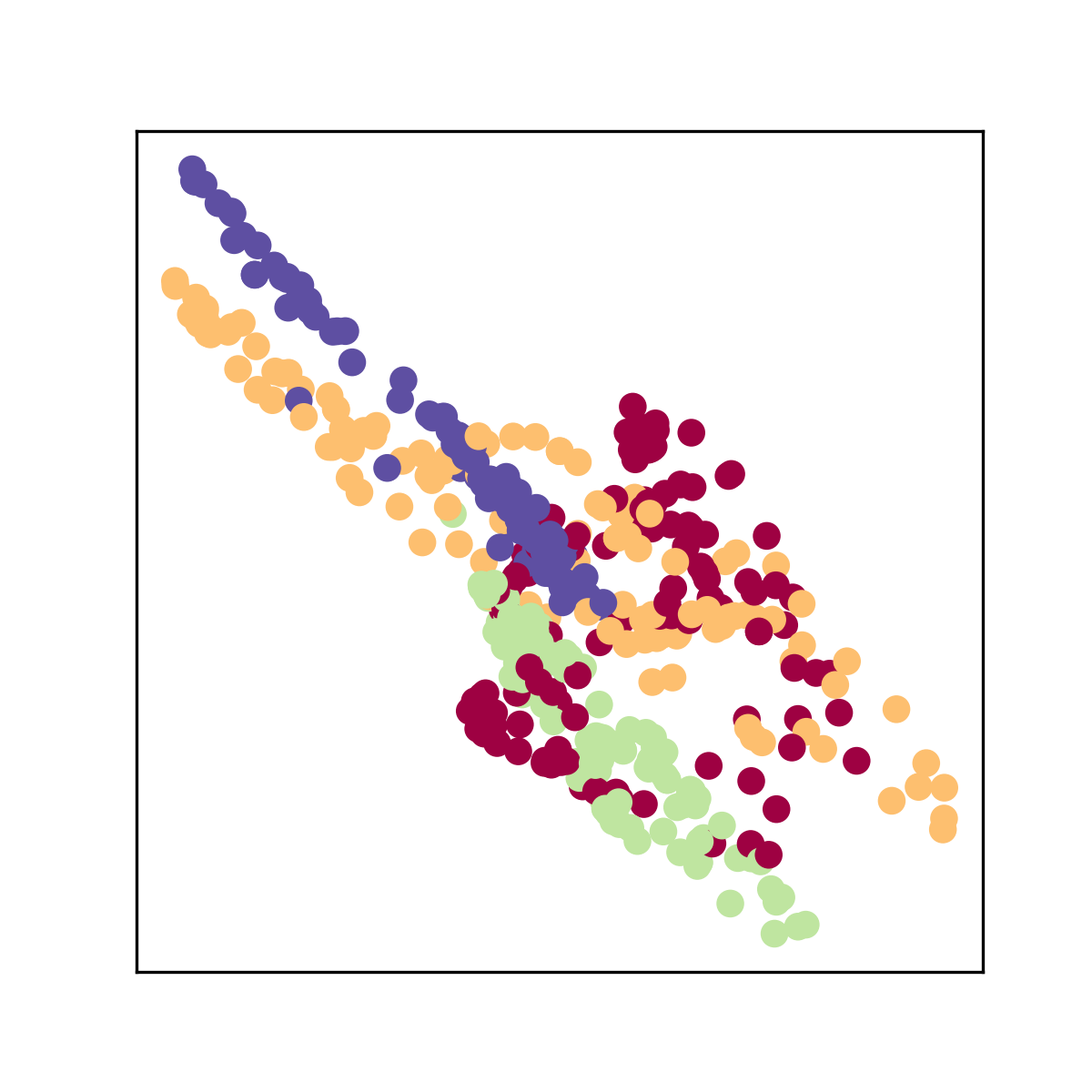} &
        \includegraphics[height=1.7cm, trim={1.4cm 1.25cm 1.2cm 1.35cm},clip]{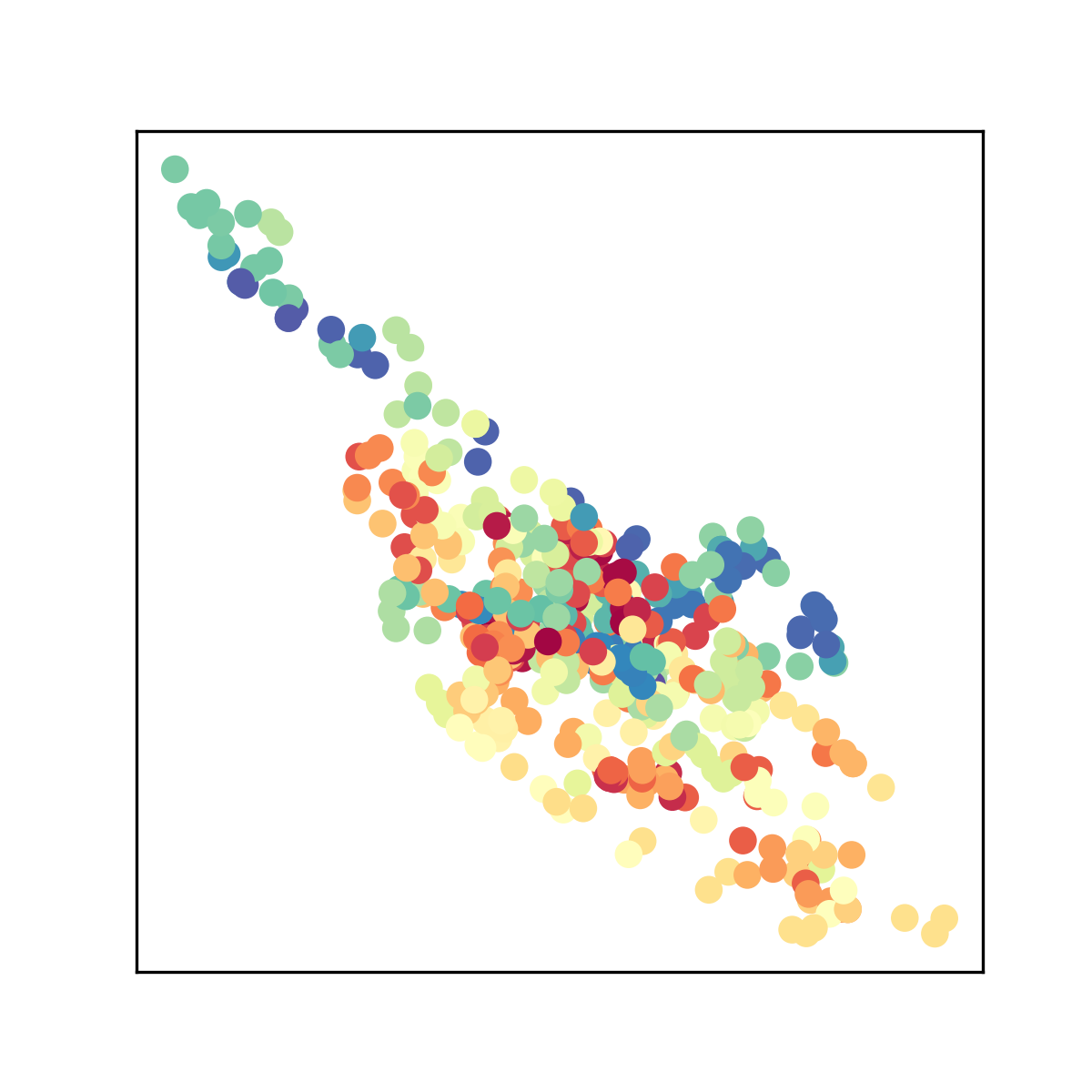} &
        \includegraphics[height=1.7cm, trim={1.4cm 1.25cm 1.2cm 1.35cm},clip]{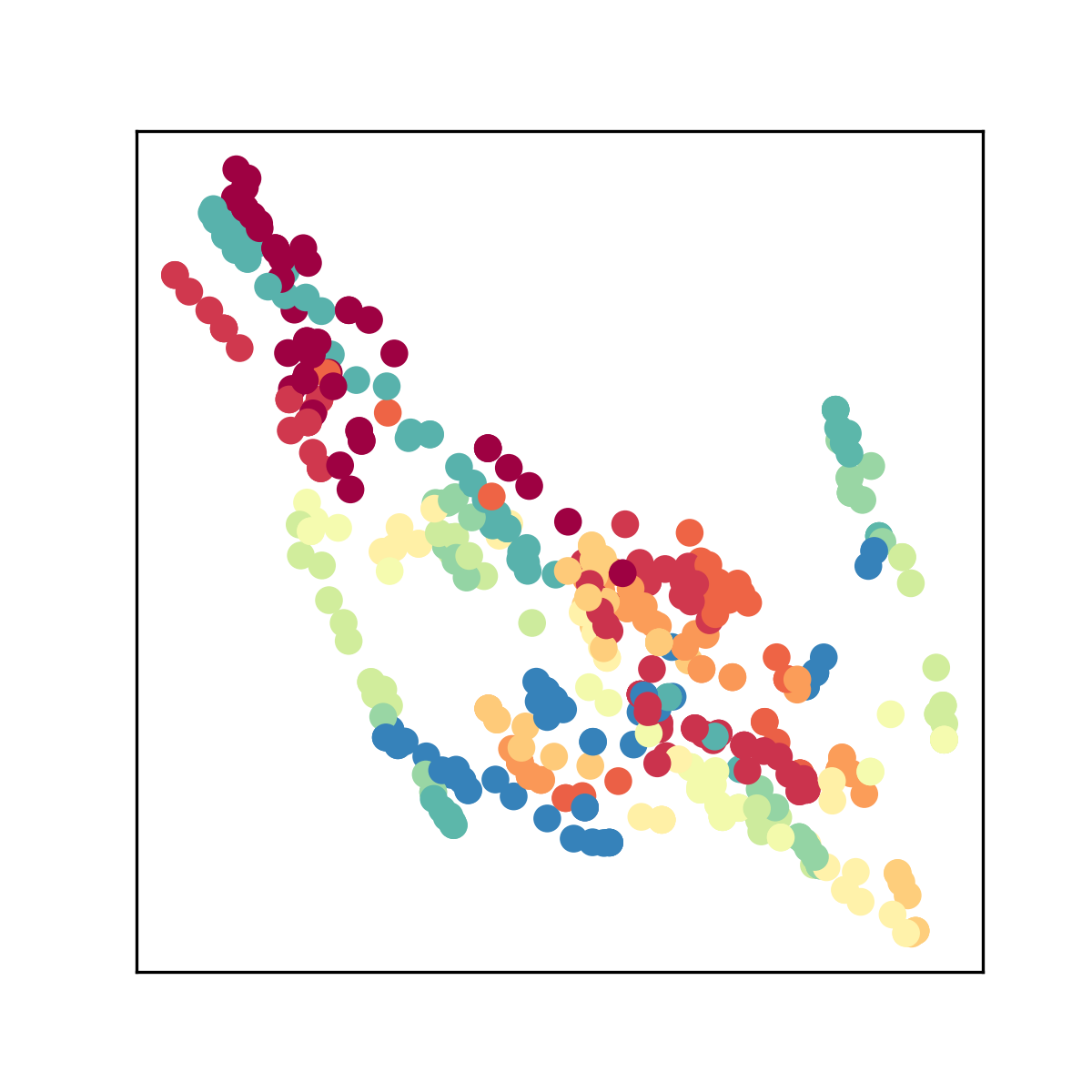} &
        \includegraphics[height=1.7cm, trim={1.4cm 1.25cm 1.2cm 1.35cm},clip]{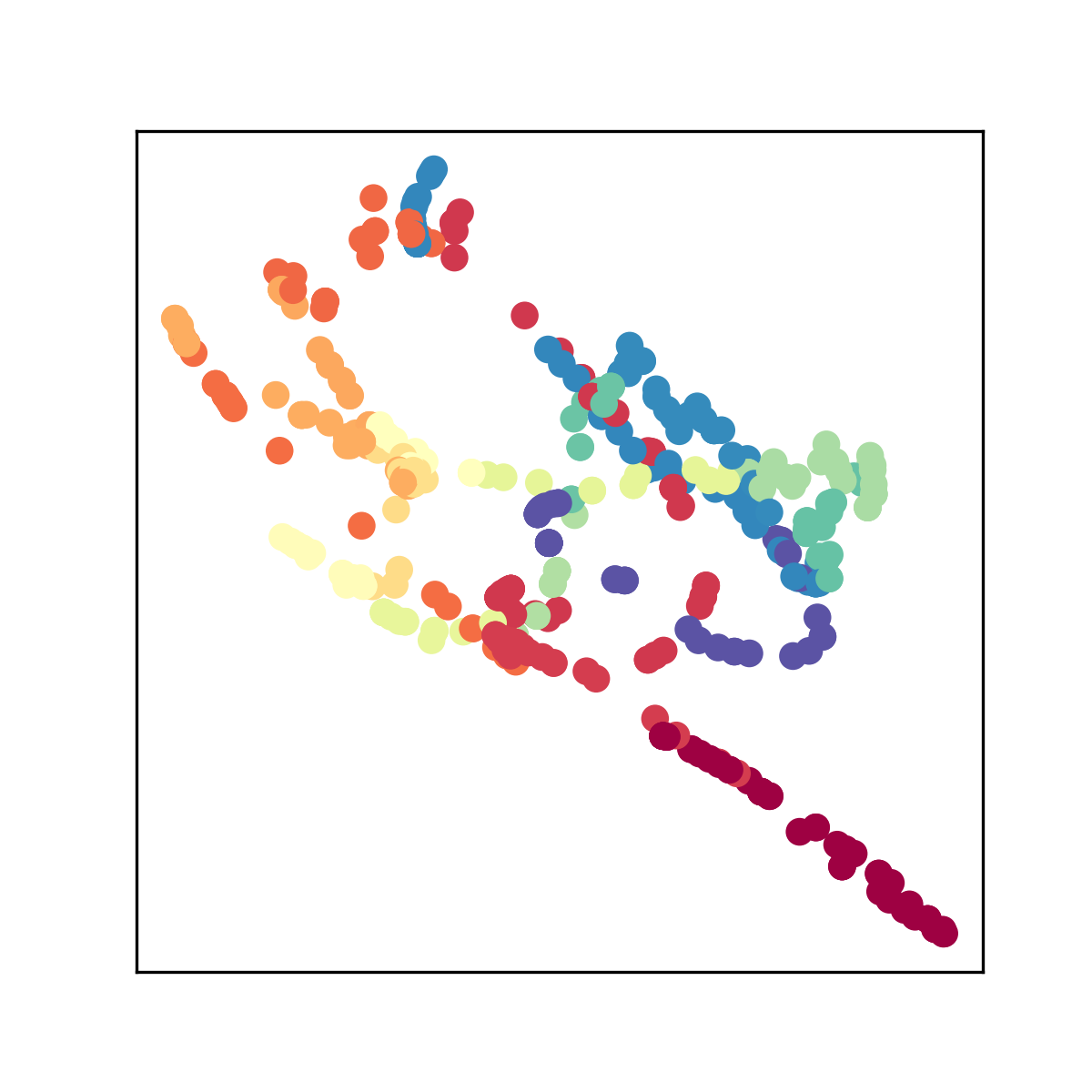} \\
        \hline
        BGRL &
        \includegraphics[height=1.7cm, trim={1.4cm 1.25cm 1.2cm 1.35cm},clip]{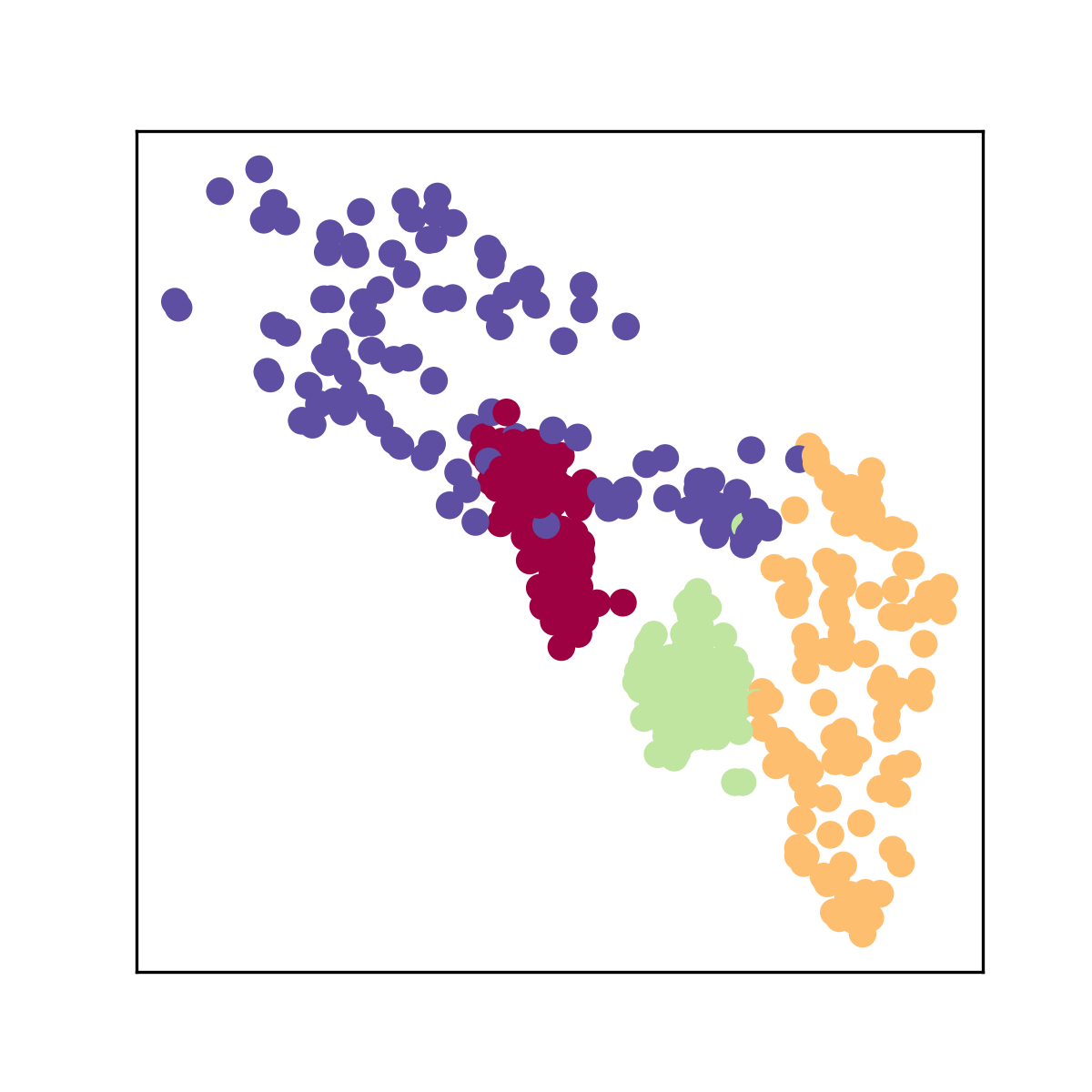} &
        \includegraphics[height=1.7cm, trim={1.4cm 1.25cm 1.2cm 1.35cm},clip]{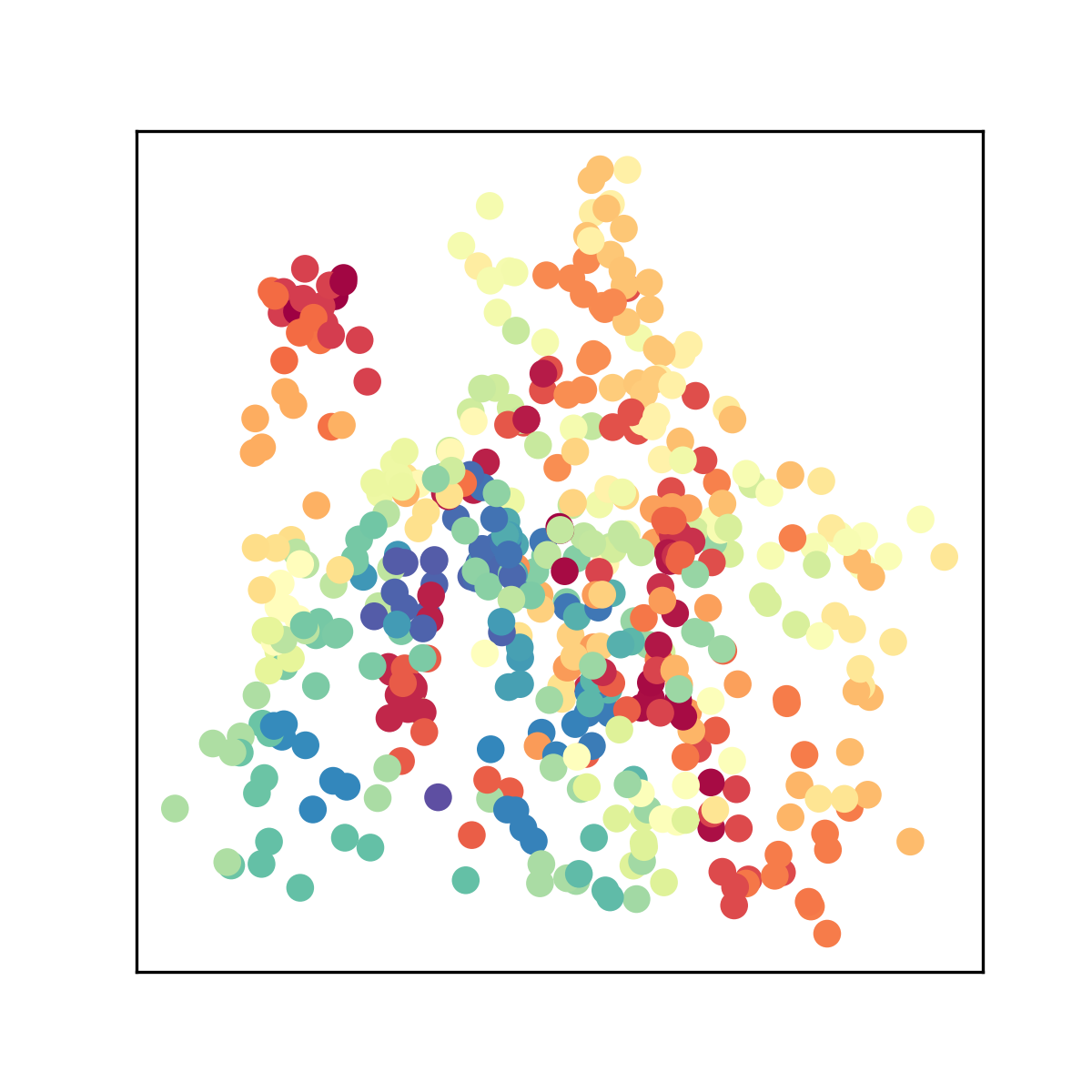} &
        \includegraphics[height=1.7cm, trim={1.4cm 1.25cm 1.2cm 1.35cm},clip]{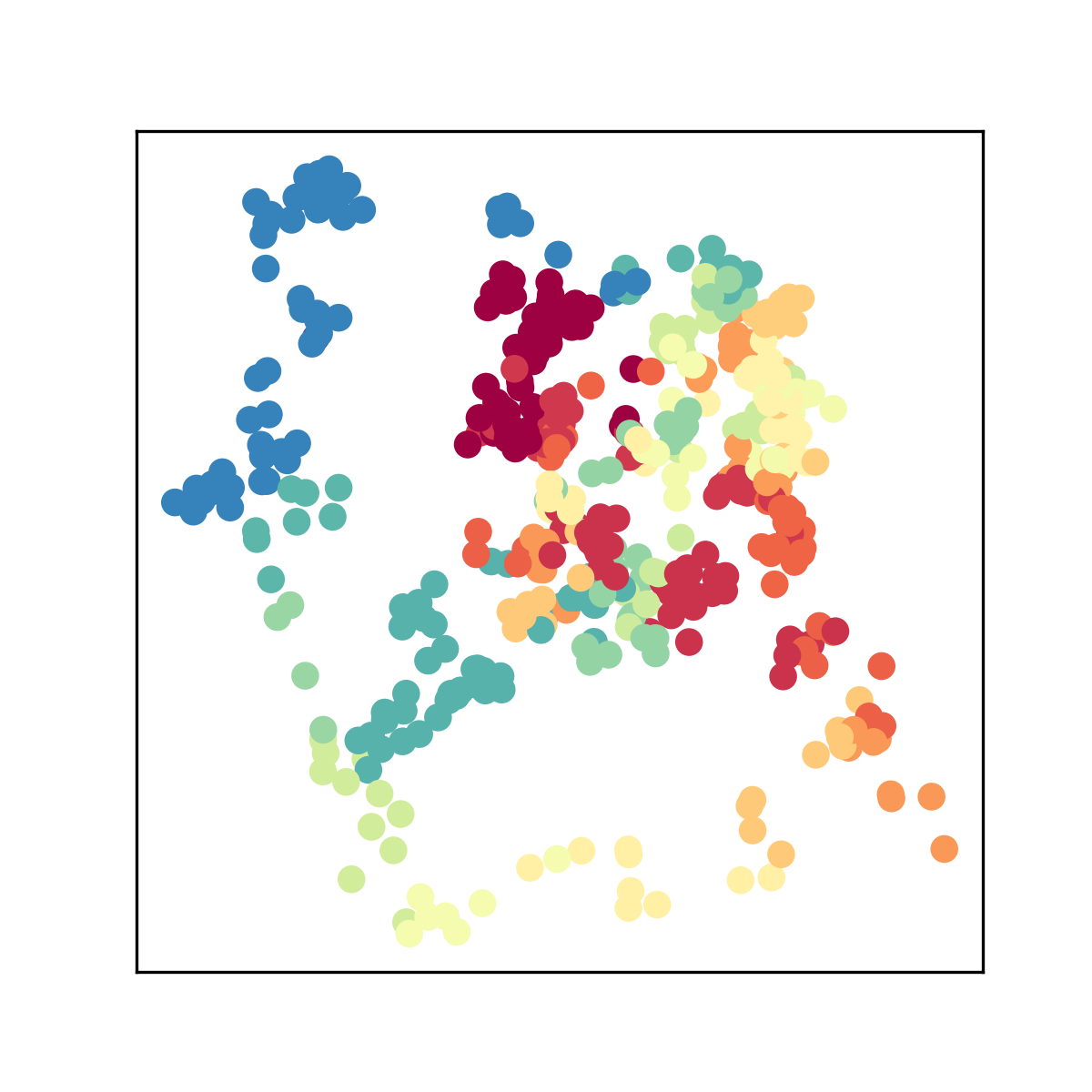} &
        \includegraphics[height=1.7cm, trim={1.4cm 1.25cm 1.2cm 1.35cm},clip]{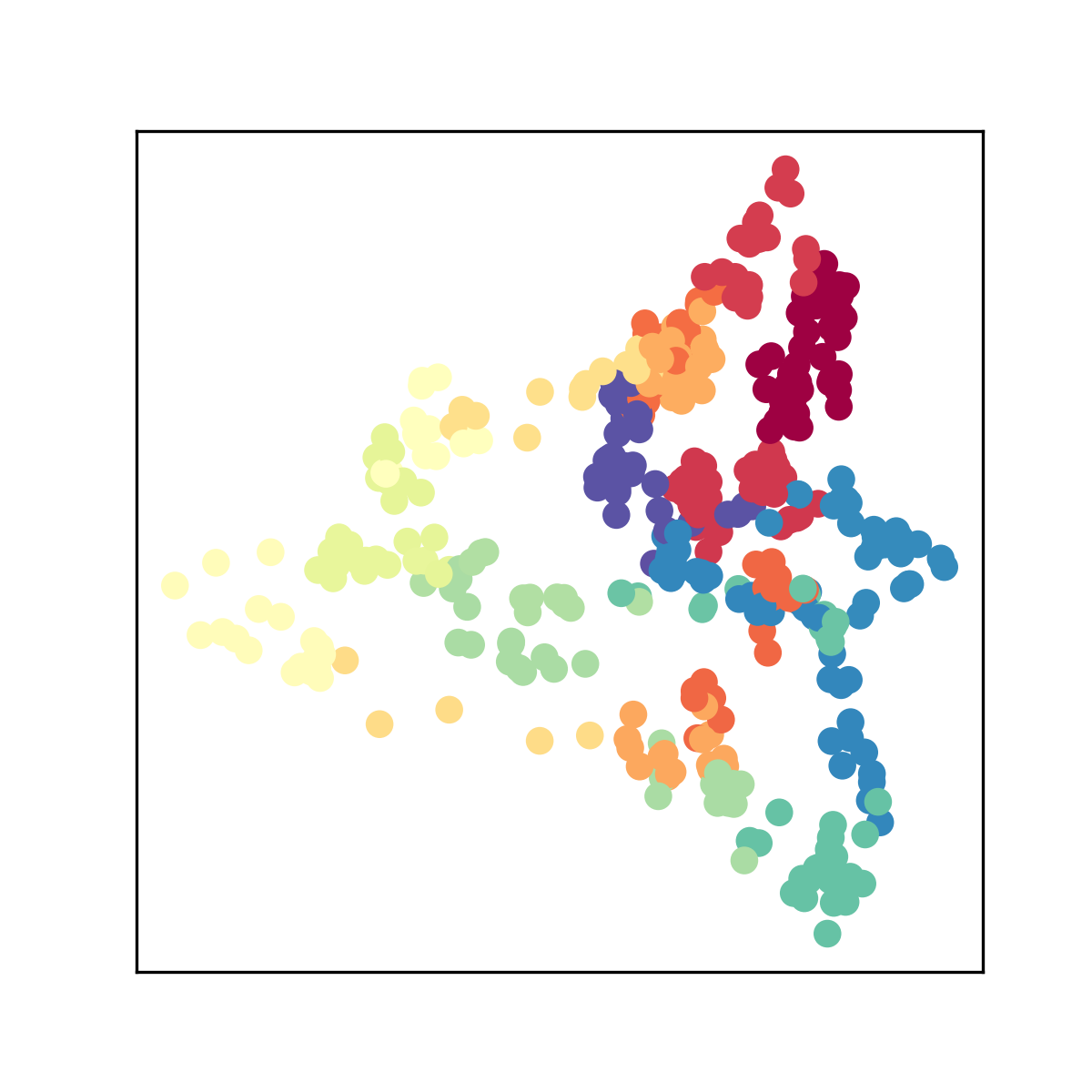} \\
        \hline
        VGAE &
        \includegraphics[height=1.7cm, trim={1.4cm 1.25cm 1.2cm 1.35cm},clip]{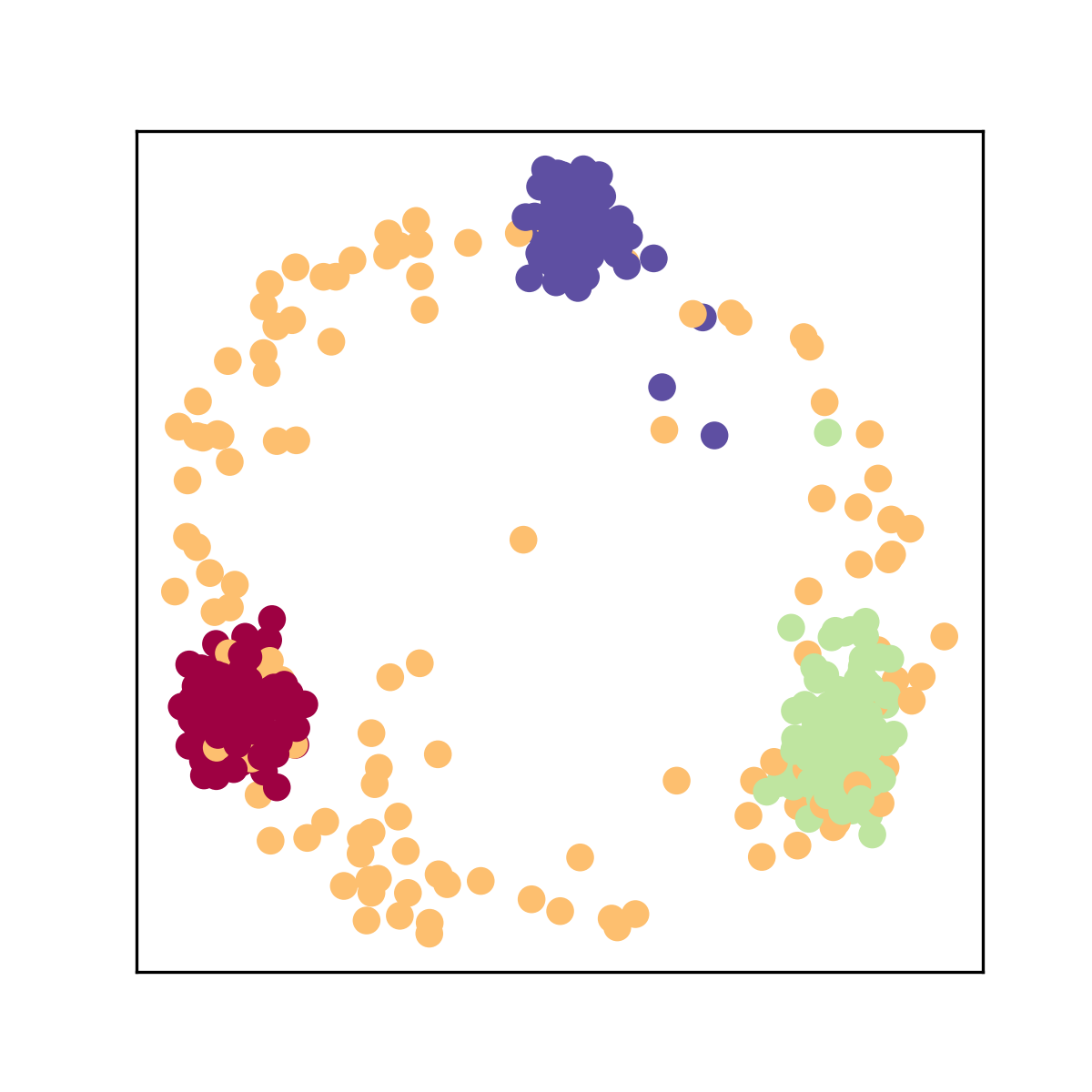} &
        \includegraphics[height=1.7cm, trim={1.4cm 1.25cm 1.2cm 1.35cm},clip]{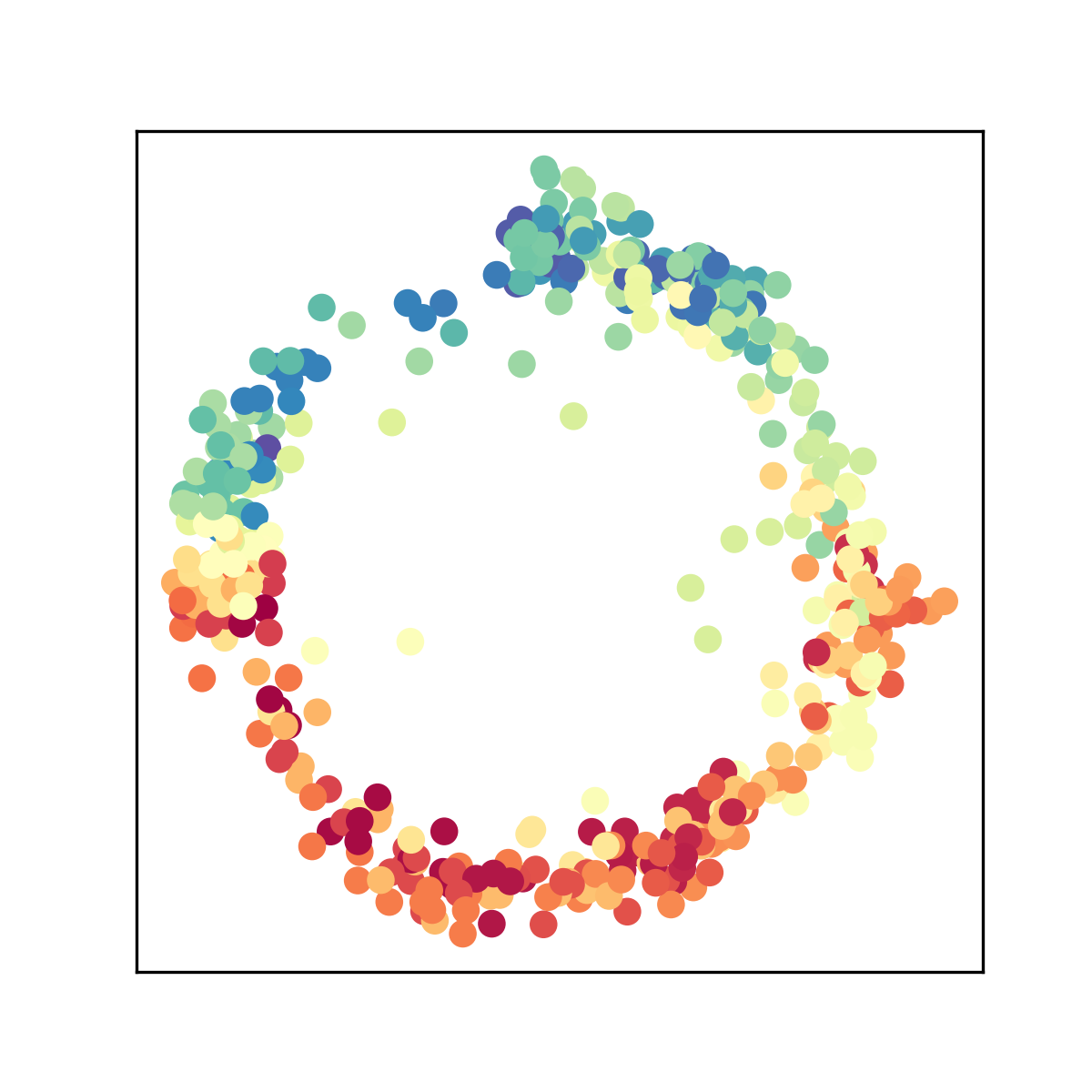} &
        \includegraphics[height=1.7cm, trim={1.4cm 1.25cm 1.2cm 1.35cm},clip]{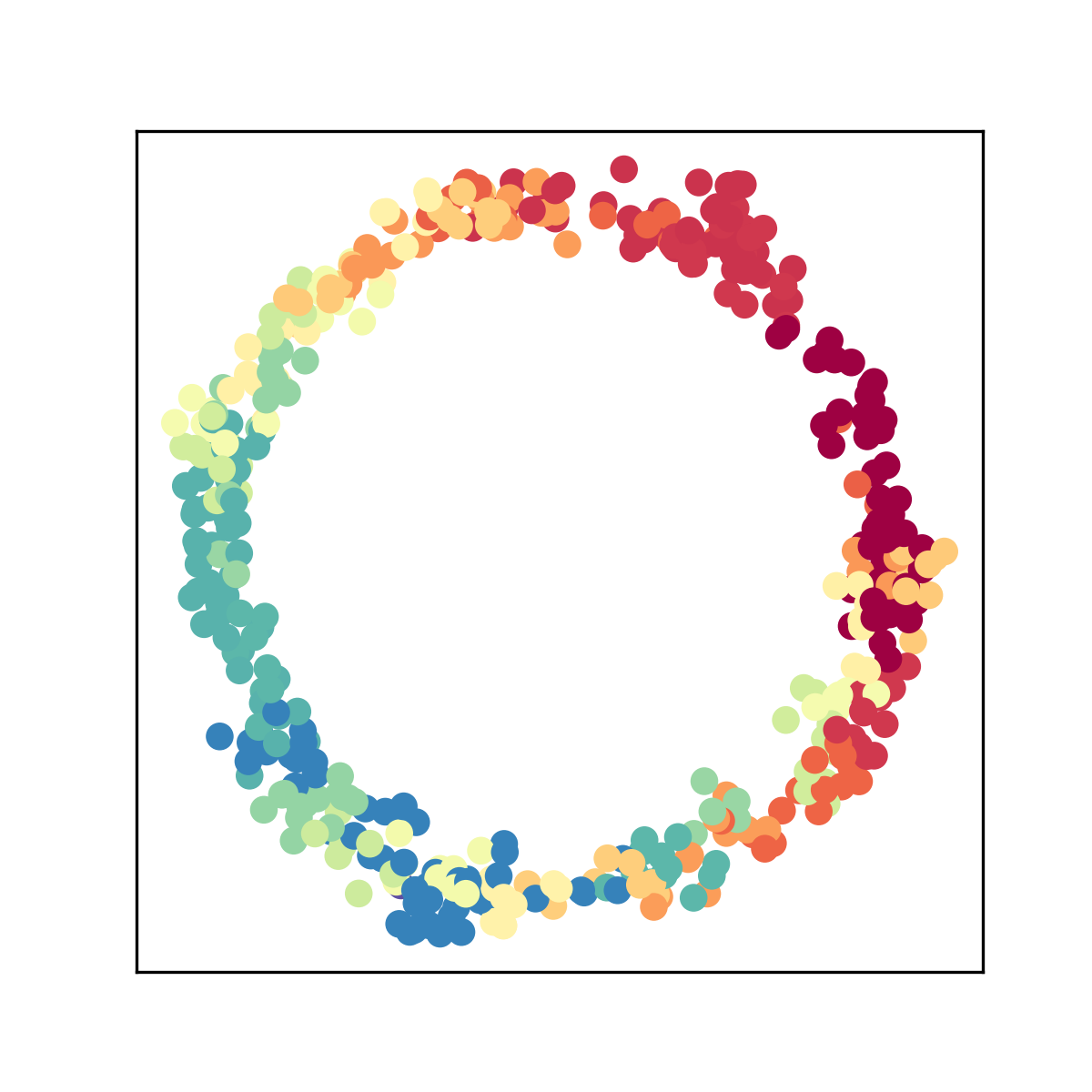} &
        \includegraphics[height=1.7cm, trim={1.4cm 1.25cm 1.2cm 1.35cm},clip]{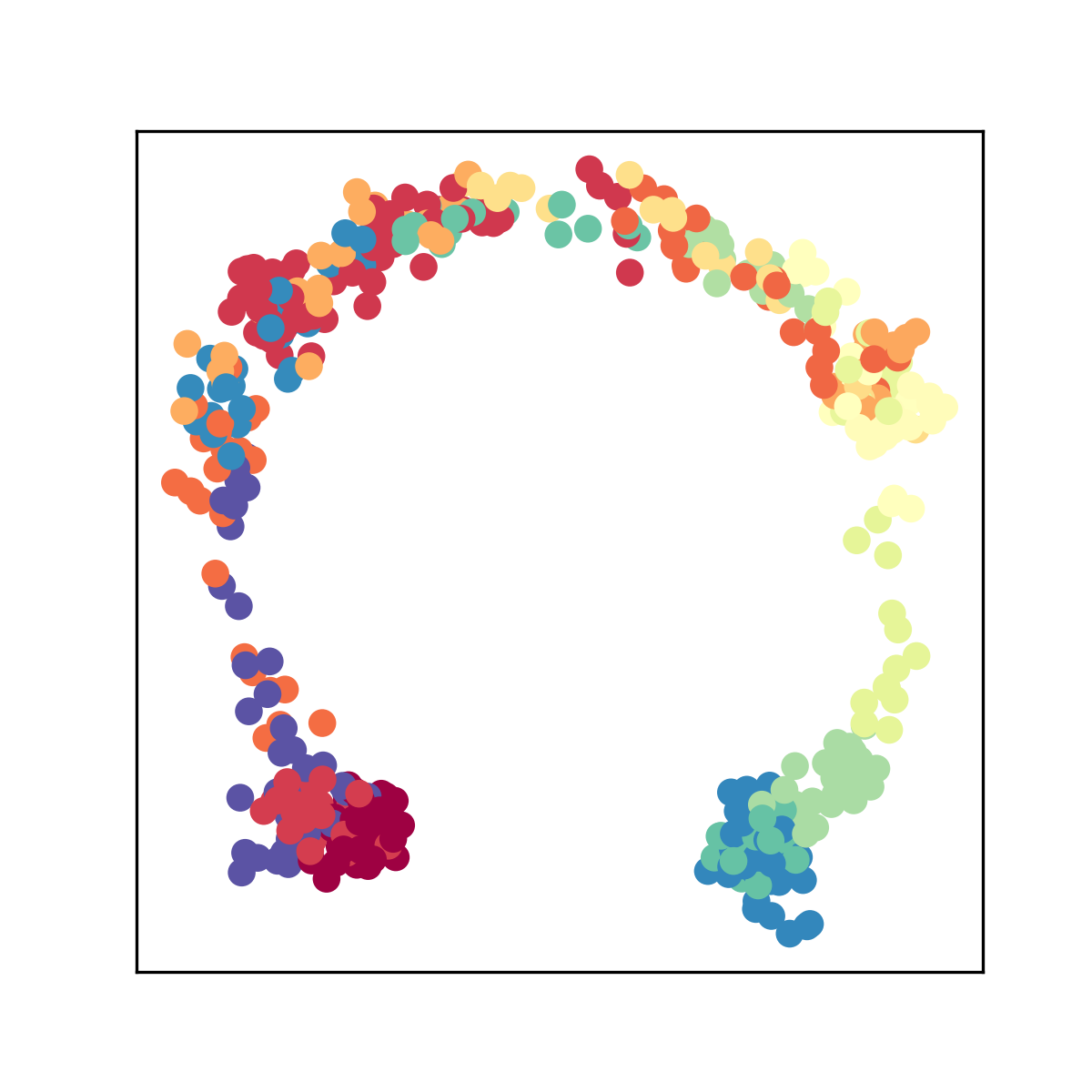} \\
        \hline
        PCA &
        \includegraphics[height=1.7cm, trim={1.4cm 1.25cm 1.2cm 1.35cm},clip]{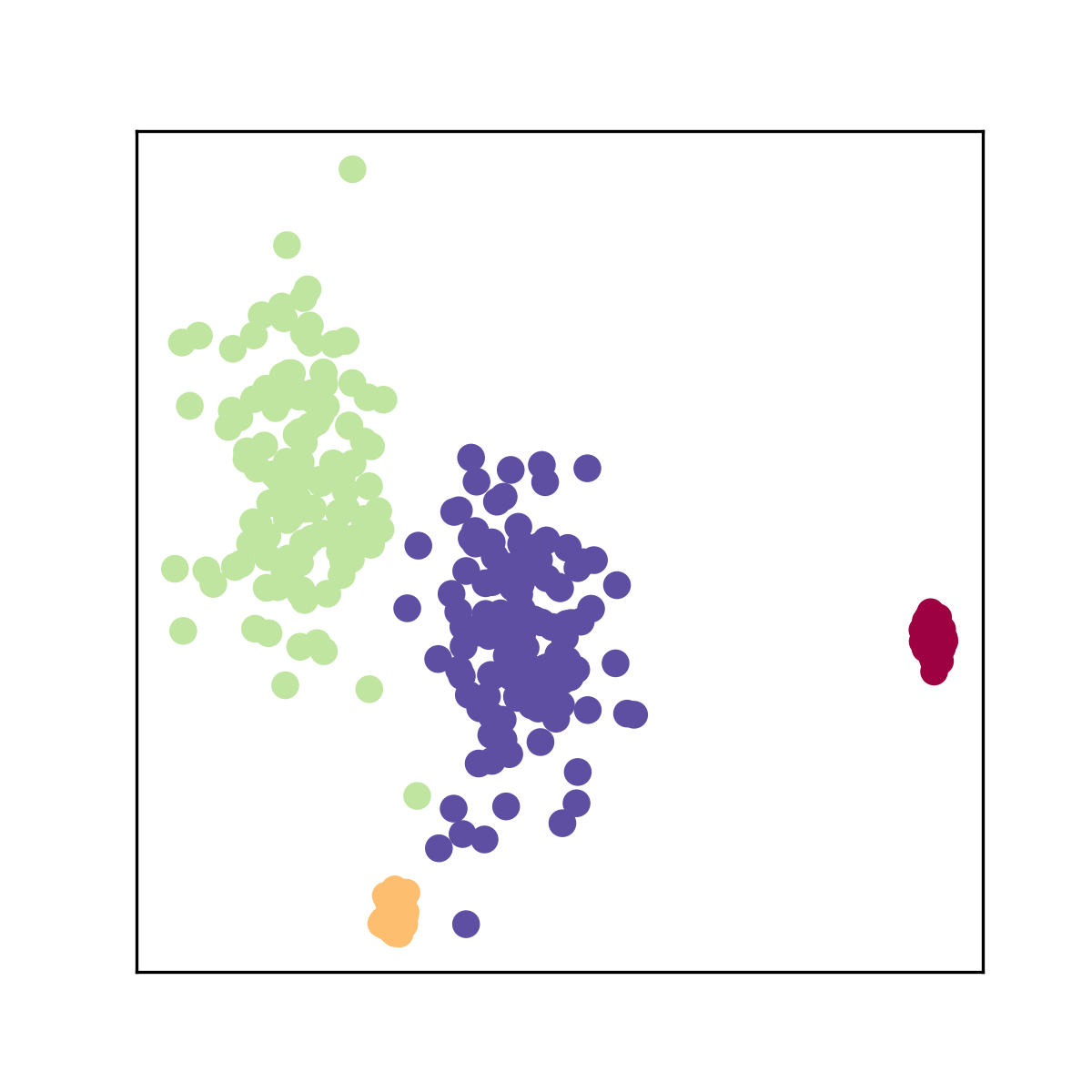} &
        \includegraphics[height=1.7cm, trim={1.4cm 1.25cm 1.2cm 1.35cm},clip]{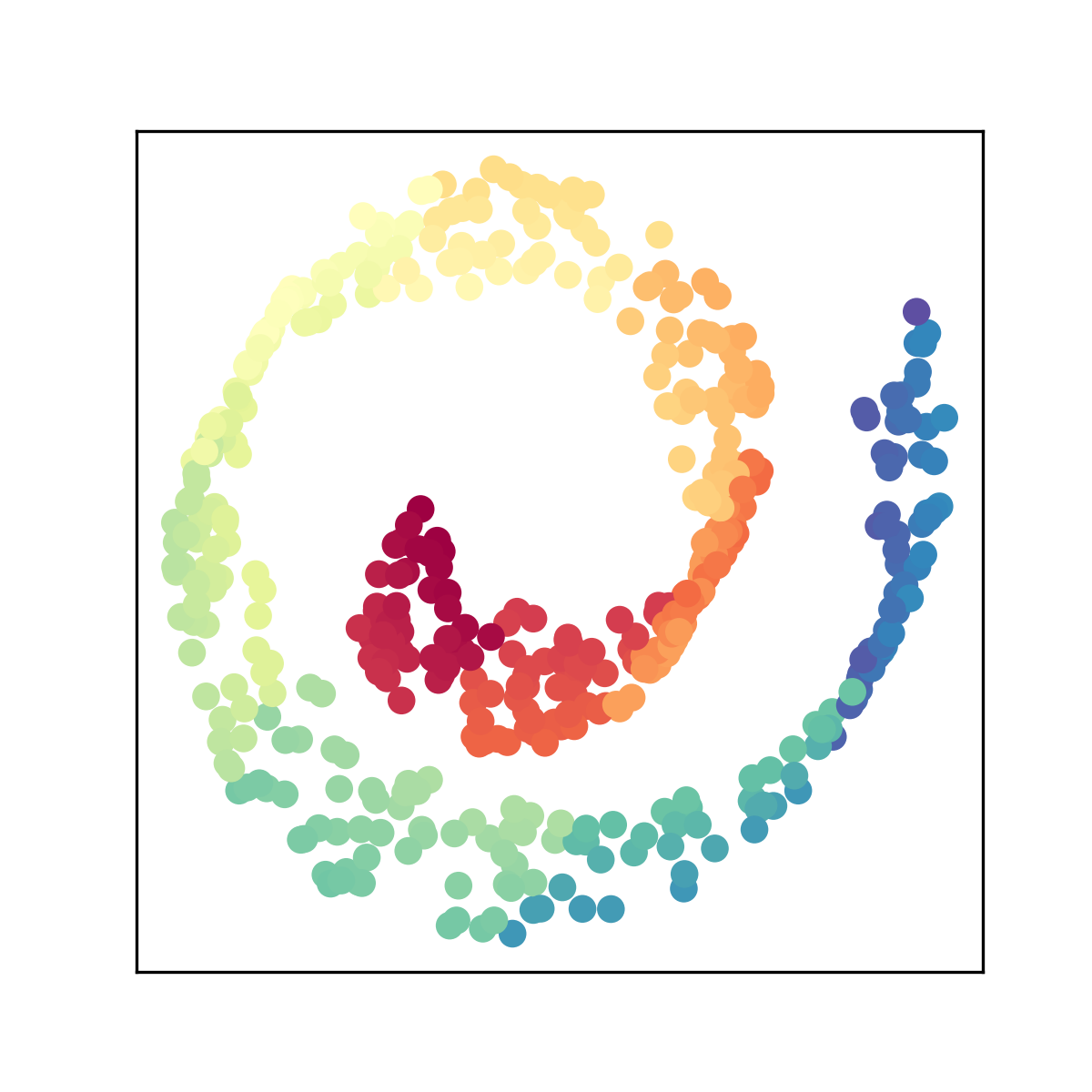} &
        \includegraphics[height=1.7cm, trim={1.4cm 1.25cm 1.2cm 1.35cm},clip]{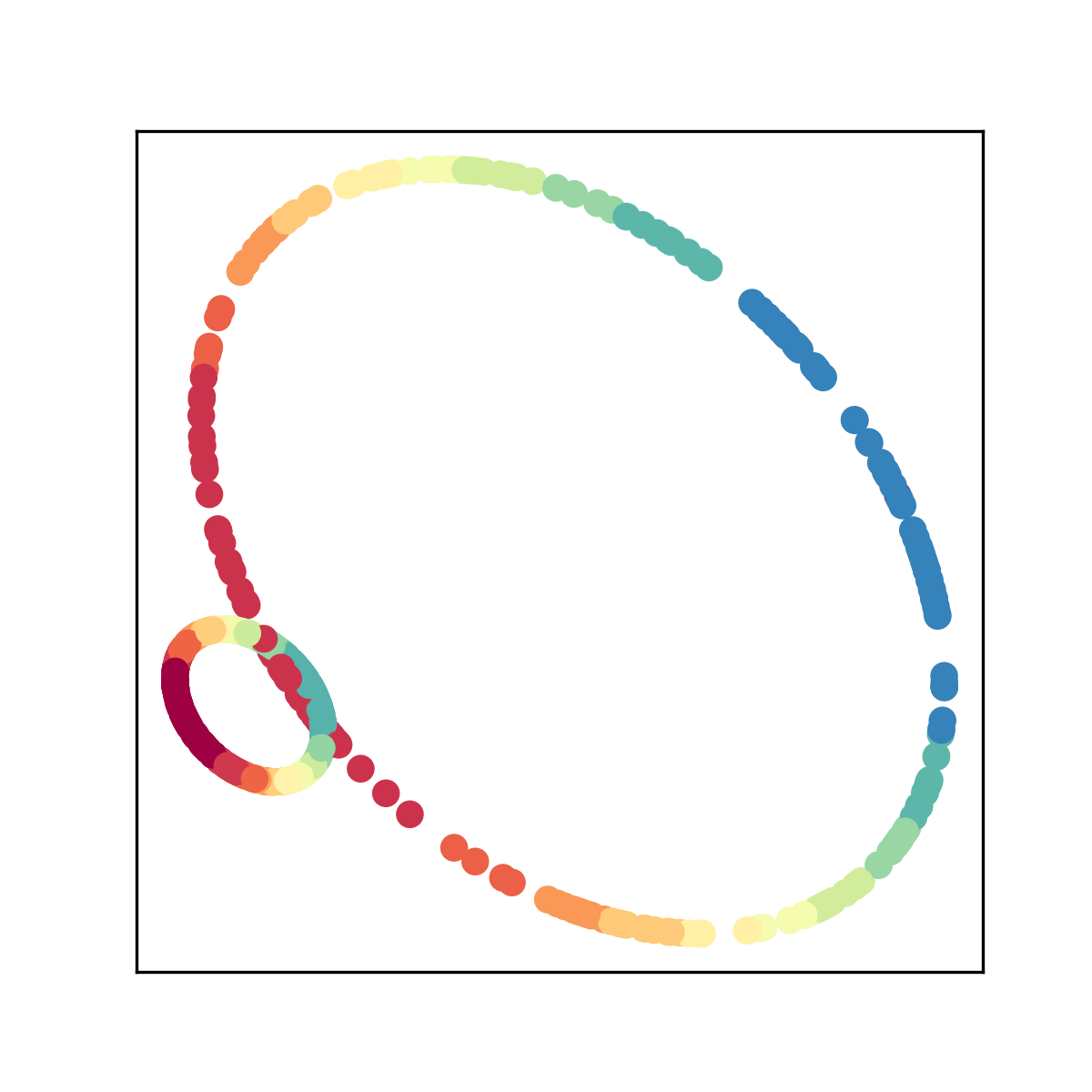} &
        \includegraphics[height=1.7cm, trim={1.4cm 1.25cm 1.2cm 1.35cm},clip]{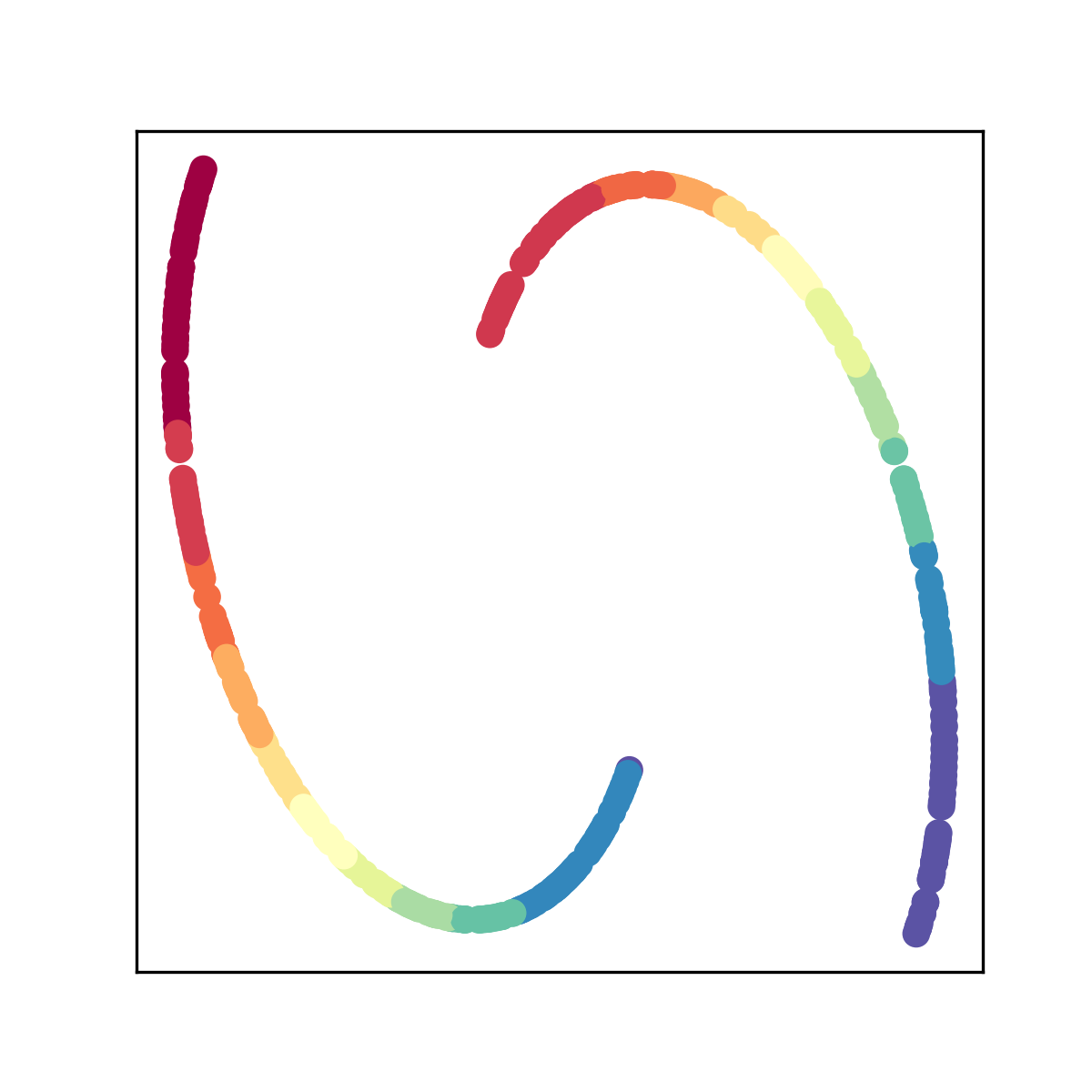} \\
        \hline
        t-SNE &
        \includegraphics[height=1.7cm, trim={1.4cm 1.25cm 1.2cm 1.35cm},clip]{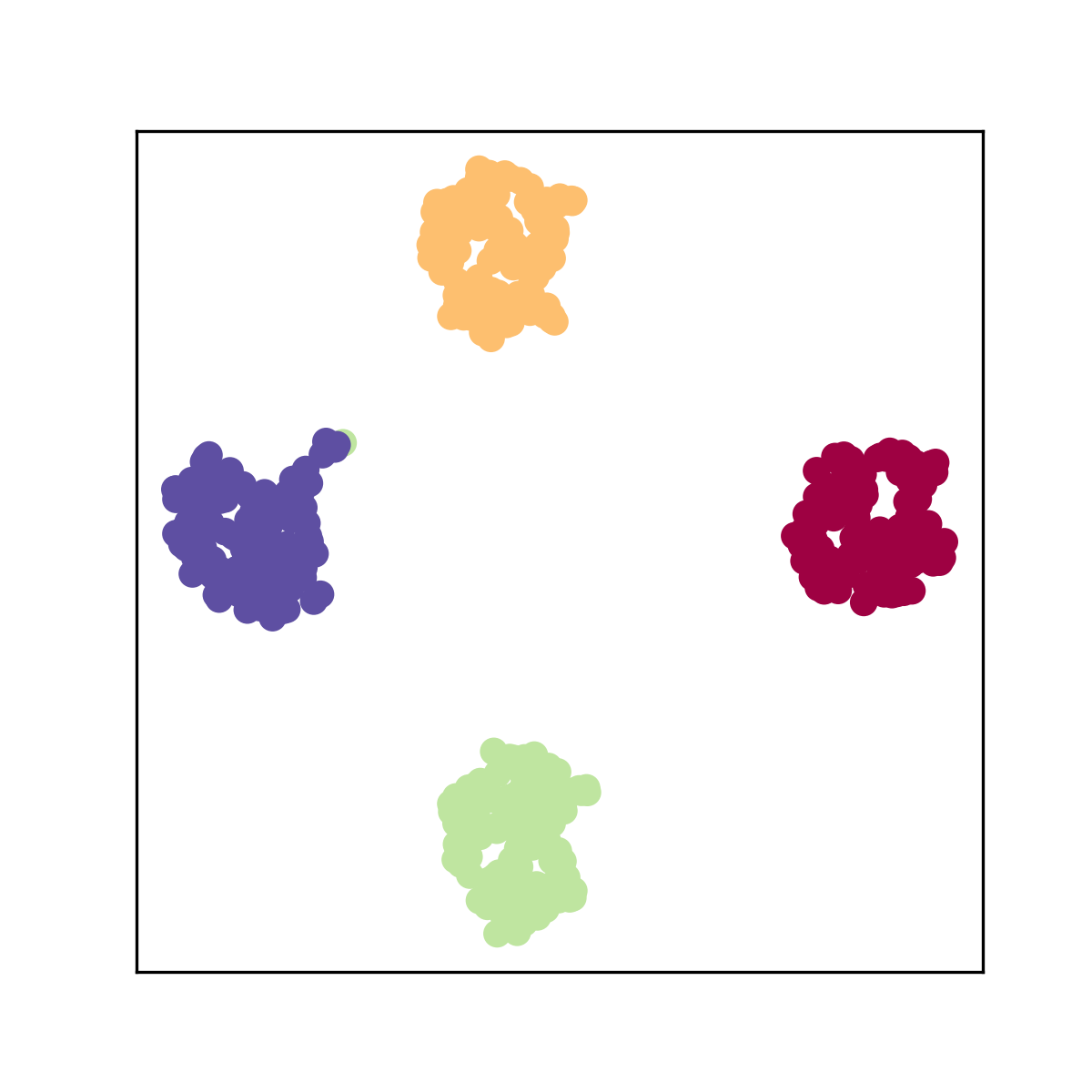} &
        \includegraphics[height=1.7cm, trim={1.4cm 1.25cm 1.2cm 1.35cm},clip]{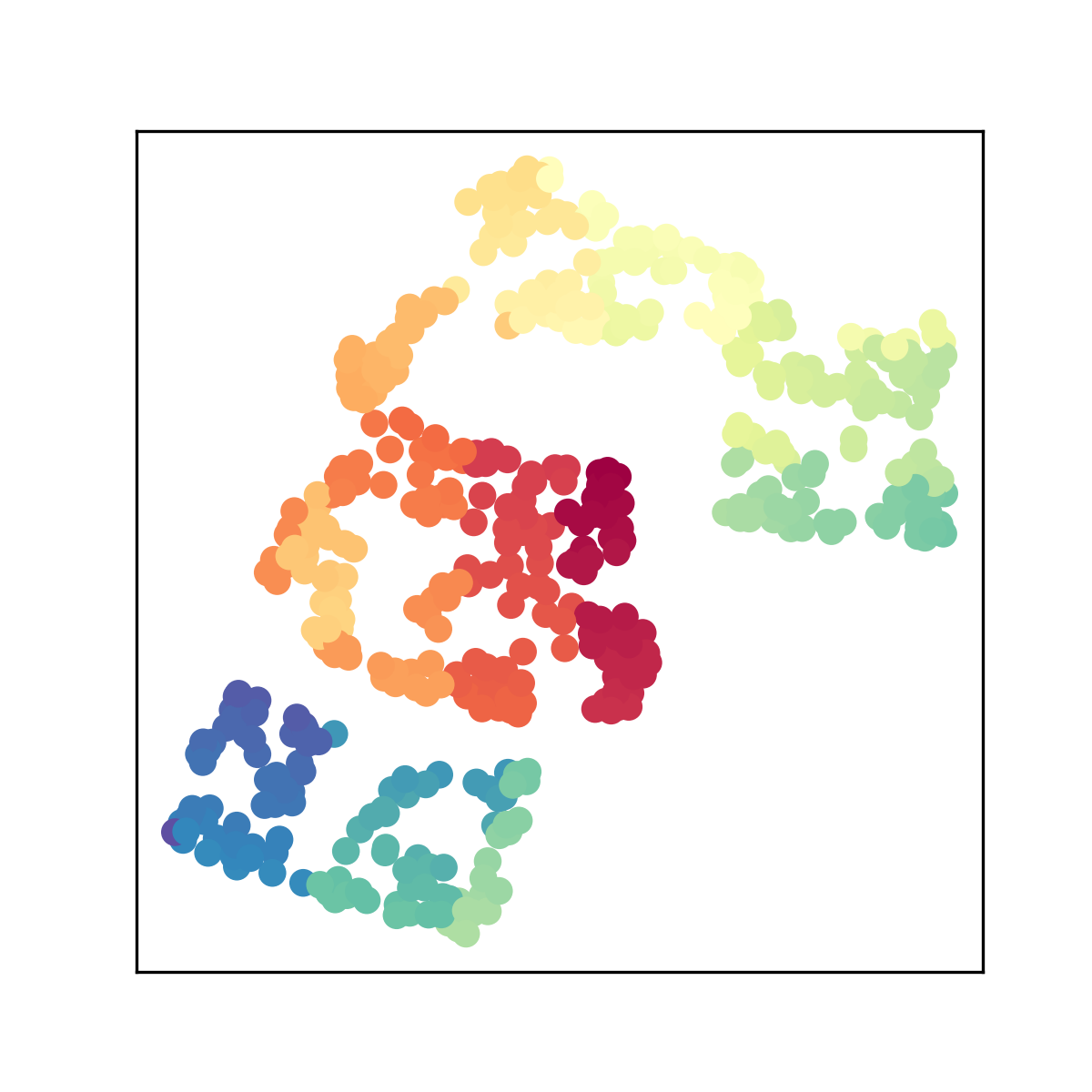} &
        \includegraphics[height=1.7cm, trim={1.4cm 1.25cm 1.2cm 1.35cm},clip]{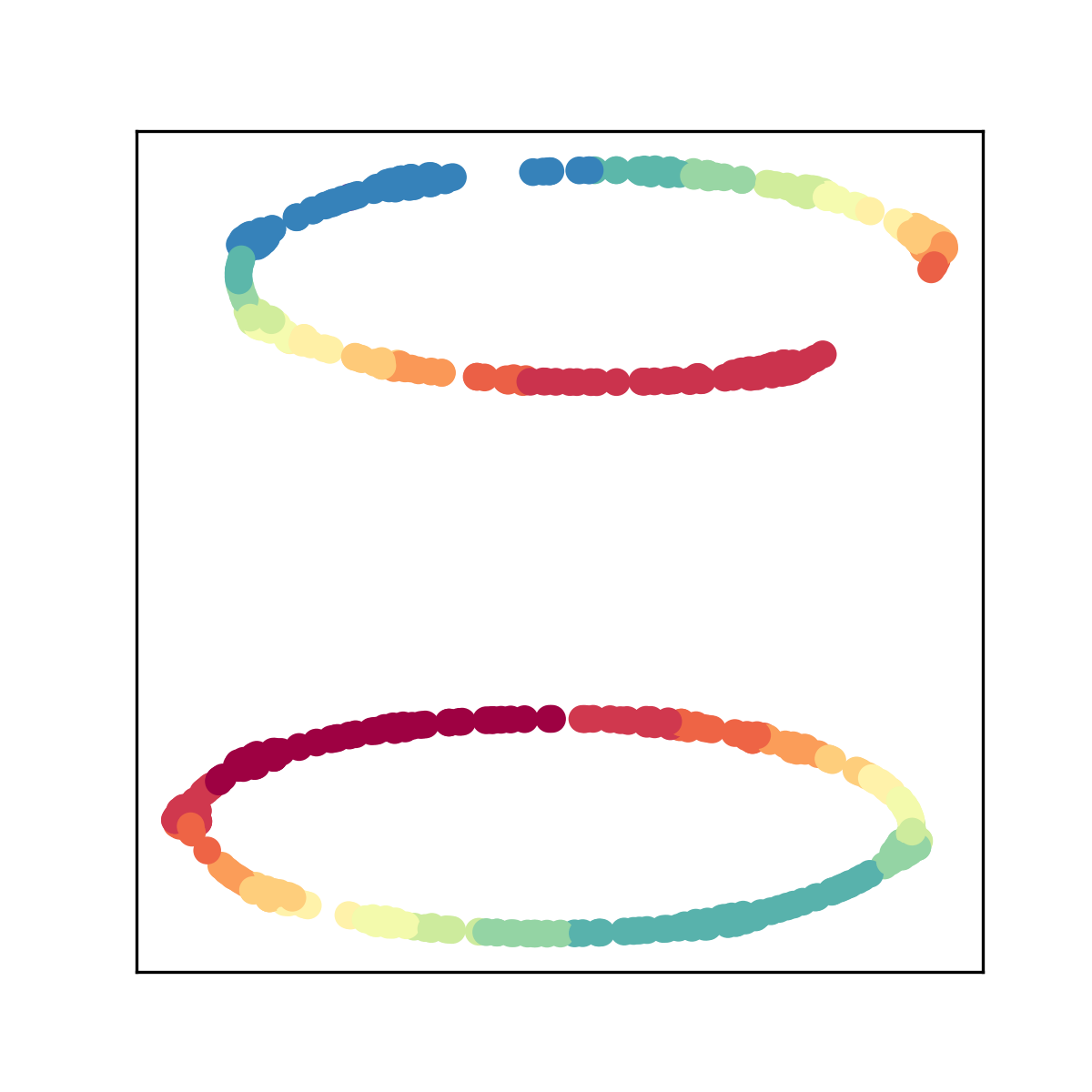} &
        \includegraphics[height=1.7cm, trim={1.4cm 1.25cm 1.2cm 1.35cm},clip]{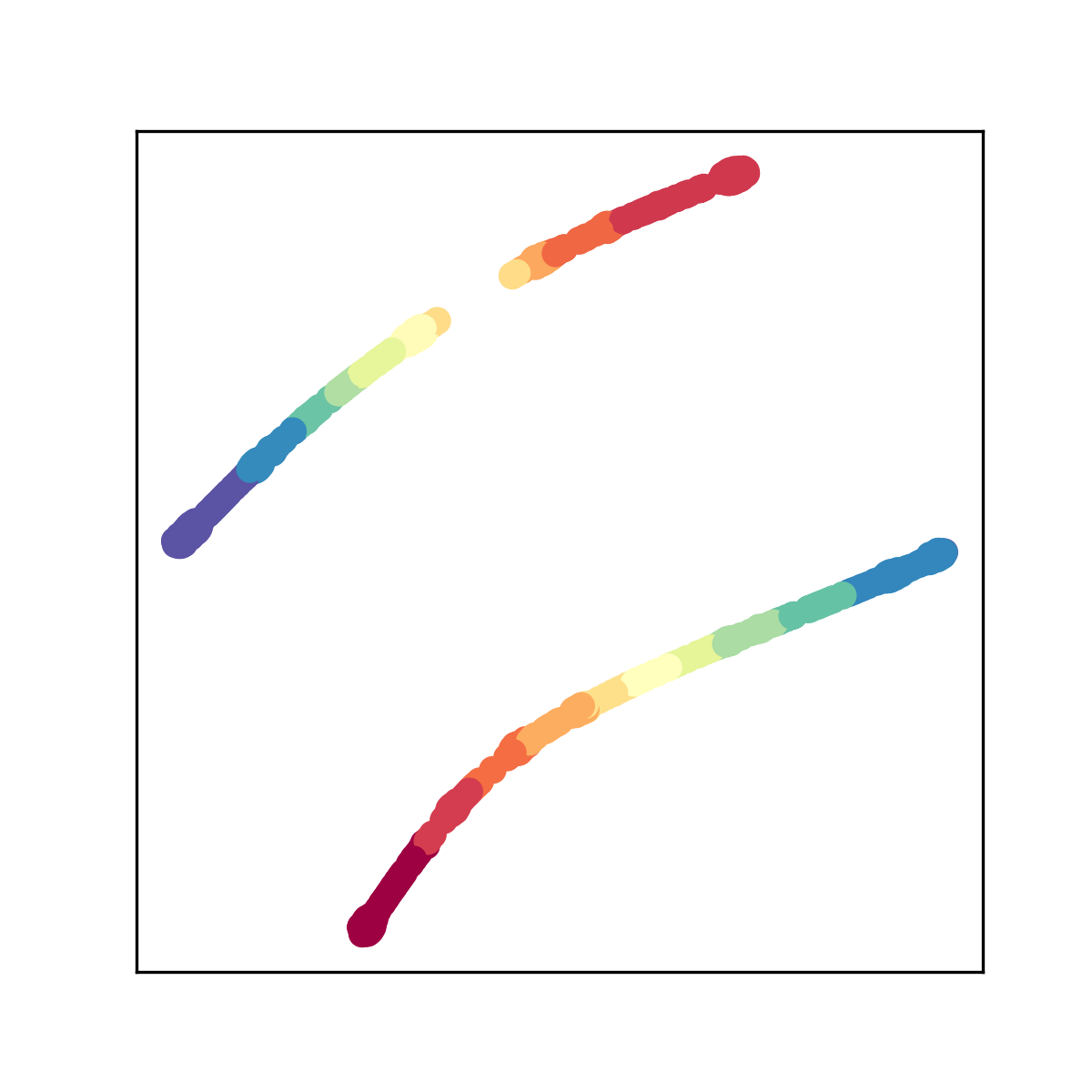} \\
        \hline
        Isomap &
        \includegraphics[height=1.7cm, trim={1.4cm 1.25cm 1.2cm 1.35cm},clip]{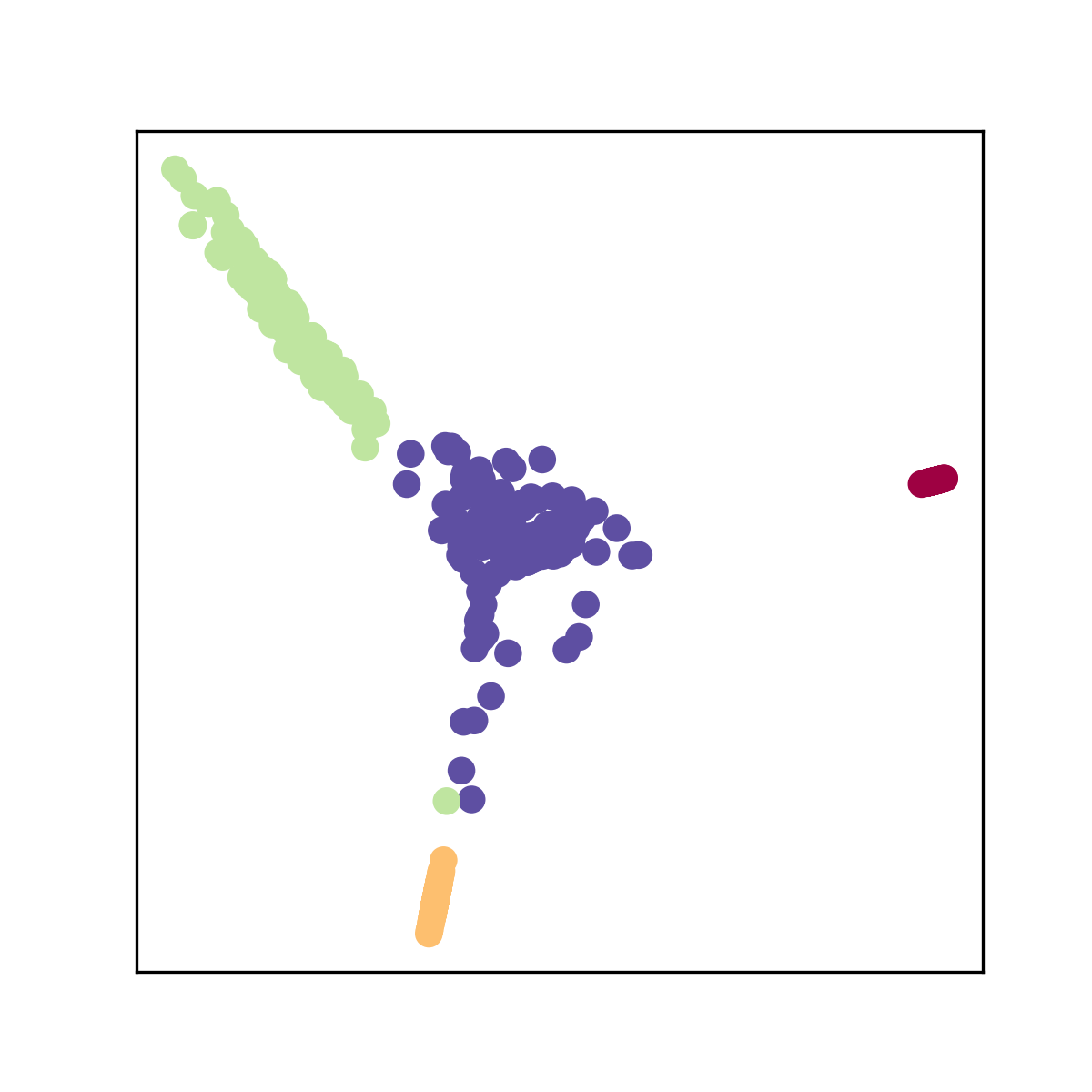} &
        \includegraphics[height=1.7cm, trim={1.4cm 1.25cm 1.2cm 1.35cm},clip]{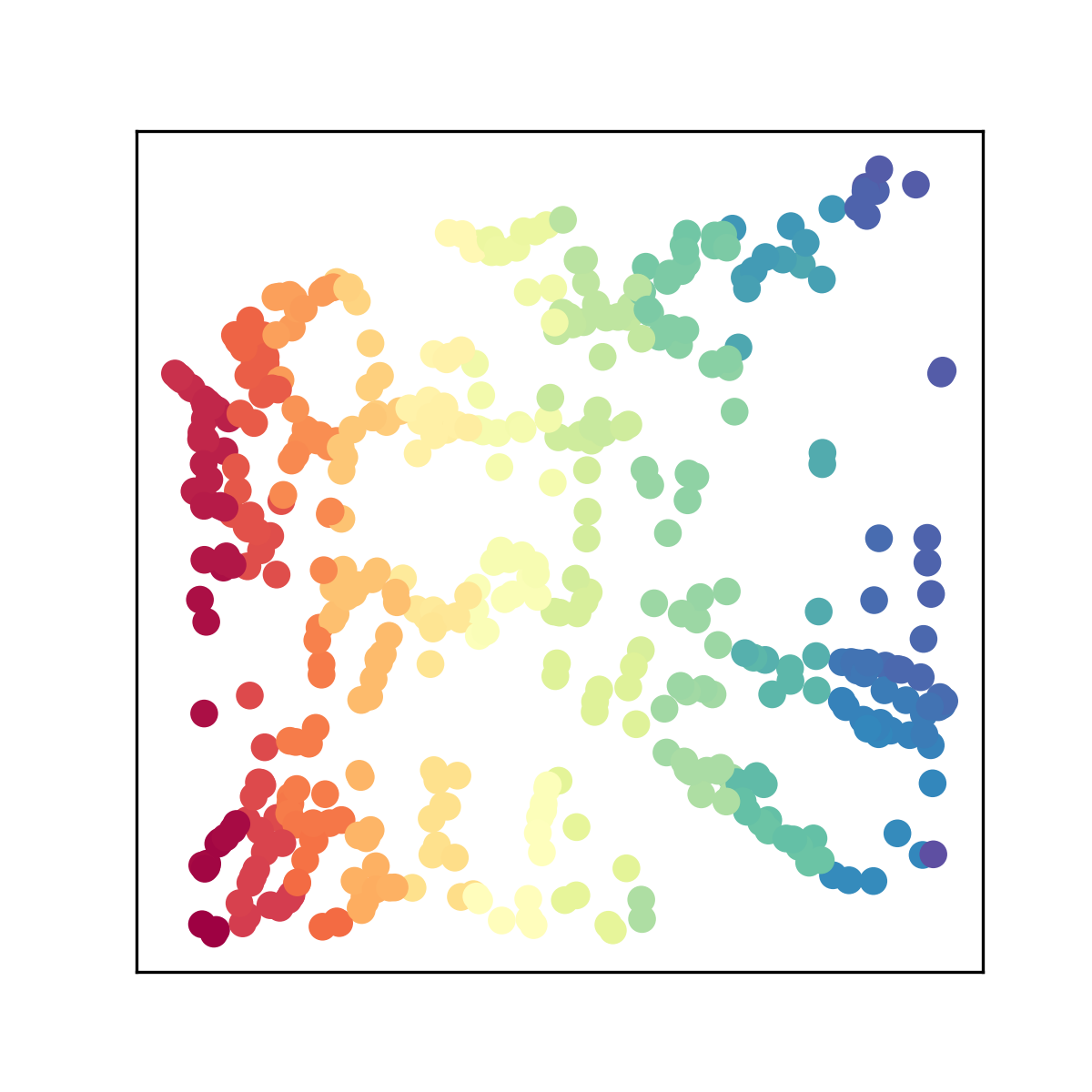} &
        \includegraphics[height=1.7cm, trim={1.4cm 1.25cm 1.2cm 1.35cm},clip]{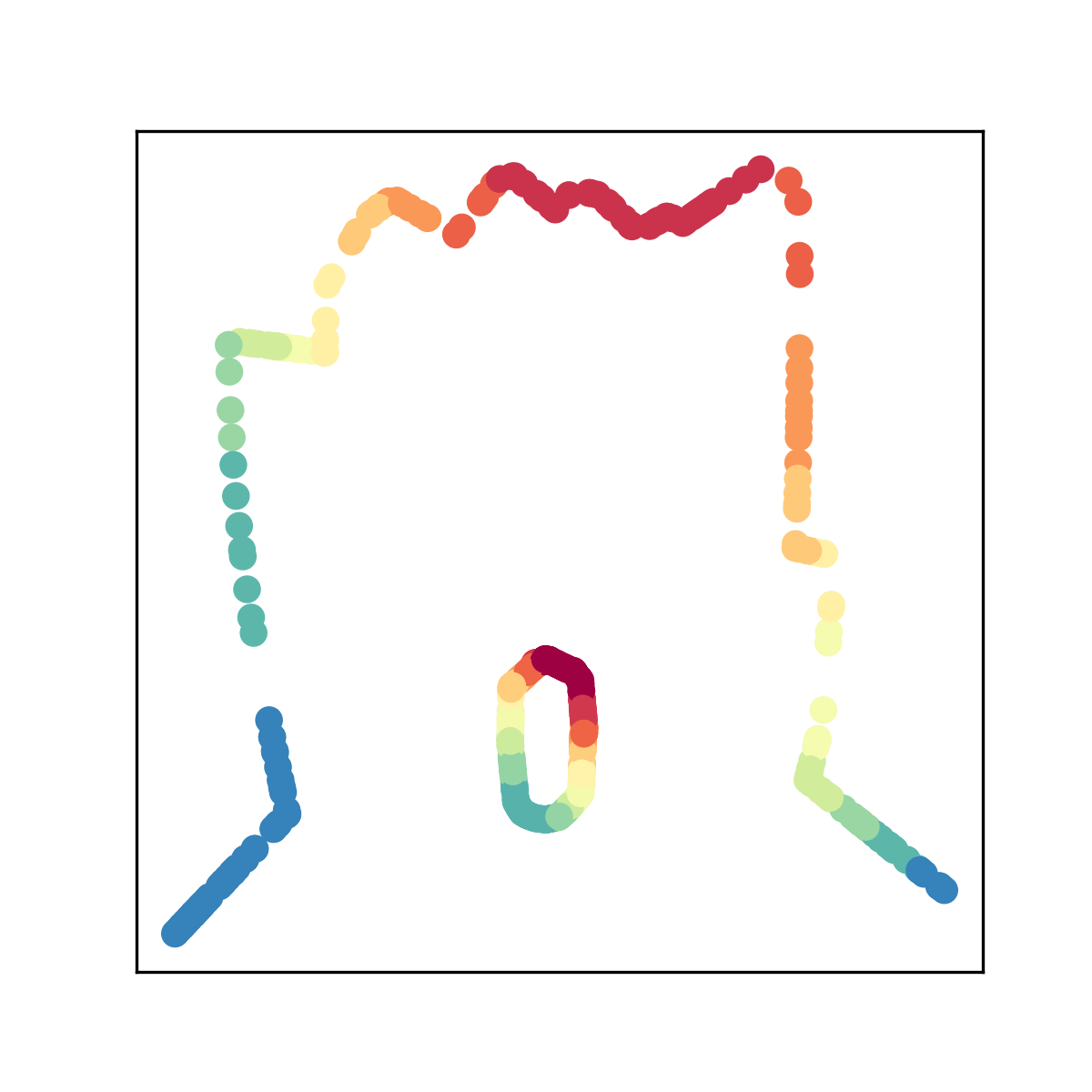} &
        \includegraphics[height=1.7cm, trim={1.4cm 1.25cm 1.2cm 1.35cm},clip]{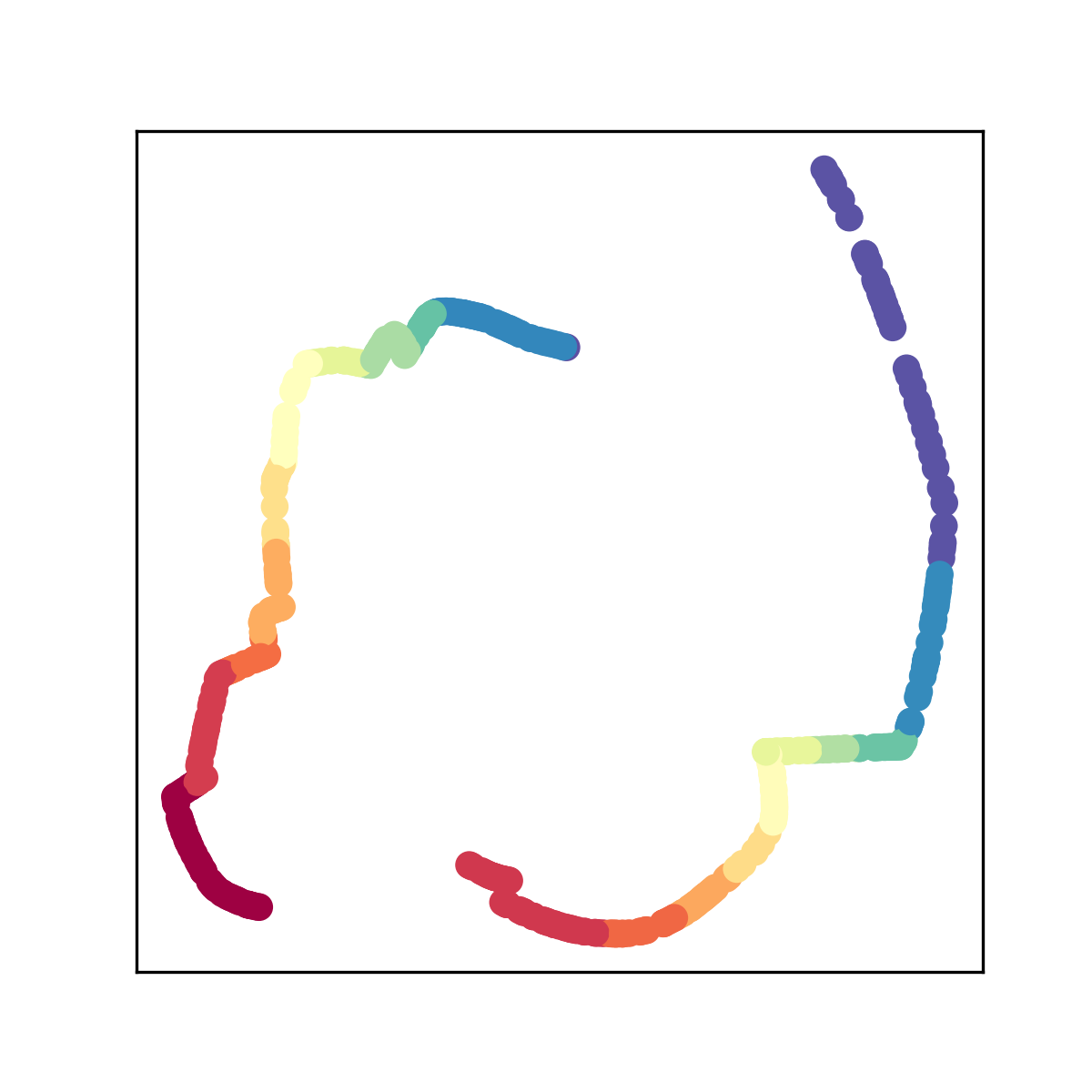} \\
        \hline
        Laplacian Eigenmap &
        \includegraphics[height=1.7cm, trim={1.4cm 1.25cm 1.2cm 1.35cm},clip]{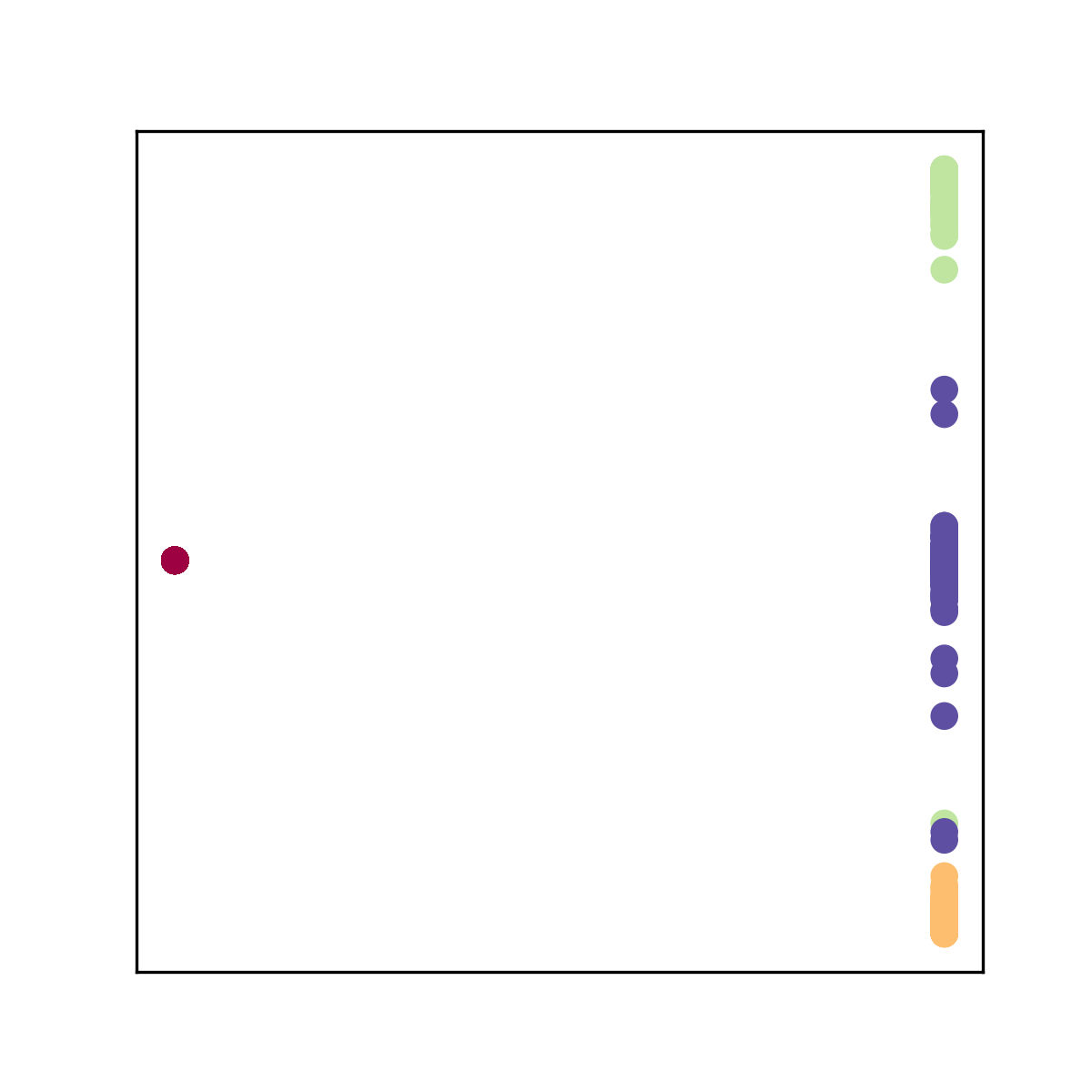} &
        \includegraphics[height=1.7cm, trim={1.4cm 1.25cm 1.2cm 1.35cm},clip]{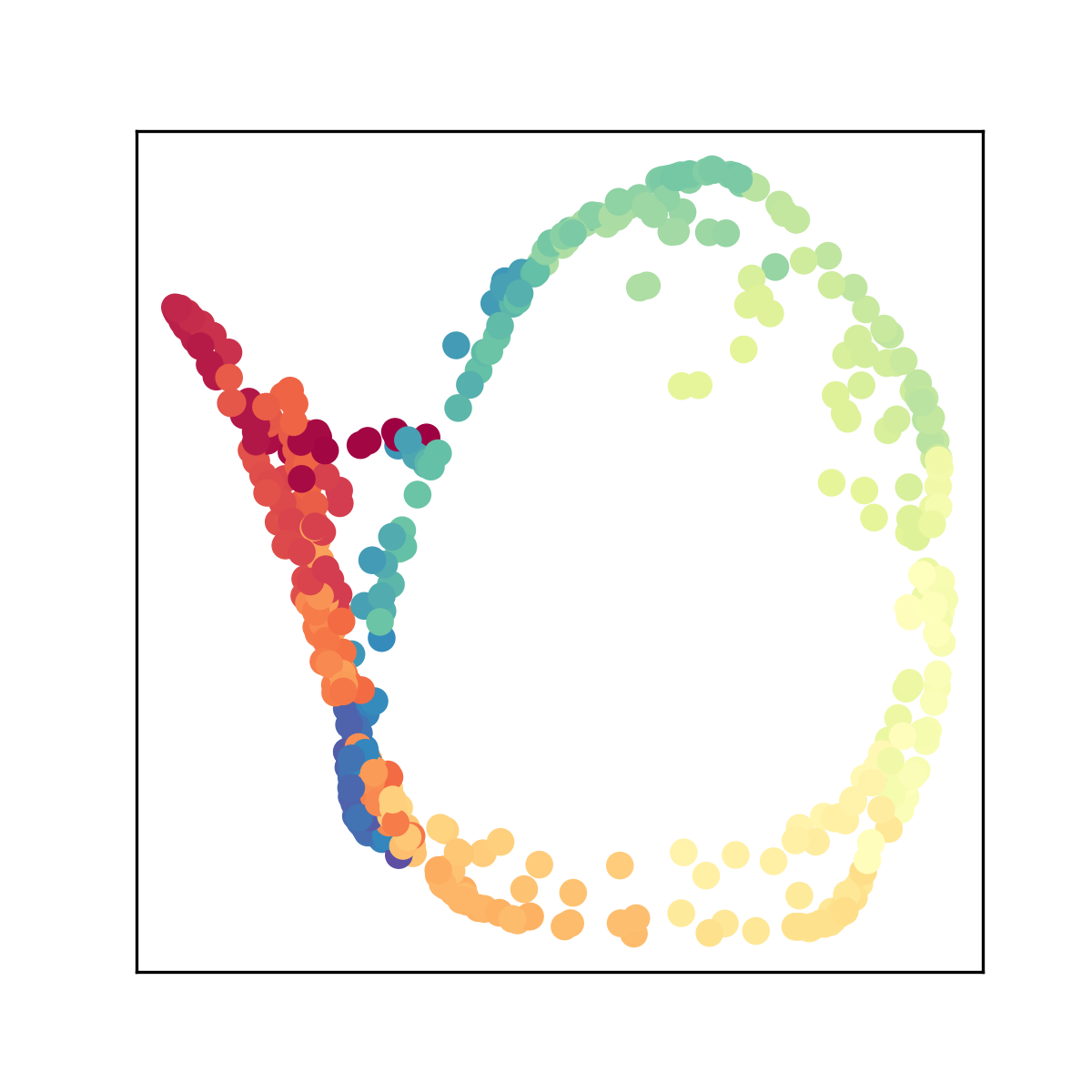} &
        \includegraphics[height=1.7cm, trim={1.4cm 1.25cm 1.2cm 1.35cm},clip]{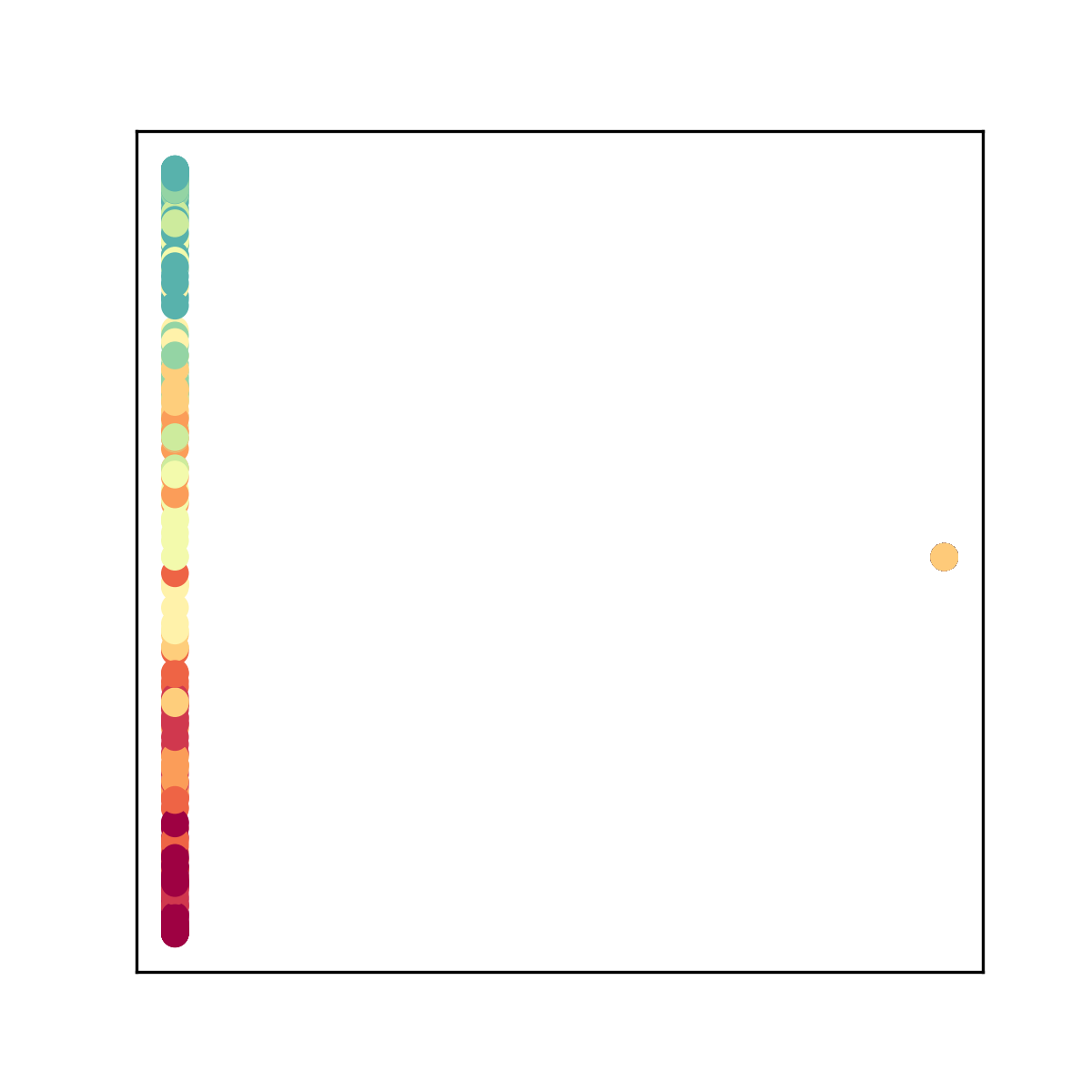} &
        \includegraphics[height=1.7cm, trim={1.4cm 1.25cm 1.2cm 1.35cm},clip]{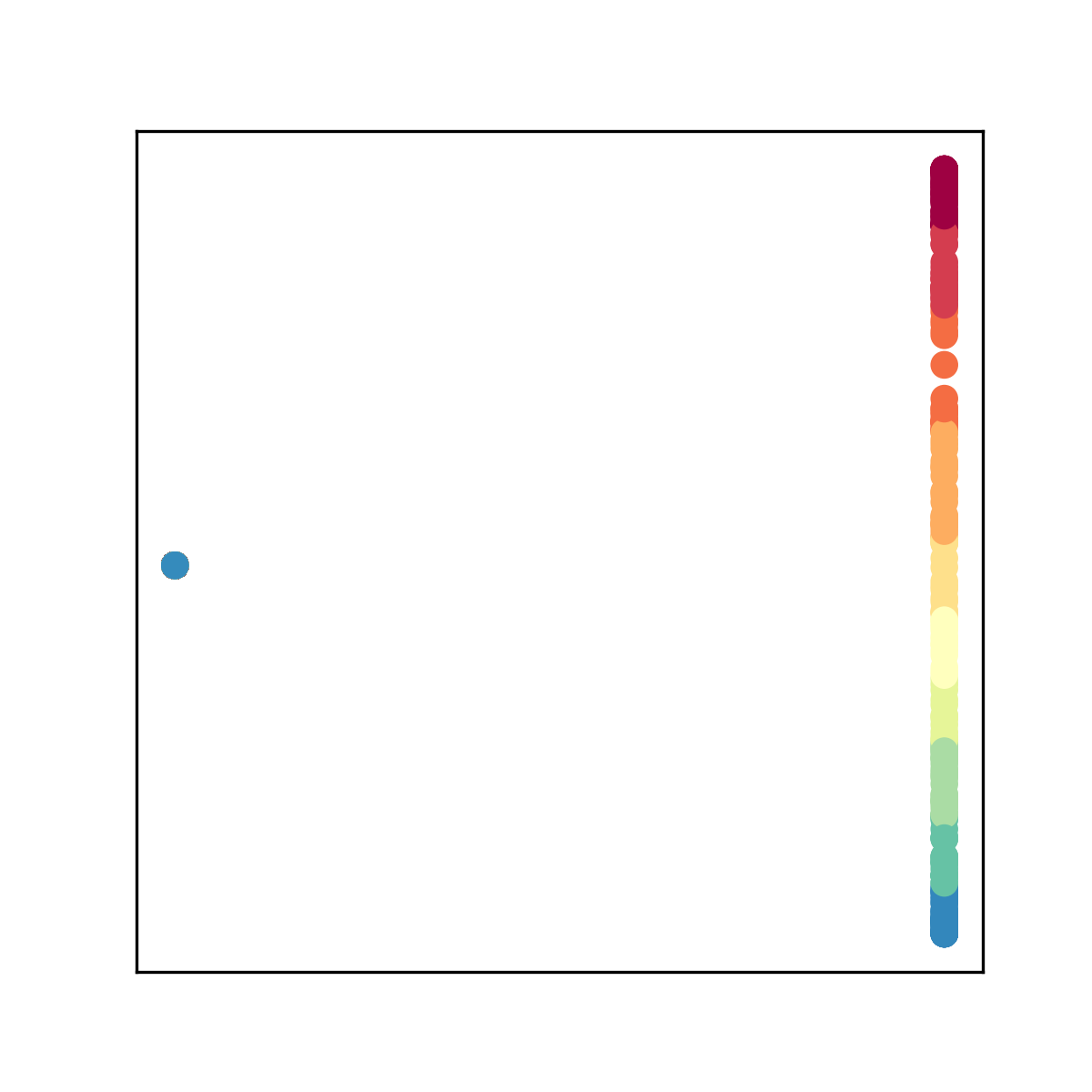} \\
        \hline
        UMAP &
        \includegraphics[height=1.7cm, trim={1.4cm 1.25cm 1.2cm 1.35cm},clip]{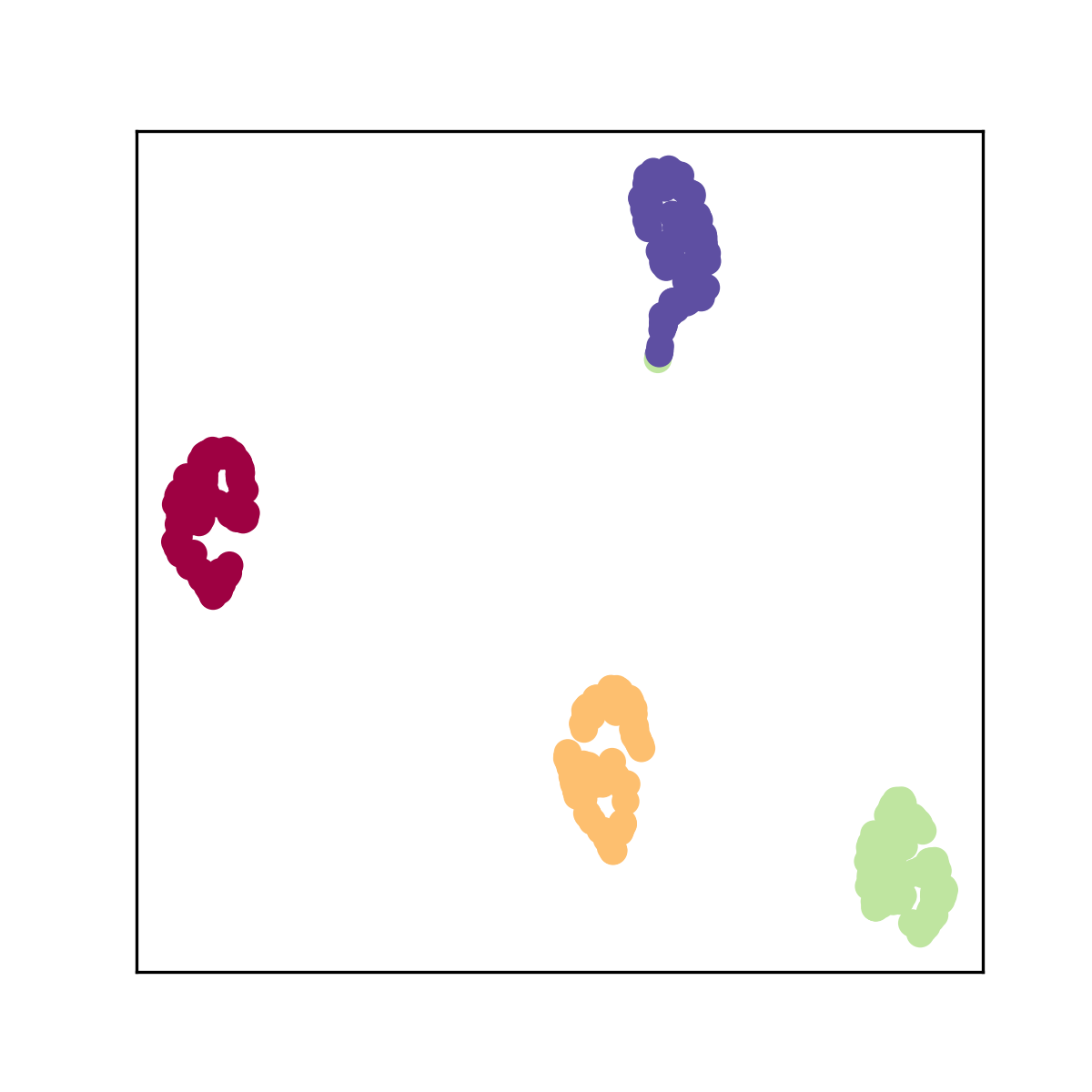} &
        \includegraphics[height=1.7cm, trim={1.4cm 1.25cm 1.2cm 1.35cm},clip]{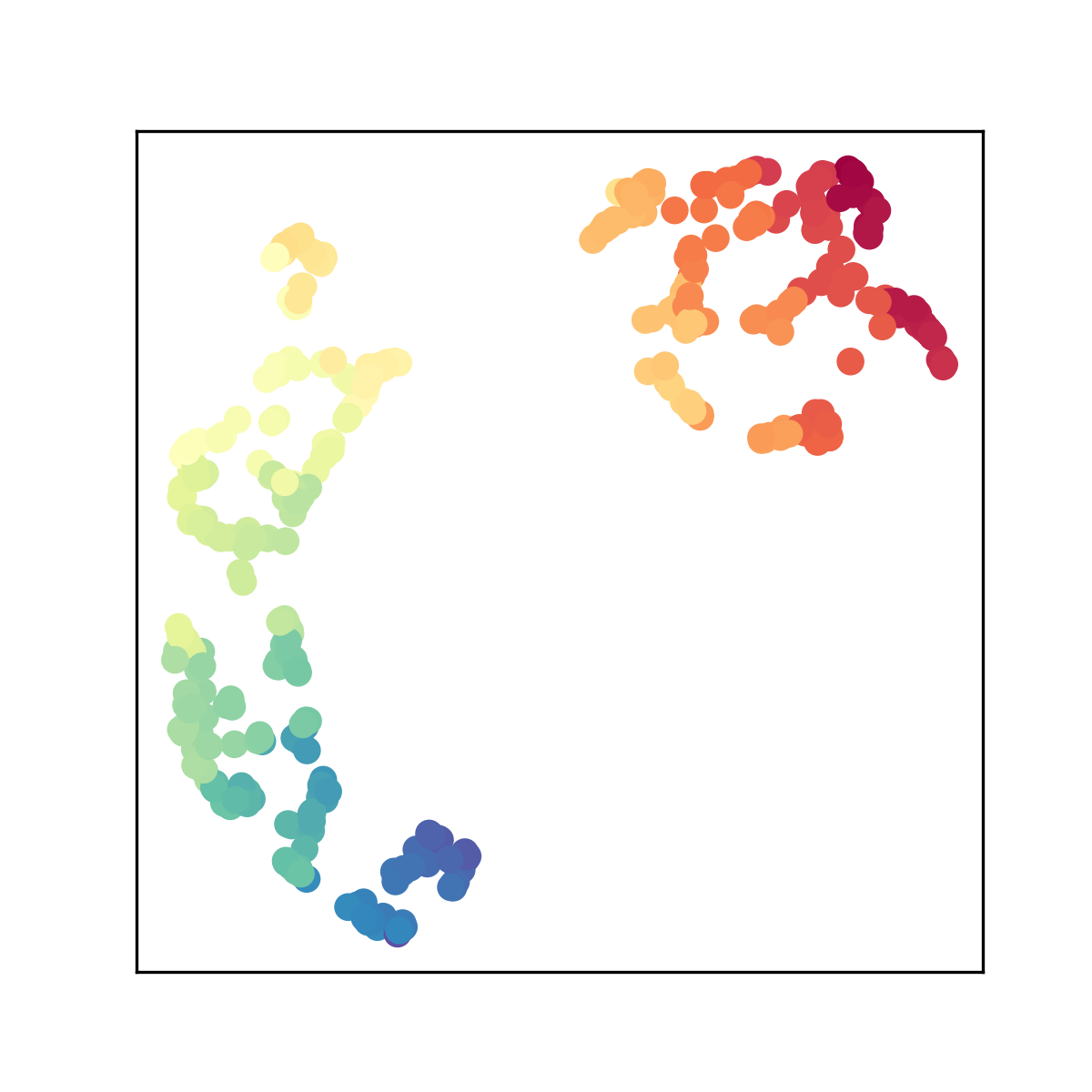} &
        \includegraphics[height=1.7cm, trim={1.4cm 1.25cm 1.2cm 1.35cm},clip]{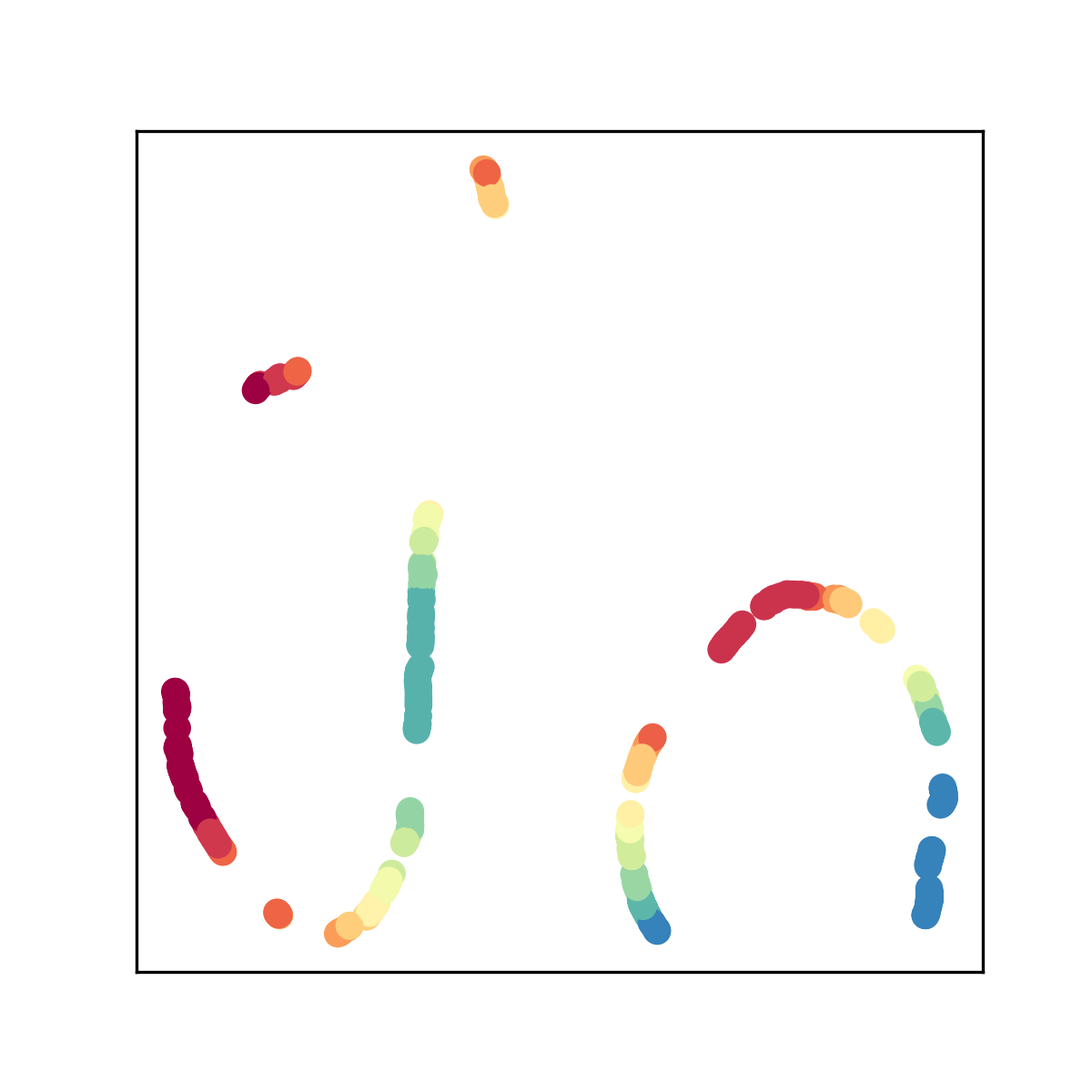} &
        \includegraphics[height=1.7cm, trim={1.4cm 1.25cm 1.2cm 1.35cm},clip]{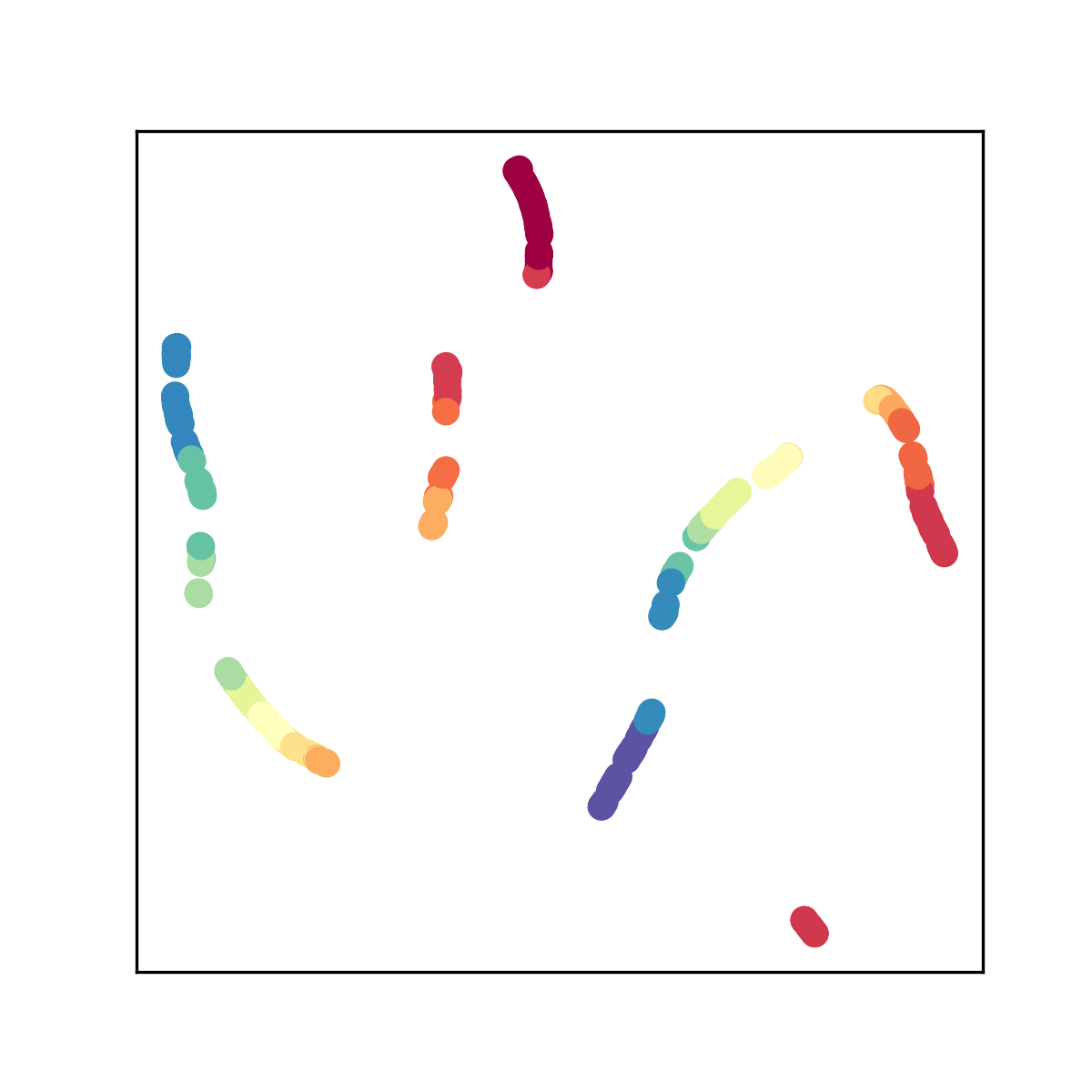} \\
        \hline
        DenseMAP &
        \includegraphics[height=1.7cm, trim={1.4cm 1.25cm 1.2cm 1.35cm},clip]{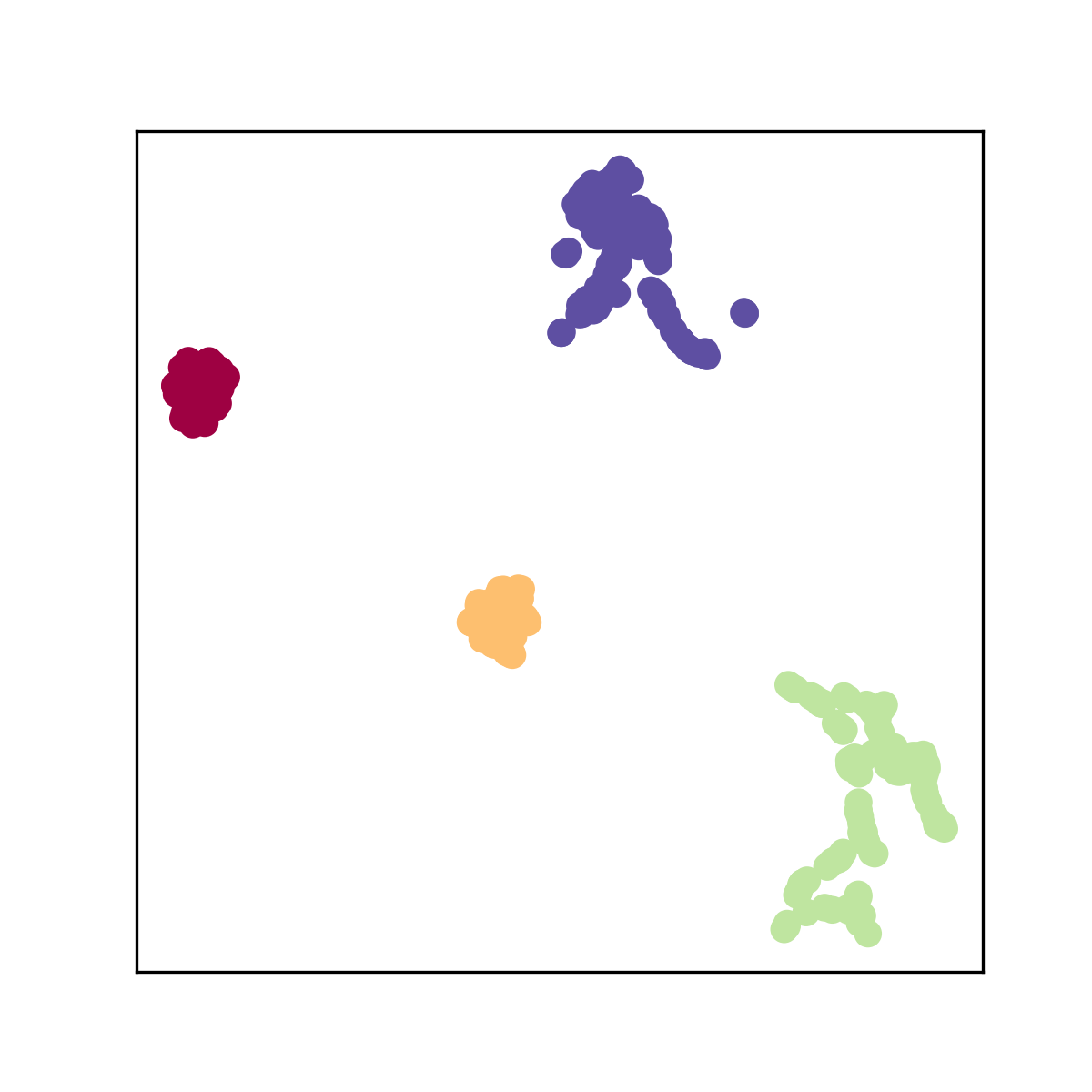} &
        \includegraphics[height=1.7cm, trim={1.4cm 1.25cm 1.2cm 1.35cm},clip]{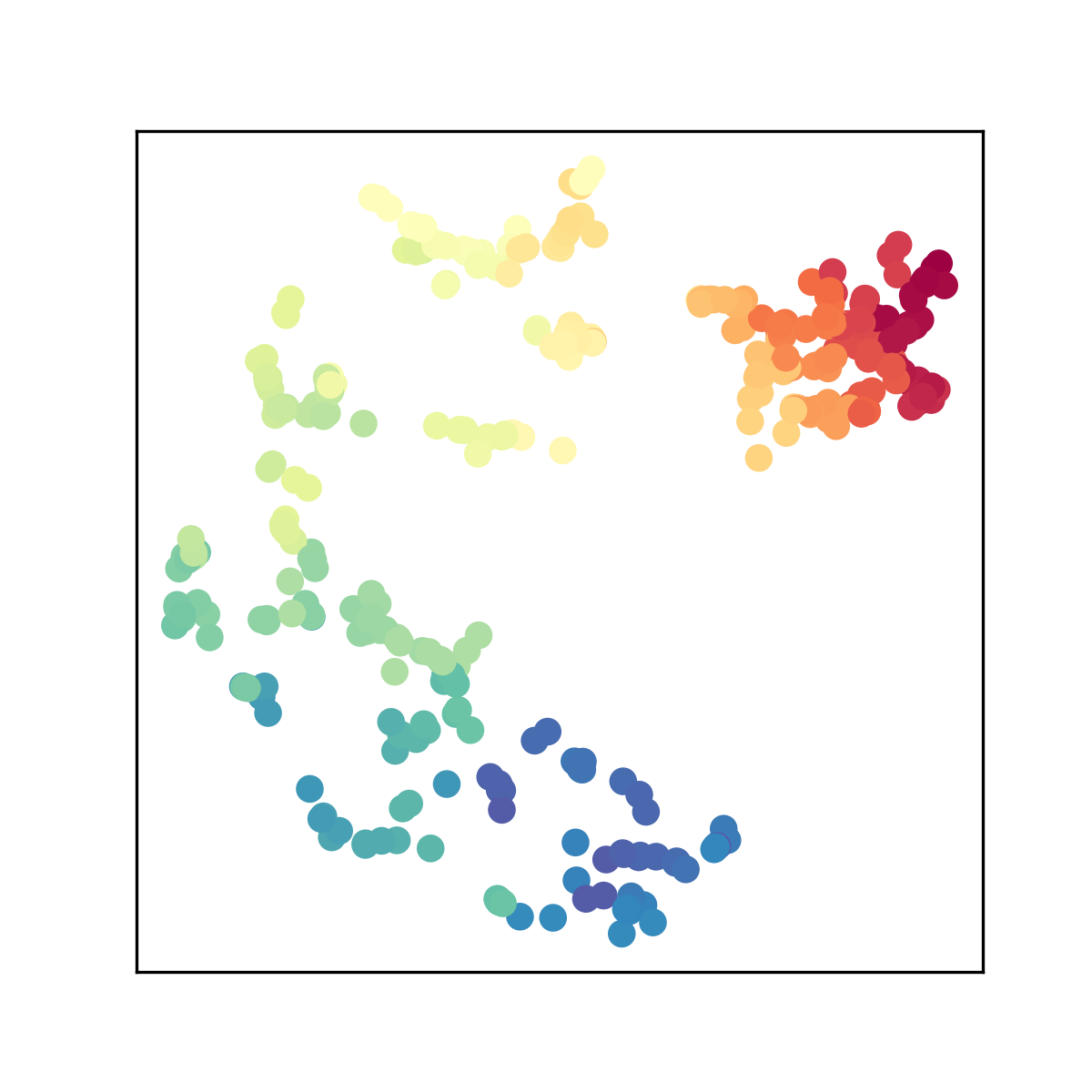} &
        \includegraphics[height=1.7cm, trim={1.4cm 1.25cm 1.2cm 1.35cm},clip]{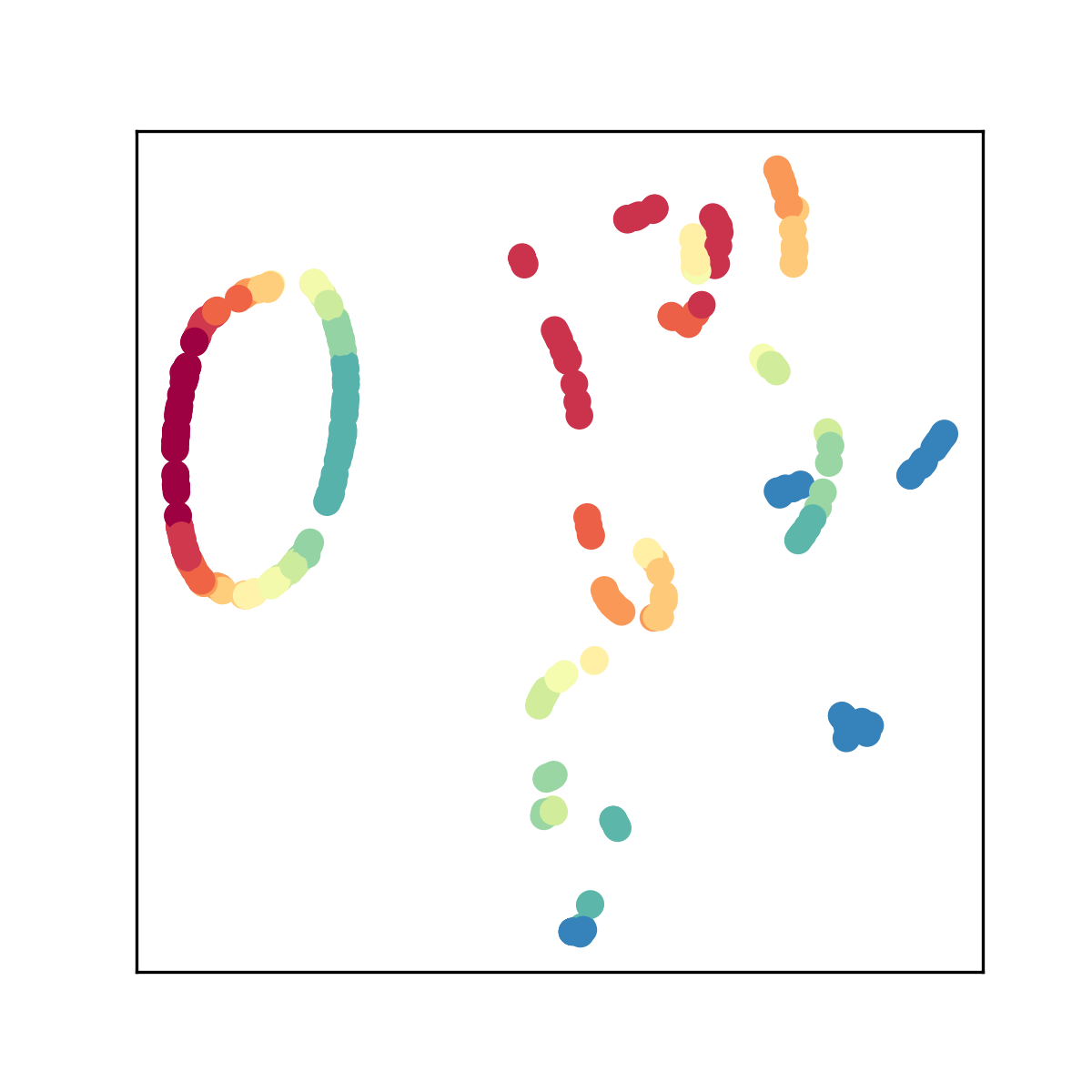} &
        \includegraphics[height=1.7cm, trim={1.4cm 1.25cm 1.2cm 1.35cm},clip]{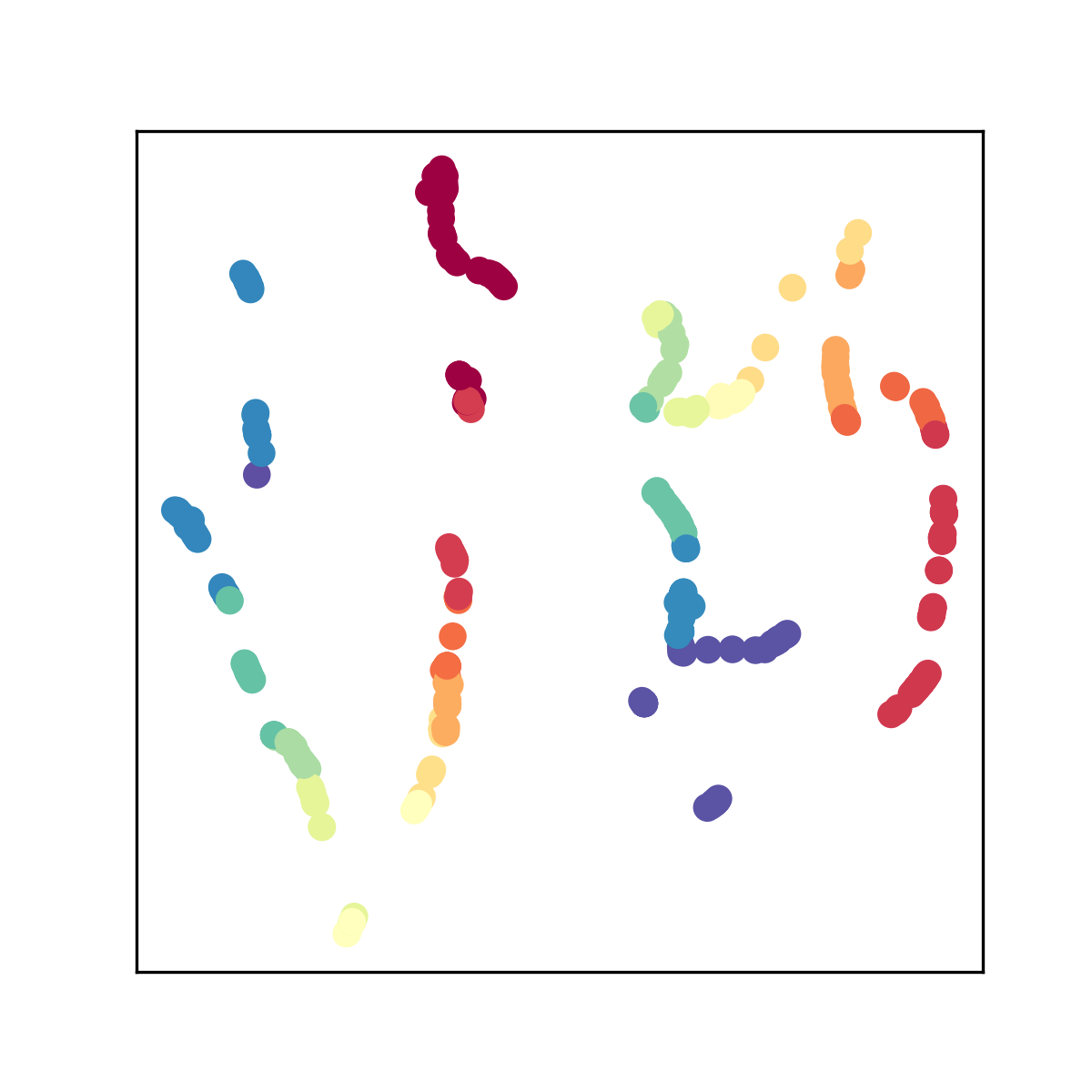} \\
        \hline
    \end{tabular}
    \caption{Node representation learning for 4 synthetic datasets: Blobs, Swissroll, Circles, and Moons. Each image represents a different method of visualization.}
    \label{fig:synthetic_appendix}
\end{figure*}
\renewcommand{\arraystretch}{1}

\end{document}